\documentclass{article}

\usepackage[main, final]{neurips_2026}
\usepackage[utf8]{inputenc} % allow utf-8 input
\usepackage[T1]{fontenc}    % use 8-bit T1 fonts
\usepackage{hyperref}       % hyperlinks
\usepackage{url}            % simple URL typesetting
\usepackage{booktabs}       % professional-quality tables
\usepackage{amsfonts}       % blackboard math symbols
\usepackage{nicefrac}       % compact symbols for 1/2, etc.
\usepackage{microtype}      % microtypography
\usepackage{xcolor}         % colors
\usepackage{graphicx}
\usepackage{amsmath}
\usepackage{booktabs}
\usepackage{multirow}
\usepackage[table]{xcolor}
\usepackage{colortbl} % 建议显式加上，避免某些模板不兼容
\usepackage{caption}
\usepackage{subcaption}
\usepackage[normalem]{ulem}
\useunder{\uline}{\ul}{}

\usepackage{makecell}
\usepackage{textcomp}
\usepackage{longtable}
\usepackage{relsize} 
\usepackage{adjustbox}

\title{From Retrieval to Recognition: \\ Knowledge-Enhanced Post-training of Time Series Foundation Models for Classification}

\author{
\textbf{Zhen Liu}\textsuperscript{1},
\textbf{Yucheng Wang}\textsuperscript{2,3},
\textbf{Yingpeng Du}\textsuperscript{1},
\textbf{Tianjun Wei}\textsuperscript{1},
\\[-1mm]
\textbf{Qianli Ma}\textsuperscript{4,*},
\textbf{Min Wu}\textsuperscript{2,3,*},
\textbf{Jie Zhang}\textsuperscript{1},
\textbf{Zhu Sun}\textsuperscript{5}
\\[1mm]
\textsuperscript{1}Nanyang Technological University, Singapore
\\
\textsuperscript{2}Institute of Advanced Intelligence and Computing (IAIC), A*STAR, Singapore
\\
\textsuperscript{3}AI Center of Excellence for Manufacturing (AIMfg), A*STAR, Singapore
\\
\textsuperscript{4}South China University of Technology, China
\\
\textsuperscript{5}Singapore University of Technology and Design, Singapore
\\
\ttfamily{cszhenliu@foxmail.com}, \ttfamily{qianlima@scut.edu.cn}, \ttfamily{wumin@a-star.edu.sg}
}

\begin{document}

\maketitle

\begingroup
\renewcommand{\thefootnote}{*}
\footnotetext{Corresponding authors.}
\endgroup

\begin{abstract}
  Time series foundation models (TSFMs) have achieved notable progress in classification tasks through large-scale pretraining. However, their post-training practice mainly focuses on implicit parameter fine-tuning, where task-relevant information is entangled with abundant irrelevant knowledge in vast parameter spaces. This may hinder TSFMs from precisely activating useful knowledge for downstream classification. 
  To this end, we propose Retrieval-to-Recognition (R2R), a plug-and-play post-training paradigm that adapts pretrained TSFMs to time series classification.
  Rather than pretraining a new foundation model, R2R improves how existing TSFMs are post-trained: it introduces a decoupled query encoder to extract discriminative patterns and retrieve instance-level numerical knowledge from a constructed knowledge base. The retrieved explicit knowledge is then fused with the TSFM’s implicit parametric embeddings, enabling joint knowledge transfer at both the parameter and feature levels. In this way, R2R allows TSFMs to recall task-relevant patterns learned during pretraining, improving the efficiency of knowledge utilization in downstream classification transfer. Experiments on 158 time series datasets show that R2R consistently improves the classification performance of state-of-the-art TSFMs. Ablation and knowledge-enhanced studies further validate the effectiveness of the proposed framework.
  The source code is available at \url{https://github.com/ZLiu21/R2R}.
\end{abstract}

\section{Introduction}

Time series data are widespread in critical domains such as traffic monitoring~\cite{li2015trend}, medical diagnosis~\cite{wang2024medformer}, and action recognition~\cite{yang2015deep,zhang2024unimts}.  
As a core task in time series analysis, time series classification (TSC) aims to extract discriminative patterns in sequences with complex temporal dynamics to distinguish categories~\cite{mohammadi2024deep}.
During the past decades, TSC methods have evolved from distance-based approaches~\cite {lines2015time,cuturi2017soft}, to feature-engineering strategies such as the Rocket family~\cite{dempster2020rocket,dempster2021minirocket,tan2022multirocket}, and to supervised deep learning models~\cite{ismail2020inceptiontime,chen2024timemil,jungmambasl2026}.
More recently, inspired by the success of pretraining, TSFMs have emerged as a promising research direction~\cite{ma2024survey,liang2024foundation,chen2026fedal}.
This shift advances TSC toward general-purpose modeling, utilizing large-scale pretraining to support diverse classification tasks.

\begin{figure}[]
\centering
\includegraphics[width=5.5in]{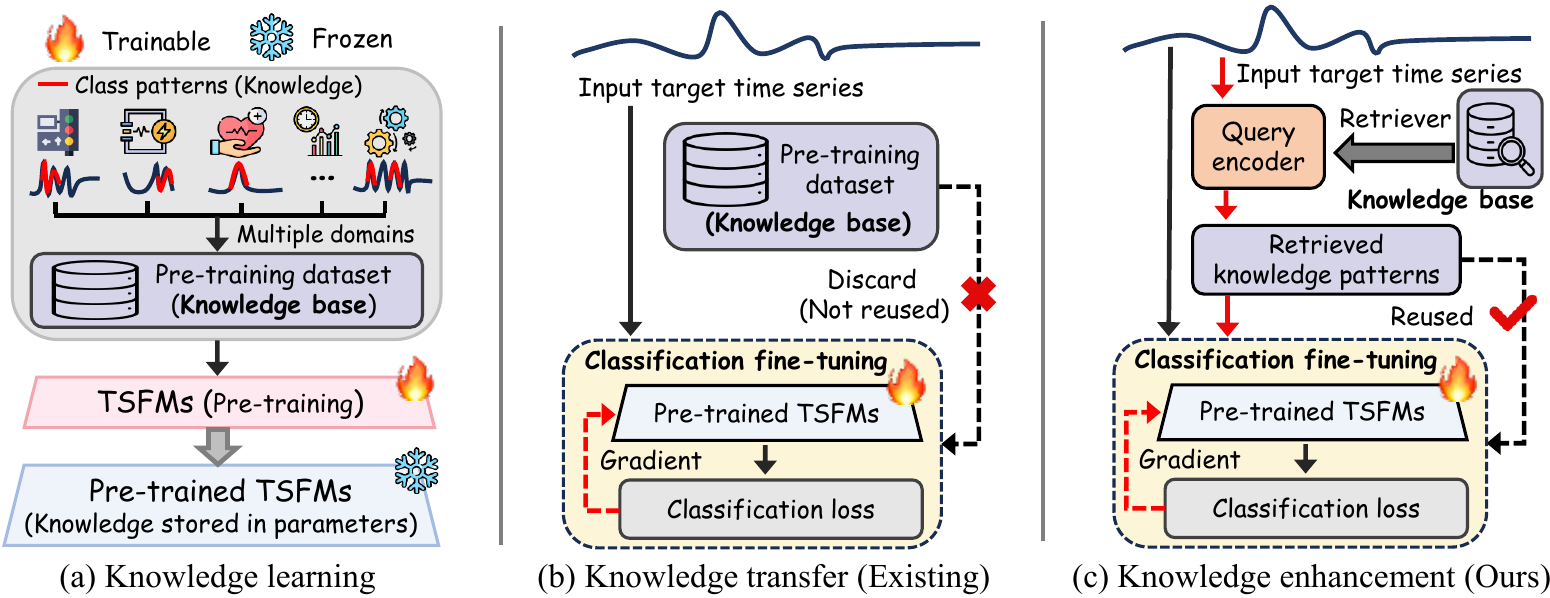}
\caption{Illustration of knowledge paradigms in TSFMs for classification.
(a) \textbf{Knowledge learning}: TSFMs learn discriminative class patterns from multi-domain pretraining data, storing knowledge implicitly in model parameters.
(b) \textbf{Knowledge transfer}: Existing methods adapt this knowledge to target tasks via parameter fine-tuning, without explicitly reusing pretraining data knowledge.
(c) \textbf{Knowledge enhancement}: R2R enhances parameter transfer with feature-level retrieval, explicitly leveraging numerical patterns from a pretraining knowledge base to boost recognition.
Stage (a) is pretraining; (b) and (c) are post-training, and this work targets the (b)$\rightarrow$(c) transition.
}
\label{fig_1_rekin_moti}
\end{figure}

Existing TSFMs for classification typically employ a pretraining paradigm that exploits extensive, multi-domain datasets~\cite{goswami2024moment,lin2024nutime}. This process can be regarded as a critical knowledge learning stage (Figure~\ref{fig_1_rekin_moti}a), where the model captures rich discriminative patterns from the pretraining dataset~\cite{feofanov2025mantis,liu2026unishape}.
Through this stage, explicit class patterns in raw time series are compressed into implicit, parameterized knowledge encoded within model weights~\cite{gao2024units,xie2026cauker}. Hence, pretrained TSFMs serve as knowledge carriers, providing superior parameter initialization compared to training from scratch.

Building on pre-trained TSFMs,  existing approaches~\cite{lin2024nutime,
feofanov2025mantis,liu2026unishape} typically employ parameter fine-tuning for knowledge transfer to downstream TSC tasks (Figure~\ref{fig_1_rekin_moti}b). This adaptation is the post-training stage of foundation models~\cite{chu2025sft,liu2026posttraining}, for which supervised fine-tuning remains the dominant instantiation in TSC.
However, parameter-only fine-tuning discards pretraining data, limiting access to informative numerical patterns in the source corpus. Task-relevant knowledge remains entangled with irrelevant information in large parameter spaces, hindering precise activation of useful knowledge for transfer.
Moreover, prior studies~\cite{gao2024units,ekambaram2026tspulse} show that simply scaling model parameters does not consistently improve TSC performance. These observations motivate a key question: \textit{beyond parameter-based transfer, can explicit data-level knowledge from the pretraining dataset be leveraged to enable more effective transfer?}

A straightforward way to incorporate data-level knowledge is to reuse pretraining data during fine-tuning.
However, large-scale pretraining datasets make direct reuse inefficient for transfer. In TSC, performance mainly depends on capturing class-relevant patterns in input data, such as discriminative subsequences~\cite{wen2024abstracted,le2024shapeformer,pandey2026timesliver}. These patterns are often sparsely distributed and mixed with substantial noise and class-irrelevant variations~\cite{schafer2015boss,huangshapex2025}. Inspired by retrieval-augmented methods in pretrained language models~\cite{lewis2020retrieval,cheng2025survey}, we argue that constructing query embeddings for input time series and retrieving feature-similar class patterns from pretraining datasets, yielding explicit instance-level knowledge. This retrieval process filters irrelevant samples while focusing on class-relevant patterns, providing more direct data-level knowledge than parameter-only fine-tuning.

Therefore, we propose \underline{\textbf{R}}etrieval-\underline{\textbf{to}}-\underline{\textbf{R}}ecognition (\textbf{R2R}), a plug-and-play post-training framework that advances pretrained TSFMs from retrieval to recognition (Figure~\ref{fig_1_rekin_moti}c). 
R2R goes beyond parameter-based transfer by introducing a dedicated query encoder to efficiently retrieve class-relevant patterns from large-scale pretraining datasets.
The query encoder captures local and global features with positional information to identify feature-level class patterns for each input time series, enabling explicit acquisition of data-level knowledge.
The retrieved patterns are then integrated with the pretrained TSFM’s parameter-derived embeddings, allowing the TSFM to effectively recall task-relevant patterns from the pretraining corpus for accurate transfer.
This retrieval-driven paradigm mitigates the limitations of parameter-only approaches by enabling feature-level transfer, offering a new post-training paradigm for TSC.

Our main contributions are summarized as follows:

\begin{itemize}
    \item We introduce R2R, a plug-and-play post-training framework for TSC that enhances pretrained TSFMs by utilizing external numerical knowledge derived from a constructed pretraining 
    knowledge base, facilitating instance-level knowledge integration beyond parameter-only transfer, without modifying the original pretraining process of TSFMs.

    \item We design a decoupled query encoder with a gradient-free backbone to efficiently extract and retrieve discriminative numerical patterns, enabling multi-domain inputs to access class-relevant patterns for each query instance and to bridge retrieval with downstream TSC.

    \item Experiments on 158 datasets demonstrate that R2R consistently improves classification performance of existing pretrained TSFMs in both supervised fine-tuning and zero-shot feature extraction, with further analyses confirming the effectiveness of each module.
\end{itemize}

\section{Related work}

\subsection{Foundation models for time series classification}

Deep learning has made significant progress in TSC due to its strong representation learning capability~\cite{ismail2019deep,formertime2023,early2024inherently,svglL2026,tran2026mix}. Early studies~\cite{wang2017time,ismail2019deep,ismail2020inceptiontime} demonstrated that end-to-end convolutional models serve as strong supervised baselines for TSC.
Subsequent work~\cite{fawaz2018transfer,kashiparekh2019convtimenet,ma2024survey} explored transfer learning, showing that pretraining improves classification performance across datasets. With the rise of large-scale pretraining, TSFMs have become a major research focus~\cite{liang2024foundation}. 
Studies~\cite{goswami2024moment,lin2024nutime,gao2024units} further indicate that models pretrained on large time series corpora learn more general and classification-oriented representations. Building on this paradigm, recent approaches~\cite{zhang2025timesbert,feofanov2025mantis,xie2026cauker,liu2026unishape,feofanov2026mantisv2}
further improve pretraining strategies and model architectures, achieving strong generalization across diverse TSC tasks.
However, existing TSFMs primarily rely on implicit parameter-based transfer, that is, on supervised fine-tuning as the sole post-training mechanism~\cite{liu2026posttraining},
overlooking the potential gains from explicit numerical instances inherent in the pretraining dataset.

\subsection{Time series retrieval augmented models}

Retrieval-augmented methods~\cite{guu2020retrieval,lewis2020retrieval,cheng2025survey} have shown strong performance in NLP by enhancing large language models with task-relevant information retrieved from external knowledge bases. Inspired by this paradigm, recent work~\cite{caoenhancing2026,ma2026www} introduces retrieval mechanisms into time series modeling. 
For forecasting tasks, prior studies~\cite{liu2024retrieval,li2025cross,han2025retrieval,caoenhancing2026} retrieve relevant historical sequences and temporal patterns to guide future value prediction, while other methods~\cite{ningts2025,zhang2025timeraf} build structured time series knowledge bases for cross-domain retrieval and improved zero-shot forecasting.
Despite these advances, retrieval-augmented methods for classification remain underexplored. This gap is non-trivial, as forecasting and classification require fundamentally different retrieval targets: forecasting extrapolates future trends from long historical observations~\cite{han2025retrieval}, whereas classification focuses on identifying discriminative patterns within fixed-length observation windows~\cite{ismail2020inceptiontime}.
To address this limitation, we design a dedicated query encoder to retrieve instance-level discriminative patterns from a class-relevant numerical feature space. The retrieved knowledge is integrated into the fine-tuning process of pretrained TSFMs, enabling a retrieval-to-recognition framework for TSC.

\section{Problem definition}

This work focuses on univariate TSC using TSFMs~\cite{liang2024foundation,ismail2019deep}. Let \(\mathbf{x} \in \mathbb{R}^L\) denote a univariate time series sample of length $L$. The goal is to predict its corresponding class label \(y \in \{1, 2, \dots, C\}\), where \(C\) represents the number of classes in the target downstream dataset \(\mathcal{D}_t\).
Given a pretrained TSFM denoted by \(\mathcal{F}(\cdot)\), conventional paradigms facilitate knowledge transfer solely through parameter fine-tuning. In contrast, we introduce an explicit knowledge base \(\mathcal{D}_k = \{\mathbf{x}_i\}_{i=1}^N\) constructed from large-scale, multi-domain pretraining corpora~\cite{lin2024nutime}. 
For a given query \(\mathbf{x}\), a retrieval mechanism is designed to identify 
numerically similar and class-discriminative patterns from \(\mathcal{D}_k\), 
yielding a knowledge-enhanced representation \(\mathbf{z}_{\text{aug}}\).
The final classification is performed by combining the implicit parametric knowledge 
stored in \(F(\cdot)\) with the retrieved explicit numerical knowledge \(\mathbf{z}_{\text{aug}}\), enabling 
effective knowledge transfer on the target dataset \(\mathcal{D}_t\).

\section{Methodology}

\begin{figure}
\centering
\includegraphics[width=5.5in]{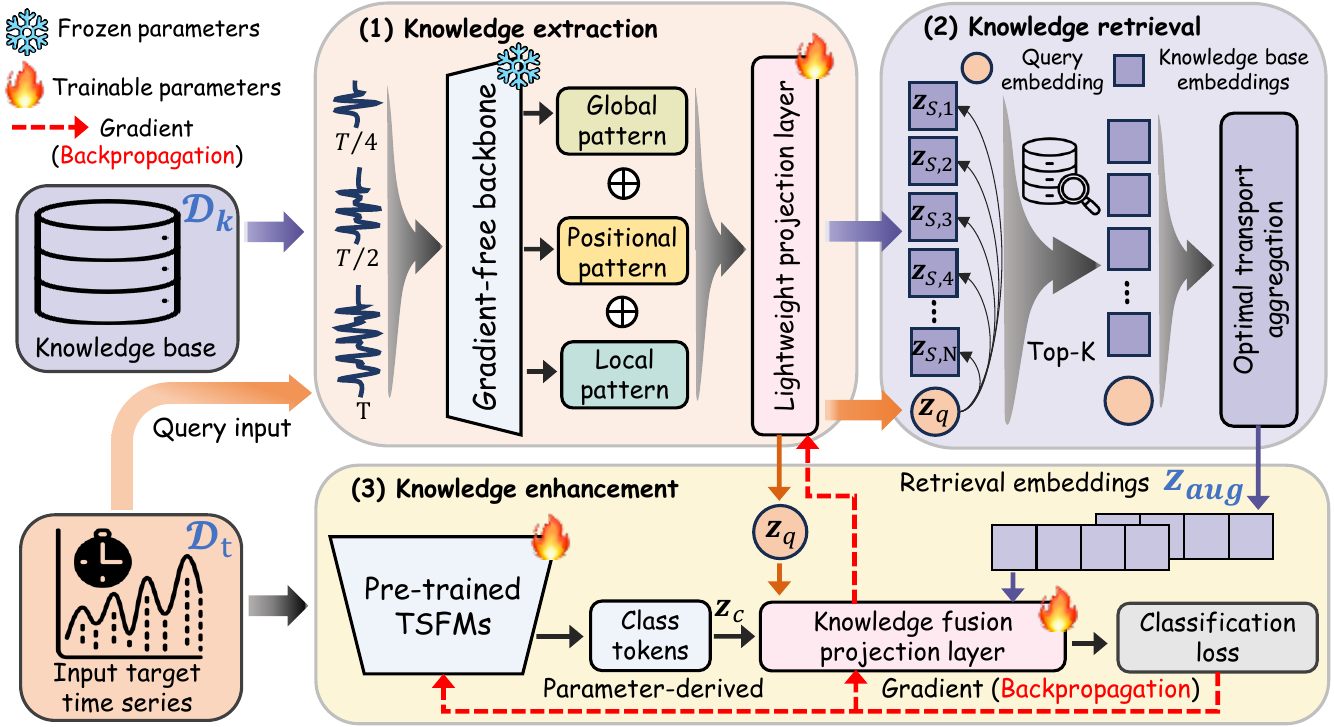}
\caption{ The overall architecture of the R2R framework. The knowledge extraction and retrieval modules constitute the query encoder. R2R uses the output of the query encoder as feature-level knowledge and fuses it with parameter-level class-token embeddings from pre-trained TSFMs for knowledge enhancement.
}
\label{fig_2_rekin}
\end{figure}

\subsection{Overview}

As shown in Figure~\ref{fig_2_rekin}, R2R consists of three modules: \textit{knowledge extraction} (§\ref{sec42}), \textit{knowledge retrieval} (§\ref{sec43}), and \textit{knowledge enhancement} (§\ref{sec44}). 
The knowledge extraction module (§\ref{sec42}) applies multi-scale downsampling to the input data and uses a gradient-free backbone to extract class-relevant features through local, global, and positional information.
These features are projected into compact embeddings for both query embedding $\mathbf{z}_q$ and knowledge base construction.
The knowledge retrieval module (§\ref{sec43}) uses knowledge base embeddings as keys and $\mathbf{z}_q$ to retrieve Top-$K$ similar instances. Optimal transport~\cite{Cuturi2013ot} is applied to compute distances and produce a weighted aggregation, yielding the augmented embedding $\mathbf{z}_{\text{aug}}$.
The knowledge enhancement module (§\ref{sec44}) fuses the TSFM class token $\mathbf{z}_c$, $\mathbf{z}_q$, and $\mathbf{z}_{\text{aug}}$ via a fusion layer. Finally, the model is trained with a classification loss, enabling parameter transfer while incorporating retrieved numerical knowledge for downstream TSC.

\subsection{Knowledge extraction}\label{sec42}

The knowledge extraction module maps both time series samples from the pretraining knowledge base $\mathcal{D}_k$ and target query inputs $\mathbf{x} \in \mathbb{R}^{L}$ from $\mathcal{D}_t$ into compact, class-discriminative embeddings for efficient retrieval. For this purpose, we adopt a gradient-free backbone as the core feature extractor, operating with frozen parameters to extract features from the input time series without training. Prior work~\cite{dempster2020rocket,dempster2021minirocket,tan2022multirocket} shows that convolutional models with randomly or fixed initialized parameters provide strong feature representations for TSC. Similarly, pretrained TSFMs~\cite{feofanov2025mantis,liu2026unishape,ekambaram2026tspulse} can produce discriminative embeddings without fine-tuning.
However, these approaches primarily focus on local features from single-scale inputs, often neglecting global dependencies and positional structure. To mitigate this issue, we design the knowledge extraction module to capture discriminative numerical information from three aspects: local patterns, global patterns, and positional patterns.

\paragraph{Local patterns.}
We employ a gradient-free backbone to extract local features directly from the raw single-scale input $\hat{\mathbf{x}} \in \mathbb{R}^{L}$, where $\hat{\mathbf{x}}$ may originate from either the pretraining knowledge base $\mathcal{D}_k$ or the target domain $\mathcal{D}_t$. The backbone is model-agnostic with frozen parameters. The resulting local embedding is defined as
\begin{equation}
\mathbf{f}_{\text{raw}} = \phi(\hat{\mathbf{x}}),
\label{eq:f_raw}
\end{equation}
where $\phi(\cdot)$ denotes the gradient-free feature, which takes an input of arbitrary length and returns a fixed-dimensional vector by pooling over the temporal axis.

Although $\mathbf{f}_{\text{raw}}$ captures local temporal patterns, it may be less sensitive to variations in signal intensity and local energy distribution~\cite{huang1998empirical}. To address this, we introduce an instantaneous envelope feature that emphasizes amplitude-related characteristics. Specifically, given $\hat{\mathbf{x}}$, we compute
\begin{equation}
f_{\text{env}} = \max_{t} \big[ \text{AvgPool1D}(|\hat{\mathbf{x}}|, l) \big]_{t} \in \mathbb{R},
\label{eq:fenv}
\end{equation}
where $l$ denotes the sliding window (or fixed-length subsequence) size, and $[\,\cdot\,]_{t}$ denotes the $t$-th pooled entry. This feature highlights local energy concentration and peak amplitude variations~\cite{abboud2017envelope}, capturing signal intensity dynamics from a feature engineering perspective.

\paragraph{Global patterns.}
To capture global temporal patterns beyond local embeddings, we construct multi-scale inputs via temporal downsampling. Existing methods~\cite{dempster2020rocket,feofanov2025mantis}, including the local pattern module, mainly operate on the original sequence, limiting their ability to model long-range discriminative patterns. Prior studies~\cite{cui2016multi,guillaume2022random} indicate that downsampling generates sparse sequences that retain key temporal structures while enlarging the receptive field, facilitating global pattern learning over longer horizons.
Given an input $\hat{\mathbf{x}} \in \mathbb{R}^{L}$, we generate scale-specific embeddings as
\begin{equation}
\mathbf{f}_{\text{ds}, s} = \phi\big( \text{Downsample}(\hat{\mathbf{x}}, s) \big), \quad s \in \{2,\dots,\left\lfloor \frac{L}{2} \right\rfloor\},
\label{eq:downsample}
\end{equation}
where $\text{Downsample}(\cdot, s)$ denotes a stride-based temporal subsampling operator that selects sparse time steps at interval $s$, i.e. it keeps every $s$-th sample without applying an anti-aliasing low-pass filter, so it is not a band-limited resampling operation, and $\phi(\cdot)$ is the backbone defined in Eq.~(\ref{eq:f_raw}). This design enables the model to capture broader temporal context by leveraging sparse observations over longer horizons.

However, larger downsampling scales may overlook critical transition points that capture fine-grained temporal dynamics.
To mitigate this issue, we compute first-order differences $\Delta x_t = x_{t+1} - x_t$ and define a turning point indicator $\mathbb{I}\big((\Delta x_t \cdot \Delta x_{t+1}) < 0\big)$ to detect sign changes, and further introduce a turning-point density-based scalar gating mechanism:
\begin{equation}
f_{\text{tp}} = \frac{1}{L-2} \sum_{t=1}^{L-2} \mathbb{I}_t, \quad
\mathbf{f}_{\text{gated}} = \mathbf{f}_{\text{ds},s} \mathbin{\color{blue}\cdot} \tanh \left( 
f_{\text{tp}}
\right),
\end{equation}
which provides a lightweight feature extraction design that enhances global patterns.

\paragraph{Positional patterns.}
In TSC, similar discriminative patterns may correspond to different classes depending on their time point positions~\cite{zuo2023svp}. Hence, we construct a positional feature based on the relative peak location by combining local patterns and turning-point information as follows:
\begin{equation}
f_{\text{pos}} = f_{\text{env}} \oplus f_{\text{tp}} \oplus \frac{\arg\max_{t} |\hat{\mathbf{x}}_t|}{L},
\end{equation}
where $\oplus$ denotes concatenation and the last term denotes the normalized time index of the maximum-amplitude point.
This feature captures the temporal placement of salient patterns and aids in distinguishing classes that differ primarily by pattern location.

Finally, we concatenate the local, global, and positional patterns and map them into a unified embedding space via a lightweight projection layer:
\begin{equation}
\mathbf{z} = \psi \left( \mathbf{f}_{\text{raw}} \oplus \mathbf{f}_{\text{ds}, s} \oplus \mathbf{f}_{\text{gated}} \oplus f_{\text{pos}} \right),
\label{eq:f_z_ke}
\end{equation}
where $\psi(\cdot)$ denotes a learnable projection function. This formulation integrates class-related features into a discriminative embedding for retrieval. Notably, the computation of $\mathbf{f}_{\text{raw}},\mathbf{f}_{\text{ds},  s}, \mathbf{f}_{\text{gated}}, f_{\text{pos}}$ do not require gradient-based optimization and can be efficiently obtained in a training-free manner.

\subsection{Knowledge retrieval}\label{sec43}

Through Eq.~(\ref{eq:f_z_ke}) in the knowledge extraction module, the embeddings of the knowledge base $\mathcal{D}_k$ are collected into the set $\mathcal{Z}_S = \{ \mathbf{z}_{S,i} \in \mathbb{R}^{d} \}_{i=1}^{N}$, while a query input $\mathbf{x} \in \mathbb{R}^{L}$ from $\mathcal{D}_t$ is mapped to a query embedding $\mathbf{z}_{q} \in \mathbb{R}^{d}$. 
The retrieval module aims to efficiently identify the Top-$K$ nearest neighbors of $\mathbf{z}_{q}$ from $\mathcal{Z}_S$, and construct a feature-level augmented representation $\mathbf{z}_{\text{aug}}$.

To perform efficient nearest neighbor search, we first normalize the embeddings obtained from the knowledge extraction module:
\begin{equation}
\tilde{\mathbf{z}}_{S,i} = \frac{\mathbf{z}_{S,i}}{\|\mathbf{z}_{S,i}\|_2}, \quad 
\tilde{\mathbf{z}}_{q} = \frac{\mathbf{z}_{q}}{\|\mathbf{z}_{q}\|_2}.
\end{equation}

The normalized embeddings are indexed via inverted file system with product quantization (IVFPQ)~\cite{jegou2010product}, which partitions the embedding space into $n_{\text{list}}$ Voronoi cells and encodes residuals with product quantization, forming an index structure:
\begin{equation}
\mathcal{I} = \text{IVFPQ}(\mathcal{Z}_S; n_{\text{list}}, m, b),
\label{eq:ivfpq}
\end{equation}
where $m$ denotes the number of subspaces and $b$ the number of bits per subspace. In practice, this index is implemented using a GPU-accelerated inverted file system based on Faiss~\cite{douze2025}, enabling a scalable and efficient creation. 
Given a query embedding, the Top-$K$ nearest neighbors are retrieved from the index $\mathcal{I}$ based on cosine similarity: 
\begin{equation}
\mathcal{N}(\mathbf{z}_{q}) = \operatorname{TopK}_{\mathbf{z}_{S,i} \in \mathcal{I}} \left\langle \tilde{\mathbf{z}}_{q}, \tilde{\mathbf{z}}_{S,i} \right\rangle.
\end{equation}
During querying, only a small subset of Voronoi cells ($n_{\text{probe}} \ll n_{\text{list}}$) is visited, leading to sublinear complexity with respect to the database size. Specifically, the search cost scales as $\mathcal{O}\left( n_{\text{probe}} \cdot \frac{N}{n_{\text{list}}} \cdot m \right)$, 
where $N$ is the database size and $m$ is the number of product quantization subspaces~\cite{jegou2010product}. This formulation allows efficient Top-$K$ retrieval even for large-scale databases. With parallel GPU execution~\cite{douze2025}, the retrieval latency remains at the millisecond level when $N \leq 10^6$.

To account for varying relevance among retrieved samples, we introduce an optimal transport (OT) weighting scheme~\cite{montesuma2024recent}. The transport cost and normalized weights are defined as
\begin{equation}
c_k = 1 - \left\langle \tilde{\mathbf{z}}_{q}, \tilde{\mathbf{z}}_{S,k} \right\rangle, \quad 
w_k = \frac{\exp(-c_k / \varepsilon)}{\sum_{j=1}^{K} \exp(-c_j / \varepsilon) + \epsilon},
\label{eq:otweight}
\end{equation}
where $\varepsilon > 0$ controls smoothness and $\epsilon$ ensures numerical stability. This formulation uses OT distances to evaluate similarity in the underlying feature distribution space~\cite{Cuturi2013ot}, yielding more reliable weights for feature aggregation. The final augmented representation is computed as
\begin{equation}
\mathbf{z}_{\text{aug}} = \sum_{k=1}^{K} w_k \mathbf{z}_{S,k} \oplus \left( \mathbf{z}_{q} - \sum_{k=1}^{K} w_k \mathbf{z}_{S,k} \right),
\end{equation}
which preserves both retrieved knowledge and query-specific information in a unified form.

% To account for varying relevance among retrieved samples, we introduce an optimal transport-based weighting scheme~\cite{Cuturi2013ot,montesuma2024recent}. The transport cost and normalized weights are defined as
% \begin{equation}
% c_k = 1 - \left\langle \tilde{\mathbf{z}}_{q}, \tilde{\mathbf{z}}_{S,k} \right\rangle, \quad 
% w_k = \frac{\exp(-c_k / \varepsilon)}{\sum_{j=1}^{K} \exp(-c_j / \varepsilon) + \epsilon},
% \label{eq:otweight}
% \end{equation}
% where $\varepsilon > 0$ controls smoothness and $\epsilon$ ensures numerical stability. 
% This formulation leverages OT-inspired distances to evaluate similarity in the underlying feature distribution space, yielding more reliable weights for feature aggregation.

% \paragraph{Optimal transport aggregation.} pass.

% \textbf{Retrieval Efficiency.}
% The computational bottleneck of the retrieval module lies in approximate nearest neighbor search. Based on the FAISS IVFPQ index, only \( n_{\text{probe}} \ll n_{\text{list}} \) Voronoi cells are visited during querying. The time complexity for a single query is:
% \[
% \mathcal{O}\left( n_{\text{probe}} \cdot \frac{N}{n_{\text{list}}} \cdot \frac{d}{m} \right),
% \]
% where \( N \) is the database size, \( d \) is the embedding dimension, and \( m \) is the number of PQ subspaces. The entire process is executed on GPU in parallel, achieving millisecond-level latency for Top-\(K\) retrieval when \( N \leq 10^6 \). Notably, the retrieval process is gradient-free and fully decoupled from backpropagation.

\subsection{Knowledge enhancement}\label{sec44}

This module integrates the implicit parametric embedding (class token) from pretrained TSFMs with the explicit numerical knowledge produced by the query encoder, which consists of the knowledge extraction and retrieval modules. The fused embedding is then used for supervised classification in the target domain, enabling joint transfer of parametric and retrieved numerical knowledge.

% \textbf{Knowledge Fusion.}
% Let $\mathbf{z}_c \in \mathbb{R}^d$ denote the class token from pretrained TSFMs, $\mathbf{z}_{q} \in \mathbb{R}^d$ the query embedding produced by the query encoder, and $\mathbf{z}_{\text{aug}} \in \mathbb{R}^d$ the retrieval-augmented representation. We perform residual fusion as
% \begin{equation}
% \mathbf{z}_{\text{fused}} = \text{LayerNorm}\left(
% \mathbf{z}_c + \mathbf{z}_{q} + \mathbf{z}_{\text{aug}}
% \right).
% \end{equation}
% Here, the inclusion of $\mathbf{z}_{q}$ not only provides task-specific semantic guidance but also induces supervised signal for the lightweight projection layer in the knowledge extraction module. In particular, by leveraging ground-truth labels from the target domain, the fusion process encourages the projection function in Eq.~(\ref{eq:f_z_ke}) to adaptively shape the embedding space, enabling the query-conditioned extraction of more discriminative patterns from the knowledge base.

\textbf{Knowledge fusion.}
Let $\mathbf{z}_c \in \mathbb{R}^d$ denote the class token from pretrained TSFMs. Given the query embedding $\mathbf{z}_{q}$ and the retrieved representation $\mathbf{z}_{\text{aug}}$ produced by the query encoder, we perform gated residual fusion:
\begin{equation}
\mathbf{g}_q = \sigma(\mathbf{W}_q \mathbf{z}_c), \quad
\mathbf{g}_r = \sigma(\mathbf{W}_r \mathbf{z}_c),
\end{equation}
\begin{equation}
\mathbf{z}_{\text{fused}} = \text{LayerNorm}\left(
\mathbf{z}_c + \mathbf{g}_q \odot \mathbf{z}_{q} + \mathbf{g}_r \odot \mathbf{z}_{\text{aug}}
\right),
\end{equation}
where $\sigma(\cdot)$ is the sigmoid function, $\mathbf{W}_q, \mathbf{W}_r$ are learnable matrices, and $\odot$ denotes element-wise multiplication. This formulation adaptively integrates parametric and retrieval-based information while maintaining class-related feature fusion. The inclusion of $\mathbf{z}_{q}$ further injects task-specific supervision into the projection learning in Eq.~(\ref{eq:f_z_ke}), improving the alignment between query-conditioned representations and discriminative patterns in the knowledge base.

\textbf{Supervised fine-tuning.}
The fused representation $\mathbf{z}_{\text{fused}}$ is directly fed into the lightweight classifier $H(\cdot)$, and the model is optimized using the cross-entropy objective:
\begin{equation}
p_i = H(\mathbf{z}_{\text{fused}, i}), \quad
\mathcal{L}_{\text{CE}} =
- \frac{1}{|\mathcal{D}_t|}
\sum_{(H(\mathbf{z}_{\text{fused}}), y_i) \in \mathcal{D}_t}
\log p_{i, y_i},
\end{equation}
where $p_{i, y_i}$ denotes the predicted probability for the ground-truth class $y_i$. During fine-tuning, gradients are propagated through the pre-trained TSFM and the projection layers in the knowledge extraction module. This enables coordinated optimization of parametric representations and retrieval-based knowledge alignment for improved classification performance.

\section{Experiments}

We evaluate R2R on univariate time series classification across 128 UCR datasets~\cite{dau2019ucr} under a fully supervised fine-tuning setting. For the retrieval process in R2R, we construct a knowledge base from the pretraining datasets of NuTime~\cite{lin2024nutime}, Mantis~\cite{feofanov2025mantis}, and UniShape~\cite{liu2026unishape}, which collectively comprise 1.89M (million) samples drawn from the training sets of multi-domain datasets~\cite{dau2019ucr,bagnall2018uea,eldele2021time,zhang2022self}. From this pool, we derive a class-balanced subset of 100K (thousand) samples to form the final knowledge base for efficient retrieval. Following~\cite{liu2026unishape}, we also evaluate generalization on 30 additional univariate time series datasets~\cite{middlehurst2024bake} in a zero-shot feature extraction setting. Further details on datasets, baselines, and implementation settings are provided in Appendix~\ref{appendA}.

\subsection{Main results}

\begin{figure*}[t]
\centering

% --- 左侧表格部分 ---
\begin{minipage}[b][0.28\textheight][t]{0.42\linewidth} 
    \centering
    \captionsetup{type=table}
    % 表格标题位于最顶部
    \caption{Test accuracy results on 128 UCR datasets under the fully supervised setting.}
    \label{tab:main128ucr}
    \vspace{2pt} 

    \vfill % 自动填充，将表格推向下方，直到与右侧对齐

    \resizebox{0.85\linewidth}{!}{ % 适当调大比例以减少中间空隙
        \begin{tabular}{@{}c|l|ccc@{}}
        \toprule
         & Method & Avg. Rank & Win & P-value \\ 
        \midrule
        \multirow{5}{*}{\rotatebox[origin=c]{90}{Non-DL}} 
         & Rocket & 7.62 & 18 & \cellcolor{gray!20} 1.01E-13 \\
         & MiniRocket & 7.75 & 13 & \cellcolor{gray!20} 1.82E-13 \\
         & RDST & 7.38 & 20 & \cellcolor{gray!20} 1.12E-11 \\
         & MR-H & 6.27 & 22 & \cellcolor{gray!20} 2.04E-10 \\
         & HC-2 & 5.16 & 31 & \cellcolor{gray!20} 3.62E-07 \\
        \noalign{\vskip 1pt} \hline \noalign{\vskip 1pt}
        \multirow{6}{*}{\rotatebox[origin=c]{90}{S-DL}}
         & InceptionTime & 9.41 & 13 & \cellcolor{gray!20} 1.56E-15 \\
         & TS2Vec & 12.65 & 8 & \cellcolor{gray!20} 1.25E-18 \\
         & PatchTST & 18.19 & 6 & \cellcolor{gray!20} 1.20E-20 \\
         & TimesNet & 17.11 & 5 & \cellcolor{gray!20} 9.47E-21 \\
         & SoftShape & 9.20 & 14 & \cellcolor{gray!20} 2.03E-14 \\
         & MambaSL & 13.13 & 14 & \cellcolor{gray!20} 7.37E-18 \\
        \noalign{\vskip 1pt} \hline \noalign{\vskip 1pt}
        \multirow{7}{*}{\rotatebox[origin=c]{90}{TSFMs}}
         & GPT4TS & 17.06 & 4 & \cellcolor{gray!20} 9.47E-21 \\
         & MOMENT & 17.58 & 2 & \cellcolor{gray!20} 8.47E-21 \\
         & UniTS & 16.74 & 2 & \cellcolor{gray!20} 5.93E-21 \\
         & NuTime & 10.26 & 15 & \cellcolor{gray!20} 2.21E-15 \\
         & TS-RAG & 9.31 & 11 & \cellcolor{gray!20} 1.44E-16 \\
         & Mantis & 8.07 & 21 & \cellcolor{gray!20} 8.43E-12 \\
         & UniShape & 4.47 & 31 & \cellcolor{gray!20} 1.68E-04 \\
        \noalign{\vskip 1pt} \hline \noalign{\vskip 1pt}
        \multirow{3}{*}{\rotatebox[origin=c]{90}{Ours}}
         & R2R$_{\text{NuTime}}$ & 7.90 & 23 & \cellcolor{gray!20} 5.64E-12 \\
         & R2R$_{\text{Mantis}}$ & 6.18 & 29 & \cellcolor{gray!20} 2.31E-07 \\
         & R2R$_{\text{UniShape}}$ & \textbf{3.01} & \textbf{48} & - \\ 
        \bottomrule
        \end{tabular}
    }
\end{minipage}
\hspace{0.02\linewidth} % 手动设置固定间距，防止中间空隙过大
% --- 右侧图片部分 ---
\begin{minipage}[b][0.26\textheight][t]{0.54\linewidth} 
    \centering
    \vspace{0pt} 

    % 图片顶部与左侧 caption 对齐
    \includegraphics[width=\linewidth, height=0.35\textheight, keepaspectratio]{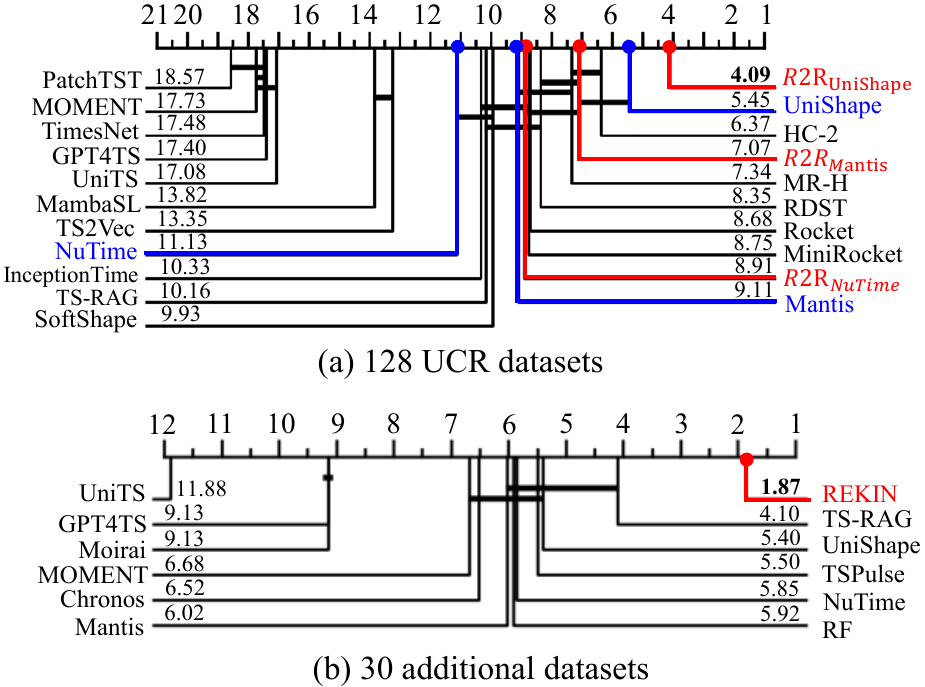}
    
    \vfill % 填充空间，确保图片标题沉到底部，与左侧表格底部对齐

    % 图片标题紧贴底部
    \caption{Critical difference diagram on 158 datasets.}
    \label{fig_2_cd_diagram}
\end{minipage}

\end{figure*}

This section evaluate R2R by integrating three advanced TSFMs~\cite{lin2024nutime,feofanov2025mantis,liu2026unishape}, denoted as R2R$_{\text{NuTime}}$, R2R$_{\text{Mantis}}$, and R2R$_{\text{UniShape}}$, and compare them with 18 baselines, including non-deep learning (Non-DL) methods (Rocket~\cite{dempster2020rocket}, Mini-Rocket~\cite{dempster2021minirocket}, RDST~\cite{guillaume2022random}, MultiRocket-Hydra (MR-H)~\cite{dempster2023hydra}, HIVE-COTE 2.0 (HC-2)~\cite{middlehurst2021hive}), supervised deep learning (S-DL) models (InceptionTime~\cite{ismail2020inceptiontime}, TS2Vec~\cite{yue2022ts2vec}, PatchTST~\cite{nietime2023}, TimesNet~\cite{wutimesnet2023}, SoftShape~\cite{liu2025learning}, MambaSL~\cite{jungmambasl2026}), and TSFM-based methods (GPT4TS~\cite{zhou2023one}, MOMENT~\cite{goswami2024moment}, UniTS~\cite{gao2024units}, NuTime~\cite{lin2024nutime}, TS-RAG~\cite{ningts2025}, Mantis~\cite{feofanov2025mantis}, UniShape~\cite{liu2026unishape}). We report average rank based on test classification accuracy and the number of datasets with the best accuracy (Win) on 128 UCR datasets to reduce bias from abnormally high or low accuracy results on individual datasets. All baselines and R2R variants are run under the same evaluation protocol: the official UCR train/test splits, the officially released hyperparameters of each method, and the same hardware budget, so the reported ranks are not confounded by extra tuning granted to R2R (see Appendix~\ref{appendA}).
In addition, we assess statistical significance using the Wilcoxon signed-rank test (p-value) and critical difference (CD) diagrams~\cite{demvsar2006statistical}. A p-value below 0.05 indicates significant pairwise differences, while CD diagrams provide a global comparison in which lower CD values denote better performance and methods not connected by a solid line differ significantly.

Table ~\ref{tab:main128ucr} reports the results of comparing the test classification accuracy among 21 methods (including three R2R variants) on 128 UCR datasets; detailed results are provided in Appendix~\ref{appendB}. R2R consistently improves all three integrated TSFMs, with R2R$_{\text{UniShape}}$ achieving the best overall performance, significantly outperforming all baselines as confirmed by the Wilcoxon $p$-values (all $< 0.05$) and CD diagram in Figure~\ref{fig_2_cd_diagram}a. Similar improvements are observed for R2R$_{\text{NuTime}}$ and R2R$_{\text{Mantis}}$, indicating that retrieval-based knowledge enhancement provides consistent gains beyond parameter-only transfer. Among baselines, Non-DL methods (e.g., Rocket family~\cite{dempster2023hydra}, HIVE-COTE 2.0~\cite{middlehurst2021hive}) remain strong competitors. 
TSFMs generally outperform S-DL methods due to large-scale pretraining, with Mantis and UniShape achieving the strongest baseline results, yet they rely solely on implicit parameter transfer. TS-RAG~\cite{ningts2025}, the only retrieval-augmented baseline, ranks 9.31 with 11 wins and thus performs worse than the classification-oriented TSFMs (i.e., Mantis and UniShape) and all three R2R variants, although it retrieves from the same knowledge base as R2R. This gap therefore reflects the classification-specific retrieval and fusion design rather than the retrieved corpus, and shows that R2R improves the post-training stage without introducing a new backbone or pretraining objective. In summary, the results demonstrate that retrieval-based knowledge effectively enhances parametric representations for TSC, enabling more positive transfer and improving TSFMs under a retrieval-to-recognition paradigm.

\subsection{Comparison results on zero-shot feature extraction}

\begin{table}[h]
\centering
\caption{Test accuracy results on 30 additional datasets under the zero-shot feature extraction setting.}
\label{tab:zero-shot-30-results}
\resizebox{\linewidth}{!}{
\begin{tabular}{@{}c|ccccccccccc|c@{}}
\toprule
Metric & RF & GPT4TS & MOMENT & Chronos & Moirai & UniTS & NuTime & Mantis & UniShape & TS-RAG & TSPulse & R2R \\ \midrule
Avg. Rank & 5.80 & 9.00 & 6.53 & 6.33 & 9.00 & 11.87 & 5.63 & 5.80 & 5.20 & 3.97 & 5.37 & \textbf{1.73} \\
Win & 2 & 1 & 1 & 2 & 1 & 0 & 0 & 1 & 1 & 5 & 1 & \textbf{21} \\
P-value 
& \cellcolor{gray!20} 8.52E-06 
& \cellcolor{gray!20} 1.28E-05 
& \cellcolor{gray!20} 1.43E-06 
& \cellcolor{gray!20} 1.30E-05 
& \cellcolor{gray!20} 9.31E-08 
& \cellcolor{gray!20} 2.05E-08 
& \cellcolor{gray!20} 2.11E-05 
& \cellcolor{gray!20} 2.64E-05 
& \cellcolor{gray!20} 8.20E-07 
& \cellcolor{gray!20} 1.33E-04 
& \cellcolor{gray!20} 9.31E-08 
& - \\ 
\bottomrule
\end{tabular}}
\end{table}

To further evaluate the effectiveness of retrieval-based numerical knowledge, we follow the setups of~\cite{feofanov2025mantis,liu2026unishape} and assess R2R under a zero-shot feature extraction setting. Specifically, both the default training and test sets are passed through parameter-frozen TSFMs to obtain embeddings, which are then used to train and evaluate a Random Forest classifier, with no model parameter fine-tuning. We conduct experiments on 30 additional datasets and compare against 11 baselines, including a Random Forest (RF)~\cite{breiman2001random} classifier on raw time series inputs and 10 TSFMs (GPT4TS~\cite{zhou2023one}, MOMENT~\cite{goswami2024moment}, Chronos~\cite{ansari2024chronos}, Moirai~\cite{woo2024unified}, UniTS~\cite{gao2024units}, NuTime~\cite{lin2024nutime}, Mantis~\cite{feofanov2025mantis}, UniShape~\cite{liu2026unishape}, TS-RAG~\cite{ningts2025}, and TSPulse~\cite{ekambaram2026tspulse}). R2R is built upon pre-trained UniShape in this setting.
Note that the same Random Forest serves both as the shared probing classifier over TSFM embeddings and, as the RF baseline, as a classifier fitted on the raw series, so Table~\ref{tab:zero-shot-30-results} reports one RF column (see Appendix~\ref{appendA}).

As shown in Table~\ref{tab:zero-shot-30-results} and Figure~\ref{fig_2_cd_diagram}b, R2R achieves the best performance and consistently yields statistically significant improvements over all baselines. 
The detailed results of Table~\ref{tab:zero-shot-30-results} are presented in Appendix~\ref {appendC}.
We observe that classification-oriented TSFMs (NuTime~\cite{lin2024nutime}, Mantis~\cite{feofanov2025mantis}, UniShape~\cite{liu2026unishape}) generally outperform forecasting-oriented models (Chronos~\cite{ansari2024chronos}, Moirai~\cite{woo2024unified}), indicating the importance of task alignment. While TS-RAG~\cite{ningts2025} improves Chronos via retrieval originally proposed for forecasting tasks, it remains inferior to R2R due to its lack of a classification-specific design. As the TS-RAG method is built on the same knowledge base as R2R in this setting (Appendix~\ref{appendC}), the gap in average rank (3.97 vs. 1.73) is attributable to the classification-specific retrieval target and fusion rather than to the retrieved corpus. On the whole, without any task-specific parameter fine-tuning, R2R uses retrieval-based feature-level knowledge for downstream time series classification, achieving substantially better performance than TSFMs that rely solely on parameter-based transfer.

\subsection{Ablation studies}

\begin{table*}[h]
\centering
\begin{minipage}[t]{0.58\linewidth}
\centering
\caption{Test accuracy results of ablation study on 128 UCR time series datasets.}
\vspace{4pt}
\label{tab:abla-ucr128}
\resizebox{\linewidth}{!}{
\begin{tabular}{@{}c|c|c|ccc@{}}
\toprule
\multicolumn{2}{c|}{Method} & Backbone & Avg. Rank & Win & P-value \\ 
\midrule
\multicolumn{2}{c|}{R2R$_{\text{UniShape}}$} & - & \textbf{1.48} & \textbf{86} & - \\ 
\hline
\multicolumn{2}{c|}{w/o Feature transfer} & - & 2.45 & 48 & \cellcolor{gray!20} 1.68E-04 \\ 
\hline
\multirow{6}{*}{\rotatebox[origin=c]{90}{\shortstack{w/o Parameter\\transfer}}}
 & with Retrieval & MiniRocket & 4.14 & 13 & \cellcolor{gray!20} 3.60E-18 \\
 & w/o Retrieval & MiniRocket & 5.09 & 11 & \cellcolor{gray!20} 1.58E-18 \\
 & w/o ReMGP & MiniRocket & 5.59 & 8 & \cellcolor{gray!20} 4.56E-19 \\
 & w/o Retrieval & Rocket & 5.91 & 8 & \cellcolor{gray!20} 9.56E-19 \\
 & w/o Retrieval & MultiRocket & 4.06 & 18 & \cellcolor{gray!20} 4.69E-17 \\
 & w/o Retrieval & UniShape & 4.52 & 13 & \cellcolor{gray!20} 5.58E-19 \\ 
\bottomrule
\end{tabular}
}
\end{minipage}
\hfill
\begin{minipage}[t]{0.39\linewidth}
\centering
\caption{Test accuracy results on 18 UCR datasets under varying knowledge base sizes and sampling strategies.}
\vspace{4pt}
\label{tab:kd-18ucr}
\resizebox{\linewidth}{!}{
\begin{tabular}{@{}cc|ccc@{}}
\toprule
Size & Sampling & Avg. Rank & Win & Time (s) \\ 
\midrule
50K & Balanced & 4.11 & 3 & 88.01 \\
100K & Balanced & \textbf{2.33} & 8 & 92.11 \\
200K & Balanced & 3.28 & \textbf{9} & 119.79 \\
500K & Balanced & 3.11 & 6 & 207.49 \\
1M & Balanced & 3.17 & 6 & 361.77 \\
1.89M & Balanced & 4.17 & 5 & 414.80 \\ \hline
100K & Random & 3.17 & 5 & 92.11 \\
100K & Unbalanced & 6.50 & 2 & 92.11 \\ 
\bottomrule
\end{tabular}
}
\end{minipage}
\end{table*}

This section presents ablation studies to analyze each component of R2R under two settings: (i) \textbf{w/o Feature transfer}, where the query encoder is removed and only the pretrained TSFM is fine-tuned; and (ii)  \textbf{w/o Parameter transfer}, where the pretrained TSFM is removed and classification relies solely on the query encoder, with further ablations on knowledge retrieval and gradient-free backbone selection. As shown in Table~\ref{tab:abla-ucr128}, removing feature transfer significantly degrades performance, indicating the importance of retrieval-based numerical knowledge. Under \textbf{w/o Parameter transfer}, using MiniRocket~\cite{dempster2021minirocket} as the backbone, the knowledge retrieval module (\textit{with Retrieval}) improves TSC results (e.g., Avg. Rank: 4.14 vs. 5.09), while removing the retrieval module (w/o $\mathbf{z}_{\text{aug}}$) and multi-scale global and positional patterns in the knowledge extraction module (\textit{w/o ReMGP}) further reduces performance, showing its role in enhancing embedding discriminability. We also compare gradient-free backbones for the query encoder, where MultiRocket~\cite{tan2022multirocket} performs best but shows limited gains over parameter-frozen UniShape~\cite{liu2026unishape}, while outperforming Rocket~\cite{dempster2020rocket}. 

Overall, all \textbf{w/o Parameter transfer} variants perform substantially worse than full R2R$_{\text{UniShape}}$ and even underperform \textbf{w/o Feature transfer}, suggesting that pretrained TSFM parameter transfer via supervised fine-tuning is the main knowledge source, while retrieval-based feature transfer provides further improvements. These variants also indicate whether the improvement is inherited from the strong Rocket-family extractors used by the query encoder: Rocket and MiniRocket rank only 7.62 and 7.75 in Table~\ref{tab:main128ucr}, well behind parameter-only UniShape (4.47) and R2R$_{\text{UniShape}}$ (3.01), and here a parameter-frozen UniShape query encoder (4.52) is comparable to MultiRocket (4.06), so R2R is not tied to the Rocket family. Since every \textbf{w/o Parameter transfer} variant stays far behind full R2R$_{\text{UniShape}}$ (1.48 with 86 wins), the gain originates from jointly fine-tuning retrieved numerical knowledge with pretrained TSFM parameters rather than from a stronger feature extractor.

Table~\ref{tab:kd-18ucr} studies the impact of the knowledge base in R2R using 18 representative UCR datasets following~\cite{liu2026unishape} for efficient analysis under large-scale retrieval. A class-balanced subset of 100K samples from the pretraining dataset achieves the best trade-off between accuracy and efficiency. Here, \textit{Time (s)}  denotes the additional average time for knowledge extraction and retrieval with the R2R query encoder across the 18 datasets. Increasing the knowledge base from 50K to 100K introduces only minor overhead (about 4 seconds per dataset), while scaling to 1.89M samples raises the time by about 4.5×. Yet, the time cost remains manageable (within 7 minutes per dataset). Moreover, Table~\ref{tab:kd-18ucr} indicates that both class diversity and distribution balance are critical for effective numerical knowledge transfer. Detailed results of Tables~\ref{tab:abla-ucr128} and~\ref{tab:kd-18ucr} are provided in Appendix~\ref{appendD}.

\subsection{Knowledge-enhanced analysis}

\begin{figure}[t]
\centering
\includegraphics[width=5.5in]{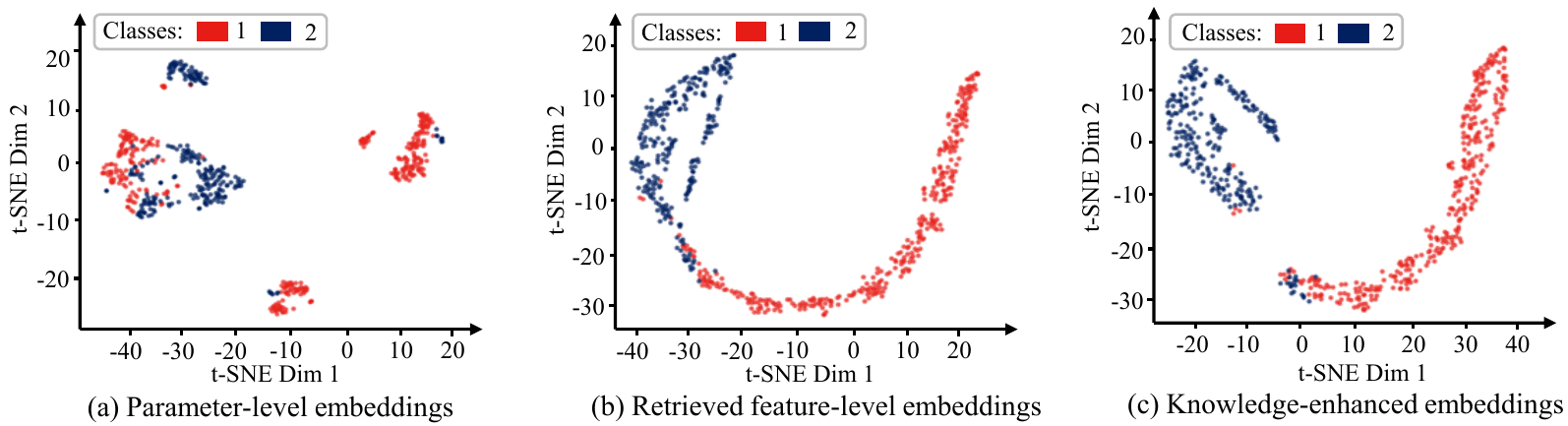}
\caption{t-SNE visualization on the \textit{SonyAIBORobotSurface1} dataset.
}
\label{fig_3_tsne}
\end{figure}

% \usepackage{booktabs}
% \usepackage[normalem]{ulem}
\begin{table}[h]
\centering
\caption{Test results on 18 UCR datasets compared with different data augmentation methods.}
\label{tab:kd-aug-ana}
\resizebox{\textwidth}{!}{% % 开始缩放
\begin{tabular}{@{}c|cccccccc|c@{}}
\toprule
Metric & Raw & Jitter & Time warp & Magnitude warp & Window slice & Time flip & Permutation & Scaling & R2R \\ \midrule
Avg. Rank & 2.39 & 5.11 & 5.06 & 4.89 & 4.33 & 5.06 & 5.50 & 4.83 & \textbf{2.00} \\
\bottomrule
\end{tabular}}
\end{table}

To analyze how R2R utilizes retrieved embeddings for knowledge enhancement, we visualize target sample representations using the t-SNE~\cite{van2008visualizing} method. Figure~\ref{fig_3_tsne}a shows embeddings from the pretrained UniShape based on parameter fine-tuning, Figure~\ref{fig_3_tsne}b shows embeddings from the retrieval module in R2R, and Figure~\ref{fig_3_tsne}c shows the fused embeddings combining parameter- and feature-level knowledge for final TSC. In Figure~\ref{fig_3_tsne}a, classes 1 and 2 are partially mixed, whereas Figures~\ref{fig_3_tsne}b and~\ref{fig_3_tsne}c exhibit clearer separation. These results indicate that R2R retrieves class-discriminative features from the pretraining knowledge base, improving inter-class separability. Additional visualizations in Appendix~\ref{appendE} further confirm consistent gains across different datasets and multi-class settings.

To verify the effectiveness of R2R for retrieval-based numerical knowledge enhancement, Table~\ref{tab:kd-aug-ana} compares 7 representative time series data augmentation methods~\cite{wen2021time,iglesias2023data}. As shown, these methods often degrade performance relative to the Raw setting (fine-tuning without augmentation), indicating negative transfer on many datasets. 
In contrast, R2R retrieves class-relevant embeddings from a knowledge base to help pretrained TSFMs recall pretraining knowledge for augmentation, thereby improving positive transfer in TSC.
This advantage likely stems from reduced randomness compared to conventional stochastic augmentations, which may disrupt intrinsic class structures in time series. 

% \usepackage{booktabs}
% \usepackage{multirow}
\begin{table}[h]
\centering
\caption{Impact of continual fine-tuning for TSC. Acc$_{A_1}$: Initial accuracy on dataset $A$; Acc$_{A_2}$: Accuracy on dataset $A$ after continual training on dataset $B$; Forgetting: Acc$_{A_1}$ - Acc$_{A_2}$.}
\label{tab:continue-ana1}
\resizebox{\textwidth}{!}{% % 开始缩放
\begin{tabular}{@{}ll|ccc|ccc@{}}
\toprule
\multicolumn{1}{c}{\multirow{2}{*}{Dataset   A}} & \multicolumn{1}{c|}{\multirow{2}{*}{Dataset B}} & \multicolumn{3}{c|}{Only parameter   transfer} & \multicolumn{3}{c}{R2R transfer} \\ \cmidrule(l){3-8} 
\multicolumn{1}{c}{} & \multicolumn{1}{c|}{} & \multicolumn{1}{c}{Acc$_{A_1}$} & \multicolumn{1}{c}{Acc$_{A_2}$} & \multicolumn{1}{c|}{Forgetting} & \multicolumn{1}{c}{Acc$_{A_1}$} & \multicolumn{1}{c}{Acc$_{A_2}$} & \multicolumn{1}{c}{Forgetting} \\ \cmidrule(r){1-8}
AsphaltRegularityUni\_eq & FloodModeling1\_disc & 0.9654 & 0.4887 & 0.4767 & 0.9734 & 0.7909 & \textbf{0.1825} \\
GesturePebbleZ1\_eq & Colposcopy & 0.9477 & 0.5233 & 0.4244 & 0.9535 & 0.7151 & \textbf{0.2384} \\
AllGestureWiimoteX\_eq & AllGestureWiimoteY\_eq & 0.7671 & 0.2429 & 0.5242 & 0.7786 & 0.2771 & \textbf{0.5015} \\
GestureMidAirD1\_eq & GestureMidAirD2\_eq & 0.7538 & 0.1231 & 0.6307 & 0.7692 & 0.1538 & \textbf{0.6154} \\ 
\bottomrule
\end{tabular}}
\end{table}

Finally, we select four datasets from the 30 additional datasets, covering different numbers of classes, to evaluate R2R under continual classification fine-tuning. Table~\ref{tab:continue-ana1} compares standard parameter-only fine-tuning with R2R transfer. R2R consistently outperforms parametric fine-tuning alone on Acc$_{A_1}$ and Acc$_{A_2}$, and exhibits less forgetting after continual fine-tuning, indicating that retrieved embeddings help mitigate pre-trained knowledge forgetting and improve transfer generalization across downstream tasks. 
Additional results for continual fine-tuning, together with data augmentation descriptions and detailed results in Table~\ref{tab:kd-aug-ana}, are presented in Appendix~\ref{appendE}. Appendix~\ref{appendF} reports hyperparameter analysis, and Appendix~\ref{appendG} reports a preliminary multivariate evaluation in which R2R$_{\text{UniShape}}$ is extended to ten UEA datasets~\cite{bagnall2018uea} through a trainable linear channel adapter. Appendix~\ref{appendH} and~\ref{appendI} 
provide limitations and broader impacts.

\section{Conclusion}

This paper presents R2R, a retrieval-to-recognition post-training framework that enhances pretrained TSFMs for time series classification. R2R employs a plug-and-play query encoder to retrieve class-relevant numerical knowledge and integrates it with parametric representations for downstream classification. Experiments demonstrate that R2R consistently improves pretrained TSFMs via joint parameter- and feature-level knowledge transfer, attaining the best average rank with 48 wins among 21 baselines on the 128 UCR datasets under supervised fine-tuning and 21 wins among 12 baselines on the 30 additional datasets under zero-shot feature extraction, with all gains being statistically significant ($p < 0.05$). However, R2R acts only at the post-training stage and contributes no new pretraining objective or backbone architecture, and it currently relies on a knowledge base built from univariate time series. A preliminary multivariate extension through a trainable linear channel adapter is reported in Appendix~\ref{appendG}, where R2R$_{\text{UniShape}}$ attains the best average rank on ten UEA datasets, although the retrieved knowledge itself is still univariate. Future work will explore more general retrieval module designs and build natively multivariate knowledge bases, so that inter-variable dependencies are captured by retrieval rather than by an input adapter.

\section*{Acknowledgments}
We thank the anonymous reviewers for their helpful feedbacks.
We thank Professor Eamonn Keogh and all the people who have contributed to the UCR archive and other time series classification datasets. 
This work was partially supported by the Singapore NRF AI4S
project (NRF-AI4SCH-2025-0001), the Ministry of Education,
Singapore, under its Academic Research Fund (AcRF) Tier 1
grant (RG16/25), the SMU-SUTD Internal Research Grant Call
(SMU-SUTD 2023\_02\_01), the SUTD Kickstarter Initiative
(SKI 2021\_06\_12), and AIMfg, A$^*$STAR, Singapore, under
its MTI WS Fund.
This work was also partially supported by the National Natural
Science Foundation of China (Grant Nos. 62272173, 62273109),
the Natural Science Foundation of Guangdong Province
(Grant Nos. 2024A1515010089, 2022A1515010179),
the Science and Technology Planning Project of Guangdong Province
(Grant No. 2023A0505050106), and the National Key R\&D
Program of China (Grant No. 2023YFA1011601).

\bibliographystyle{unsrt}
\bibliography{neurips26}

%%%%%%%%%%%%%%%%%%%%%%%%%%%%%%%%%%%%%%%%%%%%%%%%%%%%%%%%%%%%

\newpage
\appendix

\section{Experiment setting}\label{appendA}

\subsection{Datasets and knowledge base}\label{appendA-1}

This study employs the univariate UCR Time Series Archive~\cite{dau2019ucr} as the primary benchmark for classification, encompassing 128 datasets from a wide range of real-world domains. Detailed dataset statistics are reported in Table~\ref{tab:append-ucr_stats}. All 128 UCR time series datasets are utilized for fully supervised classification fine-tuning evaluation.

For the retrieval module in R2R, the knowledge base is constructed from the pretraining dataset adopted in prior studies~\cite{lin2024nutime,feofanov2025mantis,liu2026unishape}. This pretraining dataset integrates the default training sets from the UCR Time Series Archive~\cite{dau2019ucr}, the UEA Time Series Archive~\cite{bagnall2018uea}, and eight other domain-specific time series datasets~\cite{eldele2021time,zhang2022self}, resulting in approximately 1.89 million univariate samples. Given the pronounced class imbalance in this aggregated dataset, directly incorporating all samples would incur substantial computational overhead and may degrade retrieval quality. To alleviate this issue, we construct a class-balanced knowledge-base sampling strategy that ensures a minimum of 50 and a maximum of 1,000 samples per class. This procedure yields a balanced subset of approximately 100,000 (100K) samples, preserving class diversity while preventing over-representation of dominant classes and reducing retrieval cost.

Following~\cite{liu2026unishape}, we further assess zero-shot feature extraction on 30 additional univariate time series datasets~\cite{middlehurst2024bake}. This evaluation protocol examines the capability of R2R to enhance time series classification performance without gradient-based optimization. Detailed statistics of these 30 additional time series datasets are summarized in Table~\ref{tab:append-zeroshot_stats}.

\begin{table}[h]
\caption{
Summary of the 128 univariate datasets from the UCR archive used in the fully supervised classification setting. ``Train size'' and ``Test size'' indicate the number of instances in the corresponding splits. ``No. class'' represents the number of classes, and ``Length'' denotes the time series length. The label ``Vary'' in the length column signifies datasets with variable-length sequences due to missing values.
}
\label{tab:append-ucr_stats}
\resizebox{\linewidth}{!}{
\begin{tabular}{clcccc|clcccc}
\hline
\textbf{ID} & \textbf{Dataset name} & \textbf{Train size} & \textbf{Test size} & \textbf{No. classes} & \textbf{Length} & \textbf{ID} & \textbf{Dataset name} & \textbf{Train size} & \textbf{Test size} & \textbf{No. classes} & \textbf{Length} \\ \hline
1 & ACSF1 & 100 & 100 & 10 & 1460 & 65 & ItalyPowerDemand & 67 & 1029 & 2 & 24 \\
2 & Adiac & 390 & 391 & 37 & 176 & 66 & LargeKitchenAppliances & 375 & 375 & 3 & 720 \\
3 & AllGestureWiimoteX & 300 & 700 & 10 & Vary & 67 & Lightning2 & 60 & 61 & 2 & 637 \\
4 & AllGestureWiimoteY & 300 & 700 & 10 & Vary & 68 & Lightning7 & 70 & 73 & 7 & 319 \\
5 & AllGestureWiimoteZ & 300 & 700 & 10 & Vary & 69 & Mallat & 55 & 2345 & 8 & 1024 \\
6 & ArrowHead & 36 & 175 & 3 & 251 & 70 & Meat & 60 & 60 & 3 & 448 \\
7 & Beef & 30 & 30 & 5 & 470 & 71 & MedicalImages & 381 & 760 & 10 & 99 \\
8 & BeetleFly & 20 & 20 & 2 & 512 & 72 & MelbournePedestrian & 1194 & 2439 & 10 & 24 \\
9 & BirdChicken & 20 & 20 & 2 & 512 & 73 & MiddlePhalanxOutlineAgeGroup & 400 & 154 & 3 & 80 \\
10 & BME & 30 & 150 & 3 & 128 & 74 & MiddlePhalanxOutlineCorrect & 600 & 291 & 2 & 80 \\
11 & Car & 60 & 60 & 4 & 577 & 75 & MiddlePhalanxTW & 399 & 154 & 6 & 80 \\
12 & CBF & 30 & 900 & 3 & 128 & 76 & MixedShapesRegularTrain & 500 & 2425 & 5 & 1024 \\
13 & Chinatown & 20 & 343 & 2 & 24 & 77 & MixedShapesSmallTrain & 100 & 2425 & 5 & 1024 \\
14 & ChlorineConcentration & 467 & 3840 & 3 & 166 & 78 & MoteStrain & 20 & 1252 & 2 & 84 \\
15 & CinCECGTorso & 40 & 1380 & 4 & 1639 & 79 & NonInvasiveFetalECGThorax1 & 1800 & 1965 & 42 & 750 \\
16 & Coffee & 28 & 28 & 2 & 286 & 80 & NonInvasiveFetalECGThorax2 & 1800 & 1965 & 42 & 750 \\
17 & Computers & 250 & 250 & 2 & 720 & 81 & OliveOil & 30 & 30 & 4 & 570 \\
18 & CricketX & 390 & 390 & 12 & 300 & 82 & OSULeaf & 200 & 242 & 6 & 427 \\
19 & CricketY & 390 & 390 & 12 & 300 & 83 & PhalangesOutlinesCorrect & 1800 & 858 & 2 & 80 \\
20 & CricketZ & 390 & 390 & 12 & 300 & 84 & Phoneme & 214 & 1896 & 39 & 1024 \\
21 & Crop & 7200 & 16800 & 24 & 46 & 85 & PickupGestureWiimoteZ & 50 & 50 & 10 & Vary \\
22 & DiatomSizeReduction & 16 & 306 & 4 & 345 & 86 & PigAirwayPressure & 104 & 208 & 52 & 2000 \\
23 & DistalPhalanxOutlineAgeGroup & 400 & 139 & 3 & 80 & 87 & PigArtPressure & 104 & 208 & 52 & 2000 \\
24 & DistalPhalanxOutlineCorrect & 600 & 276 & 2 & 80 & 88 & PigCVP & 104 & 208 & 52 & 2000 \\
25 & DistalPhalanxTW & 400 & 139 & 6 & 80 & 89 & PLAID & 537 & 537 & 11 & Vary \\
26 & DodgerLoopDay & 78 & 80 & 7 & 288 & 90 & Plane & 105 & 105 & 7 & 144 \\
27 & DodgerLoopGame & 20 & 138 & 2 & 288 & 91 & PowerCons & 180 & 180 & 2 & 144 \\
28 & DodgerLoopWeekend & 20 & 138 & 2 & 288 & 92 & ProximalPhalanxOutlineAgeGroup & 400 & 205 & 3 & 80 \\
29 & Earthquakes & 322 & 139 & 2 & 512 & 93 & ProximalPhalanxOutlineCorrect & 600 & 291 & 2 & 80 \\
30 & ECG200 & 100 & 100 & 2 & 96 & 94 & ProximalPhalanxTW & 400 & 205 & 6 & 80 \\
31 & ECG5000 & 500 & 4500 & 5 & 140 & 95 & RefrigerationDevices & 375 & 375 & 3 & 720 \\
32 & ECGFiveDays & 23 & 861 & 2 & 136 & 96 & Rock & 20 & 50 & 4 & 2844 \\
33 & ElectricDevices & 8926 & 7711 & 7 & 96 & 97 & ScreenType & 375 & 375 & 3 & 720 \\
34 & EOGHorizontalSignal & 362 & 362 & 12 & 1250 & 98 & SemgHandGenderCh2 & 300 & 600 & 2 & 1500 \\
35 & EOGVerticalSignal & 362 & 362 & 12 & 1250 & 99 & SemgHandMovementCh2 & 450 & 450 & 6 & 1500 \\
36 & EthanolLevel & 504 & 500 & 4 & 1751 & 100 & SemgHandSubjectCh2 & 450 & 450 & 5 & 1500 \\
37 & FaceAll & 560 & 1690 & 14 & 131 & 101 & ShakeGestureWiimoteZ & 50 & 50 & 10 & Vary \\
38 & FaceFour & 24 & 88 & 4 & 350 & 102 & ShapeletSim & 20 & 180 & 2 & 500 \\
39 & FacesUCR & 200 & 2050 & 14 & 131 & 103 & ShapesAll & 600 & 600 & 60 & 512 \\
40 & FiftyWords & 450 & 455 & 50 & 270 & 104 & SmallKitchenAppliances & 375 & 375 & 3 & 720 \\
41 & Fish & 175 & 175 & 7 & 463 & 105 & SmoothSubspace & 150 & 150 & 3 & 15 \\
42 & FordA & 3601 & 1320 & 2 & 500 & 106 & SonyAIBORobotSurface1 & 20 & 601 & 2 & 70 \\
43 & FordB & 3636 & 810 & 2 & 500 & 107 & SonyAIBORobotSurface2 & 27 & 953 & 2 & 65 \\
44 & FreezerRegularTrain & 150 & 2850 & 2 & 301 & 108 & StarLightCurves & 1000 & 8236 & 3 & 1024 \\
45 & FreezerSmallTrain & 28 & 2850 & 2 & 301 & 109 & Strawberry & 613 & 370 & 2 & 235 \\
46 & Fungi & 18 & 186 & 18 & 201 & 110 & SwedishLeaf & 500 & 625 & 15 & 128 \\
47 & GestureMidAirD1 & 208 & 130 & 26 & Vary & 111 & Symbols & 25 & 995 & 6 & 398 \\
48 & GestureMidAirD2 & 208 & 130 & 26 & Vary & 112 & SyntheticControl & 300 & 300 & 6 & 60 \\
49 & GestureMidAirD3 & 208 & 130 & 26 & Vary & 113 & ToeSegmentation1 & 40 & 228 & 2 & 277 \\
50 & GesturePebbleZ1 & 132 & 172 & 6 & Vary & 114 & ToeSegmentation2 & 36 & 130 & 2 & 343 \\
51 & GesturePebbleZ2 & 146 & 158 & 6 & Vary & 115 & Trace & 100 & 100 & 4 & 275 \\
52 & GunPoint & 50 & 150 & 2 & 150 & 116 & TwoLeadECG & 23 & 1139 & 2 & 82 \\
53 & GunPointAgeSpan & 135 & 316 & 2 & 150 & 117 & TwoPatterns & 1000 & 4000 & 4 & 128 \\
54 & GunPointMaleVersusFemale & 135 & 316 & 2 & 150 & 118 & UMD & 36 & 144 & 3 & 150 \\
55 & GunPointOldVersusYoung & 136 & 315 & 2 & 150 & 119 & UWaveGestureLibraryAll & 896 & 3582 & 8 & 945 \\
56 & Ham & 109 & 105 & 2 & 431 & 120 & UWaveGestureLibraryX & 896 & 3582 & 8 & 315 \\
57 & HandOutlines & 1000 & 370 & 2 & 2709 & 121 & UWaveGestureLibraryY & 896 & 3582 & 8 & 315 \\
58 & Haptics & 155 & 308 & 5 & 1092 & 122 & UWaveGestureLibraryZ & 896 & 3582 & 8 & 315 \\
59 & Herring & 64 & 64 & 2 & 512 & 123 & Wafer & 1000 & 6164 & 2 & 152 \\
60 & HouseTwenty & 40 & 119 & 2 & 2000 & 124 & Wine & 57 & 54 & 2 & 234 \\
61 & InlineSkate & 100 & 550 & 7 & 1882 & 125 & WordSynonyms & 267 & 638 & 25 & 270 \\
62 & InsectEPGRegularTrain & 62 & 249 & 3 & 601 & 126 & Worms & 181 & 77 & 5 & 900 \\
63 & InsectEPGSmallTrain & 17 & 249 & 3 & 601 & 127 & WormsTwoClass & 181 & 77 & 2 & 900 \\
64 & InsectWingbeatSound & 220 & 1980 & 11 & 256 & 128 & Yoga & 300 & 3000 & 2 & 426 \\ \hline
\end{tabular}
}
\end{table}

\begin{table}[t]
\centering
\caption{
Summary of the 30 additional univariate time series datasets used for zero-shot feature extraction evaluation. Dataset name suffixes indicate preprocessing protocols: \texttt{\_eq} denotes unequal-length sequences normalized via mean-padding with Gaussian noise; \texttt{\_nmv} refers to datasets originally containing missing values, where only complete instances are retained; and \texttt{\_disc} indicates datasets derived from the TSER archive. ``Train size'' and ``Test size'' correspond to the number of samples in each split. ``No. class'' specifies the number of classes, ``Length'' denotes the sequence length, and ``Category'' indicates the application domain.}
\label{tab:append-zeroshot_stats}
\footnotesize % <--- 在这里添加，作用于整个表格（包括 tabular 内容）
% \resizebox{\linewidth}{!}{
\begin{tabular}{clcccccl}
\toprule
\textbf{ID} & \textbf{Dataset name} & \textbf{Train size} & \textbf{Test size} & \textbf{Length} & \textbf{No. classes} & \textbf{Category} \\
\midrule
1 & AconityMINIPrinterLarge\_eq & 2403 & 1184 & 300 & 2 & Sensor \\
2 & AconityMINIPrinterSmall\_eq & 589 & 292 & 300 & 2 & Sensor \\
3 & AllGestureWiimoteX\_eq & 300 & 700 & 500 & 10 & Motion \\
4 & AllGestureWiimoteY\_eq & 300 & 700 & 500 & 10 & Motion \\
5 & AllGestureWiimoteZ\_eq & 300 & 700 & 500 & 10 & Motion \\
6 & AsphaltObstaclesUni\_eq & 390 & 391 & 736 & 4 & Sensor \\
7 & AsphaltPavementTypeUni\_eq & 1055 & 1056 & 2371 & 3 & Sensor \\
8 & AsphaltRegularityUni\_eq & 751 & 751 & 4201 & 2 & Sensor \\
9 & Colposcopy & 99 & 101 & 180 & 6 & Image \\
10 & Covid3Month\_disc & 140 & 140 & 84 & 3 & Other \\
11 & DodgerLoopDay\_nmv & 67 & 77 & 288 & 7 & Sensor \\
12 & DodgerLoopGame\_nmv & 17 & 127 & 288 & 2 & Sensor \\
13 & DodgerLoopWeekend\_nmv & 18 & 126 & 288 & 2 & Sensor \\
14 & ElectricDeviceDetection & 624 & 3768 & 256 & 2 & Image \\
15 & FloodModeling1\_disc & 471 & 471 & 266 & 2 & Simulated \\
16 & FloodModeling2\_disc & 466 & 466 & 266 & 2 & Simulated \\
17 & FloodModeling3\_disc & 429 & 429 & 266 & 2 & Simulated \\
18 & GestureMidAirD1\_eq & 208 & 130 & 360 & 26 & Motion \\
19 & GestureMidAirD2\_eq & 208 & 130 & 360 & 26 & Motion \\
20 & GestureMidAirD3\_eq & 208 & 130 & 360 & 26 & Motion \\
21 & GesturePebbleZ1\_eq & 132 & 172 & 455 & 6 & Motion \\
22 & GesturePebbleZ2\_eq & 146 & 158 & 455 & 6 & Motion \\
23 & KeplerLightCurves & 920 & 399 & 4767 & 7 & Sensor \\
24 & MelbournePedestrian\_nmv & 1138 & 2319 & 24 & 10 & Sensor \\
25 & PhoneHeartbeatSound & 424 & 182 & 3053 & 5 & Other \\
26 & PickupGestureWiimoteZ\_eq & 50 & 50 & 361 & 10 & Motion \\
27 & PLAID\_eq & 537 & 537 & 1345 & 11 & Device \\
28 & ShakeGestureWiimoteZ\_eq & 50 & 50 & 385 & 10 & Motion \\
29 & SharePriceIncrease & 965 & 966 & 60 & 2 & Other \\
30 & Tools & 310 & 134 & 2926 & 5 & Other \\
\bottomrule
\end{tabular}
\end{table}

\subsection{Baselines}

\paragraph{Supervised classification on 128 UCR datasets.}
We consider 18 baselines spanning non-deep learning methods, end-to-end supervised deep models, and pre-trained time series foundation models.

\begin{itemize}

\item \textbf{Rocket}\footnote{\url{https://github.com/angus924/rocket}}:
Dempster et al.~\cite{dempster2020rocket} propose Rocket, which applies a large collection of random convolutional kernels with random initialization parameters to extract features, followed by a ridge classifier for efficient time series classification.

\item \textbf{MiniRocket}\footnote{\url{https://github.com/angus924/minirocket}}:
Dempster et al.~\cite{dempster2021minirocket} introduce MiniRocket, a deterministic variant of Rocket with fixed initialization parameters that significantly improves computational efficiency while maintaining similar or higher classification performance.

\item \textbf{RDST}\footnote{\url{https://github.com/baraline/convst}}:
Guillaume et al.~\cite{guillaume2022random} present RDST, which extends shapelet-based methods by incorporating multi-scale dilation into randomly sampled subsequences for time series classification.

\item \textbf{MultiROCKET-Hydra (MR-H)}\footnote{\url{https://github.com/angus924/hydra}}:
Dempster et al.~\cite{dempster2023hydra} propose Hydra, a hybrid approach that combines convolutional kernel transforms with dictionary-based representations using competing kernels.

\item \textbf{Hive-COTE 2.0 (HC-2)}\footnote{\url{https://github.com/sktime/sktime/blob/main/sktime/classification/hybrid/_hivecote_v2.py}}:
Middlehurst et al.~\cite{middlehurst2021hive} develop Hive-COTE 2.0, an ensemble framework integrating classifiers from multiple representations, including shapelets, intervals, and dictionary-based features.

\item \textbf{InceptionTime}\footnote{\url{https://github.com/timeseriesAI/tsai/blob/main/tsai/models/InceptionTime.py}}:
Ismail Fawaz et al.~\cite{ismail2020inceptiontime} propose InceptionTime, an ensemble of deep CNNs with Inception-style modules to capture multi-scale temporal dependencies for classification.

\item \textbf{TS2Vec}\footnote{\url{https://github.com/zhihanyue/ts2vec}}:
Yue et al.~\cite{yue2022ts2vec} introduce TS2Vec, which learns robust timestamp-level representations through hierarchical contrastive learning over augmented context views.

\item \textbf{PatchTST}\footnote{\url{https://github.com/yuqinie98/patchtst}}:
Nie et al.~\cite{nietime2023} propose PatchTST, which segments time series into patches as Transformer tokens and adopts channel-independent modeling for representation learning.

\item \textbf{TimesNet}\footnote{\url{https://github.com/thuml/TimesNet}}:
Wu et al.~\cite{wutimesnet2023} introduce TimesNet, which transforms 1D time series into 2D space based on periodic patterns for general time series modeling.

\item \textbf{SoftShape}\footnote{\url{https://github.com/qianlima-lab/SoftShape}}:
Liu et al.~\cite{liu2025learning} propose SoftShape, which replaces hard shapelet filtering with learnable soft shapelets for flexible subsequence selection and classification.

\item \textbf{MambaSL}\footnote{\url{https://github.com/yoom618/MambaSL}}:
Jung et al.~\cite{jungmambasl2026} introduce MambaSL, which adapts selective state space models with lightweight single-layer Mamba for time series classification.

\item \textbf{GPT4TS}\footnote{\url{https://github.com/DAMO-DI-ML/One_Fits_All}}:
Zhou et al.~\cite{zhou2023one} propose GPT4TS, a framework that leverages pre-trained language models with patch-based tokenization for time series modeling.

\item \textbf{MOMENT}\footnote{\url{https://github.com/moment-timeseries-foundation-model/moment}}:
Goswami et al.~\cite{goswami2024moment} introduce MOMENT, a family of Transformer models pre-trained via masked time series prediction on large-scale time series datasets.

\item \textbf{UniTS}\footnote{\url{https://github.com/mims-harvard/UniTS}}:
Gao et al.~\cite{gao2024units} propose UniTS, a unified Transformer framework that integrates multiple time series tasks through task tokenization.

\item \textbf{NuTime}\footnote{\url{https://github.com/chenguolin/NuTime}}:
Lin et al.~\cite{lin2024nutime} introduce NuTime, which employs multi-scale numerical embeddings to support effective pre-training on diverse time series data.

\item \textbf{TS-RAG}\footnote{\url{https://github.com/UConn-DSIS/TS-RAG}}:
Ning et al.~\cite{ningts2025} propose TS-RAG, which augments a pre-trained time series foundation model with patterns retrieved from an external knowledge base through an adaptive retrieval mixer. As TS-RAG is originally designed for zero-shot forecasting, we adapt it to supervised classification by attaching a classification head to its retrieval-augmented representations and fine-tuning it under the same protocol as the other TSFM baselines, with the retrieval knowledge base constructed in the same way as in R2R.

\item \textbf{Mantis}\footnote{\url{https://github.com/vfeofanov/mantis}}:
Feofanov et al.~\cite{feofanov2025mantis} propose Mantis, a ViT-based foundation model pre-trained via contrastive learning for time-series classification.

\item \textbf{UniShape}\footnote{\url{https://github.com/qianlima-lab/UniShape}}:
Liu et al.~\cite{liu2026unishape} introduce UniShape, a shape-aware foundation model that integrates adaptive subsequence selection and prototype-based pretraining.

\end{itemize}

In summary, among the 18 baselines, 14 methods (Rocket, MiniRocket, RDST, MR-H, InceptionTime, TS2Vec, PatchTST, TimesNet, SoftShape, GPT4TS, MOMENT, UniTS, NuTime, and Mantis) follow the same setup as UniShape~\cite{liu2026unishape}, while Hive-COTE 2.0~\cite{middlehurst2021hive}, MambaSL~\cite{jungmambasl2026}, and TS-RAG~\cite{ningts2025} are additionally introduced for comparison.

\paragraph{Zero-shot feature extraction on 30 additional datasets.}
We evaluate 11 baselines for zero-shot feature extraction, including a Random Forest~\cite{breiman2001random} classifier on raw inputs and 10 pre-trained TSFMs.

\begin{itemize}

\item \textbf{Random Forest}\footnote{\url{https://github.com/scikit-learn/scikit-learn/blob/fe2edb3cd/sklearn/ensemble/_forest.py}}:
A classical machine learning classifier without gradient-based optimization~\cite{breiman2001random} that directly operates on raw time series, rather than using representations obtained from pre-trained TSFMs. Two roles of the Random Forest should be distinguished, since a single classifier configuration is used throughout. As the \emph{probing classifier}, it is fitted on the frozen embeddings produced by each TSFM-based method and applied to the corresponding test embeddings; this is the shared evaluation protocol of the zero-shot setting and is not itself a competing method. As the \emph{RF baseline}, the same classifier is fitted on the raw series $\mathbf{x} \in \mathbb{R}^{L}$, i.e. the degenerate case in which no pretrained model is involved and the $L$ time steps of a series are passed directly as $L$ ordered input features. Table~\ref{tab:zero-shot-30-results} therefore contains a single RF column, and comparing it against the remaining columns isolates whether TSFM embeddings are more discriminative than the raw series under an identical classifier.

\item \textbf{Chronos}\footnote{\url{https://github.com/amazon-science/chronos-forecasting}}:
Ansari et al.~\cite{ansari2024chronos} propose Chronos for time series forecasting, a probabilistic Transformer that tokenizes time series into discrete vocabularies for large-scale pre-training.

\item \textbf{Moirai}\footnote{\url{https://github.com/SalesforceAIResearch/uni2ts}}:
Woo et al.~\cite{woo2024unified} introduce Moirai, a masked Transformer pre-trained on large multi-domain time series corpora for time series forecasting.

\item \textbf{TS-RAG}\footnote{\url{https://github.com/UConn-DSIS/TS-RAG}}:
Ning et al.~\cite{ningts2025} propose TS-RAG, a retrieval-augmented framework that integrates external patterns through an adaptive retrieval mixer module for zero-shot time series forecasting.

\item \textbf{TSPulse}\footnote{\url{https://github.com/ibm-granite/granite-tsfm/tree/main/notebooks/hfdemo/tspulse}}:
Vijay et al.~\cite{ekambaram2026tspulse} introduce TSPulse for time-series analysis, a lightweight pre-trained model using hybrid masked reconstruction across temporal, spectral, and semantic embedding spaces.

\end{itemize}

The remaining baselines (GPT4TS, MOMENT, UniTS, NuTime, Mantis, and UniShape) follow the same configurations as in the supervised setting. Compared to existing work~\cite{liu2026unishape}, we additionally include TS-RAG~\cite{ningts2025} and TSPulse~\cite{ekambaram2026tspulse} as new baselines for zero-shot feature extraction classification evaluation.

\subsection{Implementation details}

For supervised fine-tuning, we integrate R2R with three representative pre-trained time series foundation models, namely NuTime~\cite{lin2024nutime}, Mantis~\cite{feofanov2025mantis}, and UniShape~\cite{liu2026unishape}, and evaluate them on all 128 datasets from the UCR archive. For R2R combined with NuTime and Mantis, the maximum number of training epochs is set to 500 following the original recommendations, while UniShape is trained for up to 300 epochs. For multi-scale feature extraction in Eq.~(\ref{eq:downsample}), we employ three downsampling factors $\{1, 2, 4\}$ following~\cite{cui2016multi}. The sliding window length $l$ in Eq.~(\ref{eq:fenv}) is set to 11. In Eq.~(\ref{eq:ivfpq}), the $n_{\text{list}}$ is fixed at 256, $m$ is set to 16, and $b$ is set to 8. The number of nearest neighbors $K$ in Eq.~(\ref{eq:otweight}) is set to 20.
For the gradient-free backbone in R2R’s knowledge extraction module, we conduct model selection on each UCR dataset using a validation set sampled from the original training set.  The gradient-free backbone is then selected from MiniRocket~\cite{dempster2021minirocket}, MultiRocket~\cite{tan2022multirocket}, and the corresponding R2R-enhanced pre-trained TSFM~\cite{lin2024nutime,feofanov2025mantis,liu2026unishape} based on validation performance.
For zero-shot feature extraction on the additional 30 datasets, we fix the gradient-free backbone to MultiRocket across all experiments to ensure a zero-shot evaluation setting without parameter or model selection.
Following prior work~\cite{ismail2020inceptiontime,liu2026unishape}, we directly use the default training and test splits from the UCR archive for classification evaluation. Model selection is based on the minimum training loss due to the absence of a default validation set. To mitigate potential overfitting, we adopt an ensemble strategy during fine-tuning following~\cite{ismail2020inceptiontime,lin2024nutime}, where five independently trained models are aggregated to improve generalization with different learning ratios. This ensembling convention is applied consistently to R2R and most supervised deep learning and TSFM-based baselines, following the setting in~\cite{liu2026unishape}; MambaSL~\cite{jungmambasl2026} is reported as a single model because ensembling degraded its accuracy. The Non-DL baselines are gradient-free and follow their official single-model protocols, one of which, HIVE-COTE 2.0~\cite{middlehurst2021hive}, is itself an ensemble of about thirty algorithms drawn from four representation families.
For both R2R and all supervised baselines, the reported performance corresponds to the average test accuracy over five runs. All experiments are conducted on a server with two NVIDIA RTX A6000 GPUs, running Ubuntu 20.04.3 with Python 3.9 and PyTorch 1.12.

\textbf{Knowledge-base provenance and target-data usage.}
No test sample and no test label of any target dataset is used at any stage of R2R, including pretraining, knowledge-base construction, gradient-free backbone selection, and fine-tuning: the 1.89M source pool aggregates only the default \emph{training} splits of the datasets listed in Appendix~\ref{appendA-1}. The class-balanced sampler retains source classes with at least 50 samples and caps each class at 1{,}000, so whether a target dataset can contribute to the knowledge base at all is determined by its per-class training counts. Across the 128 UCR default training splits, no class reaches the 50-sample threshold in 72 datasets, 14 datasets are partly eligible, and 42 have every class eligible; retrieval is therefore strictly cross-dataset on those 72 datasets, whereas for the remaining 56 the knowledge base may contain training samples of the target task. Three controls indicate that the reported gains do not rest on this overlap. First, the full 1.89M pool, which contains every target training split, ranks below the 100K class-balanced subset in Table~\ref{tab:kd-18ucr} (4.17 versus 2.33), so adding target training data does not by itself improve R2R. Second, restricting Table~\ref{tab:main128ucr} to the 72 strictly cross-dataset targets leaves all three improvements intact and slightly larger: the average rank moves from 10.40 to 8.00 for NuTime, from 7.42 to 5.60 for Mantis, and from 4.97 to 3.07 for UniShape, with win/tie/loss counts of 40/15/17, 44/11/17, and 41/16/15. Third, rebuilding a 100K knowledge base that keeps only source classes with at least 381 training samples removes every target class of the 18 datasets in Table~\ref{tab:kd-18ucr}, whose largest per-class training count is 378, and reduces source-class coverage from 419 to 139; accuracy then changes by only $-0.26$ and $-0.40$ percentage points for R2R$_{\text{Mantis}}$ and R2R$_{\text{UniShape}}$, with win/tie/loss counts of 7/4/7, that is, no significant difference. We note that this last control excludes target classes by a size threshold rather than by provenance, because the released 1.89M pool does not carry dataset-of-origin labels, which makes a strict leave-one-dataset-out knowledge base infeasible. We also note that pretraining on the target training split alone is standard practice in self-supervised TSC~\cite{eldele2021time,yue2022ts2vec}, where, as here, the test split remains untouched.

\section{Details of results on 128 UCR datasets}\label{appendB}

R2R$_{\text{NuTime}}$, R2R$_{\text{Mantis}}$, and R2R$_{\text{UniShape}}$, together with the 18 selected baselines, are evaluated under the fully supervised classification setting on 128 UCR time series datasets. The detailed test classification accuracy results are reported in Table~\ref{tab:append-ucr-128}.

\normalsize
\fontsize{4.2pt}{5pt}\selectfont  % 第一个参数字号，第二个参数行距
\setlength{\tabcolsep}{1.4pt} % narrowed to absorb the extra TS-RAG column
% \begin{adjustbox}{max width=\textwidth}
\begin{longtable}[h]{@{}clccccccccccccccccccccc@{}}

\caption{Detailed test classification accuracy results on 128 UCR datasets under the fully supervised classification setting.
Due to space constraints, dataset names are abbreviated; full names can be referenced via their IDs in Table~\ref{tab:append-ucr_stats}. ``IncTime'' denotes InceptionTime. The best results are in \textbf{bold}.}
\label{tab:append-ucr-128} \\

\toprule
\multicolumn{1}{r}{ID} & Dataset & Rocket & MiniRocket & RDST & MR-H & HC2 & IncTime & TS2Vec & PatchTST & TimesNet & SoftShape & MambaSL & GPT4TS & MOMENT & UniTS & NuTime & TS-RAG & Mantis & UniShape & \makecell{R2R\\+NuTime} & \makecell{R2R\\+Mantis} & \makecell{R2R\\+UniShape} \\
\midrule
\endfirsthead

\toprule
\multicolumn{1}{r}{ID} & Dataset & Rocket & MiniRocket & RDST & MR-H & HC2 & IncTime & TS2Vec & PatchTST & TimesNet & SoftShape & MambaSL & GPT4TS & MOMENT & UniTS & NuTime & TS-RAG & Mantis & UniShape & \makecell{R2R\\+NuTime} & \makecell{R2R\\+Mantis} & \makecell{R2R\\+UniShape} \\
\midrule
\endhead

\midrule
\multicolumn{22}{r}{\textit{Continued on next page}} \\
\endfoot

% \bottomrule
\endlastfoot

1 & ACSF1 & 0.8600 & 0.9100 & 0.9300 & 0.8600 & 0.9100 & \textbf{0.9400} & 0.8800 & 0.7300 & 0.3000 & 0.8800 & 0.8600 & 0.6100 & 0.5000 & 0.7700 & 0.6700 & 0.7500 & 0.7000 & 0.8000 & 0.8200 & 0.8600 & 0.9300 \\
2 & Adiac & 0.7903 & 0.8235 & 0.7136 & 0.8344 & 0.8184 & 0.8107 & 0.7545 & 0.6675 & 0.6292 & 0.8159 & 0.7877 & 0.7826 & 0.1458 & 0.5115 & 0.8031 & 0.8235 & 0.8159 & 0.8289 & 0.8184 & \textbf{0.8440} & 0.8414 \\
3 & AllX. & 0.7629 & 0.7429 & 0.7471 & 0.7600 & 0.7671 & 0.7229 & 0.6729 & 0.2643 & 0.4529 & 0.7871 & 0.6114 & 0.5029 & 0.5157 & 0.6329 & 0.7086 & 0.7600 & \textbf{0.7914} & 0.7743 & 0.6929 & 0.7829 & 0.7743 \\
4 & AllY. & 0.7486 & 0.7571 & 0.7386 & 0.7886 & 0.7586 & 0.7943 & 0.6771 & 0.2529 & 0.4557 & 0.7643 & 0.7043 & 0.5129 & 0.5986 & 0.6443 & 0.7557 & 0.8000 & 0.8243 & 0.7929 & 0.7429 & \textbf{0.8253} & 0.8086 \\
5 & AllZ. & \textbf{0.7886} & 0.7314 & 0.7100 & 0.7529 & \textbf{0.7886} & 0.7771 & 0.7371 & 0.1757 & 0.4071 & 0.7443 & 0.5900 & 0.4071 & 0.4886 & 0.5700 & 0.6614 & 0.7300 & 0.7586 & 0.7429 & 0.6743 & 0.7300 & 0.7657 \\
6 & Arrd. & 0.8057 & 0.8400 & 0.8686 & 0.8629 & 0.9086 & 0.8400 & 0.8400 & 0.6971 & 0.7486 & 0.8457 & 0.7429 & 0.8057 & 0.5657 & 0.7543 & 0.8343 & 0.8971 & 0.8686 & 0.8857 & \textbf{0.9257} & 0.9200 & \textbf{0.9257} \\
7 & Beef & 0.8000 & 0.8667 & 0.8333 & 0.8667 & 0.8667 & 0.7333 & 0.7000 & 0.7333 & 0.6000 & 0.9200 & 0.6000 & 0.8000 & 0.5333 & 0.7333 & 0.9167 & 0.9333 & 0.7333 & 0.8667 & \textbf{0.9667} & 0.7667 & 0.8333 \\
8 & Beey. & 0.9000 & 0.8500 & \textbf{0.9500} & 0.8400 & 0.9000 & 0.9000 & 0.8500 & 0.7000 & 0.6500 & 0.9000 & 0.7500 & 0.8500 & \textbf{0.9500} & 0.7000 & 0.7500 & \textbf{0.9500} & 0.9000 & 0.8500 & 0.7000 & 0.9000 & \textbf{0.9500} \\
9 & Birn. & 0.9000 & 0.9000 & 0.9000 & 0.9000 & 0.9000 & 0.9000 & 0.8000 & 0.9000 & 0.8000 & \textbf{1.0000} & 0.8000 & 0.5500 & 0.8500 & 0.8000 & 0.9000 & 0.8000 & \textbf{1.0000} & 0.9500 & 0.9000 & 0.9000 & 0.9000 \\
10 & BME & \textbf{1.0000} & 0.9933 & 0.9933 & \textbf{1.0000} & \textbf{1.0000} & \textbf{1.0000} & 0.9800 & 0.9733 & 0.9933 & 0.9867 & 0.9933 & 0.9533 & 0.8800 & 0.9733 & 0.9933 & \textbf{1.0000} & \textbf{1.0000} & \textbf{1.0000} & \textbf{1.0000} & \textbf{1.0000} & \textbf{1.0000} \\
11 & Car & 0.9000 & 0.9167 & \textbf{0.9333} & 0.9233 & 0.9000 & 0.8833 & 0.7833 & 0.7167 & 0.7667 & 0.8667 & 0.8167 & 0.8333 & 0.6667 & 0.7833 & 0.8333 & 0.8833 & 0.8833 & \textbf{0.9333} & \textbf{0.9333} & 0.9167 & 0.9167 \\
12 & CBF & \textbf{1.0000} & 0.9989 & 0.9922 & 0.9956 & \textbf{1.0000} & 0.9956 & \textbf{1.0000} & 0.7489 & 0.9356 & 0.9978 & 0.9944 & 0.8278 & 0.9700 & 0.8767 & 0.9878 & 0.9978 & \textbf{1.0000} & \textbf{1.0000} & 0.9956 & \textbf{1.0000} & \textbf{1.0000} \\
13 & Chin. & 0.9826 & 0.9855 & 0.9855 & 0.9497 & 0.9825 & 0.9855 & 0.9768 & 0.9072 & \textbf{0.9884} & 0.9826 & 0.9767 & 0.9652 & 0.9855 & 0.9768 & 0.9797 & 0.9825 & 0.9855 & 0.9565 & 0.9739 & 0.9710 & 0.9855 \\
14 & Chln. & 0.8156 & 0.7630 & 0.7576 & 0.8193 & 0.7622 & 0.8654 & 0.8281 & 0.6523 & 0.8229 & \textbf{0.8971} & 0.7216 & 0.8242 & 0.5664 & 0.8578 & 0.7630 & 0.8526 & 0.7633 & 0.8539 & 0.7622 & 0.7495 & 0.8635 \\
15 & Cino. & 0.8304 & 0.8688 & 0.9335 & 0.9288 & \textbf{0.9993} & 0.6638 & 0.7993 & 0.7457 & 0.8159 & 0.9036 & 0.6283 & 0.7558 & 0.5565 & 0.8058 & 0.9609 & 0.8688 & 0.8449 & 0.9500 & 0.9623 & 0.9043 & 0.9732 \\
16 & Coffee & \textbf{1.0000} & \textbf{1.0000} & \textbf{1.0000} & \textbf{1.0000} & \textbf{1.0000} & \textbf{1.0000} & \textbf{1.0000} & 0.9643 & \textbf{1.0000} & \textbf{1.0000} & 0.9643 & \textbf{1.0000} & 0.8929 & \textbf{1.0000} & \textbf{1.0000} & \textbf{1.0000} & \textbf{1.0000} & \textbf{1.0000} & \textbf{1.0000} & \textbf{1.0000} & \textbf{1.0000} \\
17 & Coms. & 0.7720 & 0.7360 & 0.7360 & 0.7618 & 0.7680 & 0.7880 & 0.6600 & 0.5080 & 0.5680 & 0.6560 & 0.7760 & 0.5760 & 0.6480 & 0.5480 & \textbf{0.7920} & 0.7320 & 0.7400 & 0.7680 & \textbf{0.7920} & 0.7360 & 0.7800 \\
18 & CriX. & 0.8282 & 0.8103 & 0.8026 & 0.8026 & \textbf{0.8308} & 0.7923 & 0.7923 & 0.5205 & 0.5385 & 0.8256 & 0.5821 & 0.5333 & 0.6949 & 0.6872 & 0.7821 & 0.8282 & 0.8179 & 0.8282 & 0.8077 & 0.8205 & \textbf{0.8308} \\
19 & CriY. & 0.8564 & 0.8385 & 0.8308 & 0.8251 & 0.8410 & 0.7974 & 0.7564 & 0.6026 & 0.5538 & 0.8051 & 0.7154 & 0.5615 & 0.6564 & 0.6692 & 0.8026 & 0.8385 & 0.8308 & 0.8333 & 0.7821 & \textbf{0.8590} & 0.8487 \\
20 & CriZ. & 0.8590 & 0.8282 & 0.8462 & 0.8385 & 0.8436 & 0.8282 & 0.8128 & 0.5615 & 0.6026 & 0.8462 & 0.6846 & 0.5564 & 0.7308 & 0.6949 & 0.7974 & 0.8410 & 0.8590 & 0.8436 & 0.8205 & \textbf{0.8599} & 0.8462 \\
21 & Crop & 0.7574 & 0.7622 & 0.7637 & 0.7678 & 0.7602 & 0.7727 & 0.7540 & 0.0766 & 0.7197 & 0.7733 & 0.2406 & 0.6665 & 0.6605 & 0.6548 & 0.7888 & 0.7730 & 0.7877 & 0.7725 & \textbf{0.7891} & 0.7842 & 0.7839 \\
22 & Dian. & 0.9771 & 0.9281 & 0.9248 & 0.9641 & 0.9641 & 0.9150 & 0.9869 & \textbf{0.9902} & 0.9150 & 0.9771 & 0.9739 & 0.9542 & 0.7222 & 0.9118 & 0.9085 & 0.9837 & 0.9739 & 0.9706 & 0.9575 & 0.9281 & 0.9739 \\
23 & Disp. & 0.7410 & 0.7338 & 0.7842 & 0.7626 & 0.7554 & 0.7482 & 0.7122 & 0.7338 & 0.6763 & 0.7770 & 0.7698 & 0.7482 & 0.6691 & 0.7122 & 0.7266 & 0.7194 & 0.7050 & 0.7698 & 0.7338 & 0.7770 & \textbf{0.7914} \\
24 & Dist. & 0.7609 & 0.7681 & 0.7717 & 0.7616 & 0.7609 & 0.7536 & 0.7572 & 0.6848 & 0.7681 & 0.7790 & 0.7428 & 0.7391 & 0.6775 & 0.7246 & 0.7609 & 0.7826 & 0.7681 & 0.7899 & 0.7717 & \textbf{0.8116} & 0.7862 \\
25 & DisW. & 0.7050 & 0.6763 & \textbf{0.7266} & 0.6978 & 0.7122 & 0.7122 & 0.6978 & 0.6331 & 0.6547 & 0.6835 & 0.6763 & 0.6043 & 0.6619 & 0.6547 & 0.6835 & 0.6619 & 0.6547 & 0.7194 & 0.7194 & 0.6978 & 0.7050 \\
26 & Dody. & 0.6125 & 0.6875 & 0.6750 & 0.5750 & 0.6500 & 0.6000 & 0.5500 & 0.4250 & 0.2500 & 0.6500 & 0.5750 & 0.4500 & 0.3000 & 0.5125 & 0.5125 & 0.5250 & 0.5750 & 0.6250 & 0.5375 & 0.6875 & \textbf{0.7250} \\
27 & Dode. & 0.8696 & 0.8768 & 0.8478 & 0.8986 & 0.9203 & 0.9348 & 0.8986 & 0.6522 & 0.5145 & 0.8696 & 0.8333 & 0.6884 & 0.8333 & 0.8261 & 0.8623 & 0.9348 & 0.8406 & 0.9203 & 0.8333 & \textbf{0.9565} & 0.9420 \\
28 & Dodd. & 0.9710 & \textbf{0.9855} & 0.9783 & 0.9755 & 0.9783 & 0.9783 & 0.9638 & 0.9348 & 0.7609 & 0.9783 & \textbf{0.9855} & 0.9710 & 0.9493 & 0.9710 & 0.9710 & 0.9710 & 0.9783 & \textbf{0.9855} & 0.9783 & \textbf{0.9855} & \textbf{0.9855} \\
29 & Ears. & 0.7554 & 0.7626 & 0.7338 & 0.7338 & 0.7482 & 0.7554 & 0.7482 & 0.7266 & 0.7554 & 0.7482 & 0.7338 & 0.7410 & 0.7482 & 0.6978 & 0.7482 & 0.7482 & 0.7482 & 0.7554 & 0.7626 & 0.7554 & \textbf{0.7770} \\
30 & ECG2. & 0.9051 & 0.9100 & 0.9000 & 0.9100 & 0.8600 & 0.9100 & 0.8700 & 0.8400 & 0.8300 & 0.9150 & 0.8800 & 0.8500 & 0.8700 & 0.8800 & 0.8000 & 0.8900 & 0.9000 & 0.9200 & 0.8800 & 0.9300 & \textbf{0.9400} \\
31 & ECG5. & 0.9478 & 0.9449 & 0.9429 & 0.9460 & 0.9478 & 0.9453 & 0.9404 & 0.9209 & 0.8993 & 0.9436 & 0.9369 & 0.9309 & 0.9424 & 0.9376 & 0.9440 & 0.9453 & 0.9422 & \textbf{0.9498} & 0.9418 & 0.9456 & 0.9462 \\
32 & ECGs. & \textbf{1.0000} & \textbf{1.0000} & 0.9977 & \textbf{1.0000} & \textbf{1.0000} & \textbf{1.0000} & \textbf{1.0000} & 0.7689 & 0.8920 & 0.9965 & 0.7352 & 0.9547 & 0.8281 & 0.8211 & 0.9768 & \textbf{1.0000} & 0.9570 & \textbf{1.0000} & 0.9640 & 0.9884 & \textbf{1.0000} \\
33 & Eles. & 0.7312 & 0.7496 & 0.7408 & 0.7424 & \textbf{0.7605} & 0.7256 & 0.7255 & 0.3167 & 0.6457 & 0.6676 & 0.7024 & 0.5774 & 0.6388 & 0.6090 & 0.7506 & 0.7574 & 0.7373 & 0.7248 & 0.7475 & 0.7399 & 0.7567 \\
34 & EOGl. & 0.6492 & 0.5856 & 0.6768 & 0.6354 & 0.6519 & 0.6326 & 0.5552 & 0.4586 & 0.3094 & 0.6298 & 0.5912 & 0.2680 & 0.4558 & 0.4724 & 0.6657 & 0.6381 & 0.6768 & 0.6796 & 0.6657 & \textbf{0.6851} & 0.6796 \\
35 & EOGl. & 0.5470 & 0.5470 & 0.5497 & 0.5138 & 0.5525 & 0.4834 & 0.4420 & 0.3398 & 0.2624 & 0.5608 & 0.4337 & 0.2541 & 0.3978 & 0.5083 & 0.5552 & 0.5525 & 0.5359 & 0.5193 & \textbf{0.5718} & 0.5525 & 0.5442 \\
36 & Ethl. & 0.5800 & 0.6000 & 0.6600 & 0.6380 & 0.6820 & \textbf{0.8520} & 0.4620 & 0.4260 & 0.5140 & 0.8100 & 0.5560 & 0.6740 & 0.2640 & 0.5600 & 0.6340 & 0.6820 & 0.4460 & 0.8500 & 0.6460 & 0.5940 & 0.8140 \\
37 & Facl. & \textbf{0.9485} & 0.8077 & 0.7941 & 0.8077 & 0.8787 & 0.8041 & 0.7669 & 0.4876 & 0.6799 & 0.8006 & 0.8396 & 0.7479 & 0.6982 & 0.7467 & 0.7976 & 0.7935 & 0.8178 & 0.8994 & 0.7793 & 0.7864 & 0.7852 \\
38 & Facr. & 0.9773 & 0.9886 & 0.9773 & 0.9545 & 0.9886 & 0.8864 & 0.8977 & 0.5341 & 0.8182 & 0.9659 & 0.9091 & 0.7614 & 0.8636 & 0.6818 & 0.9545 & 0.9773 & 0.9886 & 0.9545 & 0.9659 & \textbf{1.0000} & 0.9773 \\
39 & FacR. & 0.9615 & 0.9595 & 0.9629 & 0.9595 & \textbf{0.9654} & 0.9473 & 0.9341 & 0.6902 & 0.8317 & 0.9624 & 0.8346 & 0.8307 & 0.7956 & 0.8098 & 0.9210 & 0.9444 & 0.9522 & 0.9620 & 0.9483 & 0.9551 & 0.9566 \\
40 & Fifs. & 0.8242 & 0.8418 & 0.8484 & 0.8415 & 0.8330 & 0.7473 & 0.7648 & 0.6440 & 0.5824 & 0.8308 & 0.6593 & 0.6879 & 0.7714 & 0.7143 & 0.7890 & 0.8352 & \textbf{0.8637} & 0.8462 & 0.8022 & 0.8527 & 0.8462 \\
41 & Fish & 0.9829 & 0.9771 & \textbf{0.9943} & 0.9829 & 0.9886 & 0.9771 & 0.9429 & 0.7543 & 0.7600 & 0.9714 & 0.8571 & 0.8800 & 0.7543 & 0.8800 & 0.9600 & 0.9543 & 0.9771 & 0.9886 & 0.9829 & 0.9899 & 0.9829 \\
42 & FordA & 0.9311 & 0.9470 & 0.9470 & 0.9553 & 0.9545 & 0.9515 & 0.9348 & 0.7841 & 0.9136 & 0.9606 & 0.9333 & 0.8818 & 0.9174 & 0.9311 & 0.9348 & 0.9591 & 0.9439 & 0.9629 & 0.9470 & 0.9500 & \textbf{0.9644} \\
43 & FordB & 0.8049 & 0.8198 & 0.8160 & 0.8309 & 0.8309 & 0.8580 & 0.7938 & 0.5716 & 0.7519 & 0.8679 & 0.7938 & 0.7395 & 0.7802 & 0.7716 & 0.8074 & 0.8370 & 0.8160 & 0.8605 & 0.8000 & 0.8210 & \textbf{0.8704} \\
44 & Fren. & 0.9979 & \textbf{0.9996} & 0.9989 & \textbf{0.9996} & \textbf{0.9996} & 0.9905 & 0.9814 & 0.9825 & 0.9628 & 0.9909 & 0.9961 & 0.9523 & 0.7747 & 0.9926 & 0.9944 & 0.9926 & 0.9958 & 0.9972 & 0.9968 & 0.9877 & 0.9993 \\
45 & Free. & 0.9523 & 0.9716 & 0.9944 & 0.9951 & \textbf{0.9989} & 0.8832 & 0.8986 & 0.9004 & 0.7337 & 0.8154 & 0.9568 & 0.6744 & 0.7733 & 0.6375 & 0.9530 & 0.7228 & 0.8140 & 0.9828 & 0.9582 & 0.8895 & 0.9846 \\
46 & Fungi & \textbf{1.0000} & \textbf{1.0000} & \textbf{1.0000} & \textbf{1.0000} & \textbf{1.0000} & \textbf{1.0000} & 0.9355 & 0.3978 & 0.8656 & \textbf{1.0000} & 0.9624 & 0.7312 & 0.9785 & 0.8011 & 0.8065 & 0.9946 & 0.9946 & \textbf{1.0000} & 0.8441 & 0.9946 & \textbf{1.0000} \\
47 & Ges1. & 0.7462 & 0.7308 & 0.7615 & 0.7846 & 0.7462 & 0.7538 & 0.6615 & 0.3462 & 0.4538 & 0.7538 & 0.6308 & 0.5615 & 0.6692 & 0.6846 & 0.7462 & 0.7462 & \textbf{0.8077} & 0.8000 & 0.7769 & 0.7769 & \textbf{0.8077} \\
48 & Ges2. & 0.6615 & 0.6385 & 0.6538 & 0.7000 & 0.6846 & 0.6154 & 0.4846 & 0.1462 & 0.3615 & 0.6308 & 0.4769 & 0.4615 & 0.5462 & 0.4769 & 0.6846 & 0.6308 & 0.6923 & 0.7231 & 0.7154 & \textbf{0.7385} & 0.7231 \\
49 & Ges3. & 0.4615 & 0.4769 & 0.5308 & \textbf{0.5462} & 0.5385 & 0.3615 & 0.3769 & 0.1846 & 0.2923 & 0.3692 & 0.3462 & 0.3615 & 0.2846 & 0.3769 & \textbf{0.5462} & 0.4846 & 0.5154 & 0.4692 & \textbf{0.5462} & 0.4923 & 0.5385 \\
50 & GeZ1. & 0.9012 & 0.8895 & 0.9012 & 0.9128 & 0.8953 & 0.8837 & 0.5581 & 0.5349 & 0.7558 & 0.9186 & 0.8256 & 0.6686 & 0.7616 & 0.8663 & 0.9012 & 0.8953 & 0.9244 & \textbf{0.9477} & 0.9012 & 0.9244 & 0.9419 \\
51 & GeZ2. & 0.8924 & 0.9051 & 0.8987 & 0.8671 & 0.8987 & 0.9051 & 0.5127 & 0.5063 & 0.5886 & 0.9494 & 0.8418 & 0.5759 & 0.8038 & 0.8924 & 0.8354 & 0.8671 & 0.8861 & \textbf{0.9747} & 0.8354 & 0.8924 & 0.9557 \\
52 & Gunt. & \textbf{1.0000} & 0.9933 & \textbf{1.0000} & \textbf{1.0000} & \textbf{1.0000} & \textbf{1.0000} & 0.9800 & 0.9400 & 0.9667 & \textbf{1.0000} & 0.9733 & 0.9600 & 0.9267 & 0.9667 & 0.9933 & 0.9867 & \textbf{1.0000} & \textbf{1.0000} & 0.9933 & \textbf{1.0000} & \textbf{1.0000} \\
53 & Gunn. & 0.9968 & 0.9968 & 0.9968 & \textbf{1.0000} & 0.9968 & 0.9747 & 0.9557 & 0.7342 & 0.8797 & 0.9810 & 0.9937 & 0.9494 & 0.9177 & 0.9209 & 0.9873 & 0.9842 & 0.9968 & 0.9968 & 0.9937 & 0.9747 & \textbf{1.0000} \\
54 & Gune. & \textbf{1.0000} & \textbf{1.0000} & \textbf{1.0000} & \textbf{1.0000} & \textbf{1.0000} & 0.9968 & 0.9937 & 0.9905 & 0.9905 & 0.9842 & 0.9968 & 0.9968 & 0.9778 & 0.9937 & \textbf{1.0000} & 0.9968 & \textbf{1.0000} & \textbf{1.0000} & \textbf{1.0000} & \textbf{1.0000} & \textbf{1.0000} \\
55 & Gung. & 0.9873 & \textbf{1.0000} & \textbf{1.0000} & \textbf{1.0000} & \textbf{1.0000} & 0.7587 & \textbf{1.0000} & \textbf{1.0000} & 0.9365 & 0.9746 & \textbf{1.0000} & 0.9175 & 0.8667 & 0.9460 & \textbf{1.0000} & 0.9841 & \textbf{1.0000} & \textbf{1.0000} & \textbf{1.0000} & \textbf{1.0000} & \textbf{1.0000} \\
56 & Ham & 0.7429 & 0.6952 & 0.7429 & 0.7429 & 0.7048 & 0.7429 & 0.7524 & 0.6000 & 0.6952 & \textbf{0.7714} & 0.6952 & 0.6762 & 0.6095 & 0.7333 & 0.6952 & 0.7238 & 0.7333 & 0.7619 & 0.7333 & 0.6667 & 0.7333 \\
57 & Hans. & 0.9432 & 0.9432 & 0.9514 & 0.9459 & 0.9459 & 0.9514 & 0.9189 & 0.9135 & 0.8649 & 0.9324 & 0.9270 & 0.8892 & 0.8108 & 0.9054 & 0.9270 & 0.9135 & 0.9486 & 0.9649 & 0.9351 & 0.9405 & \textbf{0.9703} \\
58 & Haptics & 0.5455 & 0.5195 & \textbf{0.5747} & 0.5260 & 0.5584 & 0.4935 & 0.5130 & 0.4091 & 0.4383 & 0.4805 & 0.4578 & 0.4383 & 0.4091 & 0.4481 & 0.5195 & 0.5390 & 0.4968 & 0.5682 & 0.5422 & 0.5552 & \textbf{0.5747} \\
59 & Herring & 0.6250 & 0.6875 & 0.6875 & 0.7344 & 0.6406 & 0.6250 & 0.5625 & 0.6250 & 0.4844 & 0.5938 & 0.5781 & 0.5625 & 0.6406 & 0.6094 & 0.5312 & 0.6094 & 0.5469 & \textbf{0.7969} & 0.6719 & 0.6094 & 0.7656 \\
60 & Houy. & 0.9580 & 0.9664 & 0.9664 & \textbf{0.9916} & 0.9748 & 0.9832 & 0.7899 & 0.6891 & 0.6639 & 0.9580 & \textbf{0.9916} & 0.6050 & 0.9244 & 0.8235 & 0.9076 & 0.9412 & 0.9832 & 0.9832 & 0.9160 & 0.9328 & \textbf{0.9916} \\
61 & Inle. & 0.4764 & 0.4764 & 0.4436 & 0.4964 & \textbf{0.5418} & 0.4327 & 0.3836 & 0.3236 & 0.2309 & 0.3764 & 0.3473 & 0.2982 & 0.2255 & 0.2927 & 0.4055 & 0.4218 & 0.4164 & 0.4909 & 0.4018 & 0.4473 & 0.5200 \\
62 & Insn. & \textbf{1.0000} & \textbf{1.0000} & \textbf{1.0000} & \textbf{1.0000} & \textbf{1.0000} & \textbf{1.0000} & \textbf{1.0000} & \textbf{1.0000} & \textbf{1.0000} & \textbf{1.0000} & \textbf{1.0000} & \textbf{1.0000} & 0.8996 & 0.7470 & \textbf{1.0000} & \textbf{1.0000} & \textbf{1.0000} & \textbf{1.0000} & \textbf{1.0000} & \textbf{1.0000} & \textbf{1.0000} \\
63 & Inse. & 0.9839 & 0.9960 & 0.9960 & \textbf{1.0000} & \textbf{1.0000} & \textbf{1.0000} & \textbf{1.0000} & \textbf{1.0000} & \textbf{1.0000} & 0.9357 & \textbf{1.0000} & \textbf{1.0000} & 0.8755 & 0.7269 & \textbf{1.0000} & 0.9277 & \textbf{1.0000} & \textbf{1.0000} & \textbf{1.0000} & 0.9839 & \textbf{1.0000} \\
64 & Innd. & 0.6697 & 0.6702 & 0.6566 & 0.6572 & 0.6631 & 0.6298 & 0.6338 & 0.5328 & 0.6232 & 0.6646 & 0.4798 & 0.6167 & 0.6247 & 0.6071 & 0.6480 & 0.6611 & 0.6470 & 0.6682 & \textbf{0.6758} & 0.6641 & 0.6737 \\
65 & Itad. & \textbf{0.9689} & 0.9640 & 0.8756 & 0.9679 & \textbf{0.9689} & 0.9670 & 0.9572 & 0.9252 & 0.9514 & 0.9640 & 0.9378 & 0.9553 & 0.9349 & 0.9572 & 0.9572 & 0.9650 & 0.9543 & 0.9640 & 0.9660 & 0.9660 & 0.9650 \\
66 & Lars. & 0.9040 & 0.8720 & 0.8453 & 0.9088 & 0.8987 & 0.9067 & 0.8667 & 0.5493 & 0.5227 & 0.8773 & 0.8480 & 0.5040 & 0.7600 & 0.7280 & 0.8640 & 0.8827 & 0.8640 & \textbf{0.9253} & 0.8667 & 0.8693 & 0.9200 \\
67 & Lig2. & 0.7869 & 0.7541 & 0.7541 & 0.7213 & 0.7869 & 0.8689 & 0.8525 & 0.7049 & 0.6393 & 0.7869 & \textbf{0.8852} & 0.6721 & 0.7213 & 0.7705 & 0.7541 & 0.8361 & 0.8197 & 0.8525 & 0.7705 & 0.8525 & 0.8361 \\
68 & Lig7. & 0.8082 & 0.7808 & 0.7945 & 0.8219 & 0.8082 & 0.8082 & 0.8493 & 0.4932 & 0.7123 & 0.8493 & 0.8219 & 0.7534 & 0.7260 & 0.6575 & 0.7397 & 0.7397 & 0.8082 & \textbf{0.8630} & 0.7945 & 0.8356 & 0.8356 \\
69 & Mallat & 0.9561 & 0.9450 & 0.9258 & 0.9288 & 0.9757 & 0.9629 & 0.8772 & 0.8333 & 0.9011 & 0.9557 & 0.8648 & 0.8789 & 0.4307 & 0.8985 & 0.9441 & 0.9403 & 0.9608 & 0.9659 & 0.9394 & 0.9697 & \textbf{0.9898} \\
70 & Meat & 0.9500 & 0.9500 & 0.9667 & 0.9333 & 0.9333 & 0.9167 & 0.9667 & 0.6500 & 0.9000 & 0.9333 & 0.9167 & 0.9000 & 0.6667 & 0.8833 & 0.9333 & 0.8500 & 0.7167 & 0.9500 & 0.9500 & 0.9167 & \textbf{0.9833} \\
71 & Meds. & 0.8000 & 0.8013 & 0.7658 & 0.7974 & 0.8039 & 0.8000 & 0.7895 & 0.7039 & 0.6908 & 0.7276 & 0.7632 & 0.6987 & 0.7211 & 0.6868 & 0.7868 & 0.7855 & 0.7961 & 0.8039 & 0.8158 & \textbf{0.8237} & 0.8066 \\
72 & Meln. & 0.9122 & 0.9616 & 0.9645 & 0.9690 & 0.9331 & 0.9216 & 0.9514 & 0.4498 & 0.9224 & 0.9000 & 0.8852 & 0.9155 & 0.7976 & 0.8710 & \textbf{0.9784} & 0.9225 & 0.9731 & 0.9702 & 0.9661 & 0.9735 & 0.9751 \\
73 & Midp. & 0.5649 & 0.5714 & 0.5974 & 0.5649 & 0.5714 & 0.5519 & 0.6429 & 0.5519 & 0.5844 & 0.5519 & 0.6429 & 0.4870 & \textbf{0.6558} & 0.4610 & 0.4870 & 0.5519 & 0.5065 & 0.5714 & 0.5779 & 0.6364 & 0.6039 \\
74 & Midt. & 0.8282 & 0.8419 & 0.8247 & 0.8488 & 0.8351 & 0.8419 & 0.8076 & 0.7354 & 0.7869 & 0.8351 & 0.7423 & 0.7698 & 0.5739 & 0.7904 & 0.8213 & 0.8247 & 0.8213 & 0.8660 & 0.8179 & 0.8591 & \textbf{0.8763} \\
75 & MidW. & 0.5455 & 0.5455 & 0.5325 & 0.5130 & 0.5779 & 0.5325 & 0.6039 & 0.5519 & 0.5455 & 0.5519 & \textbf{0.6104} & 0.5649 & 0.5974 & 0.4805 & 0.4740 & 0.5065 & 0.5065 & 0.5974 & 0.5065 & 0.5844 & 0.5714 \\
76 & Mixn. & 0.9720 & 0.9707 & 0.9810 & \textbf{0.9823} & 0.9769 & 0.9452 & 0.9241 & 0.6177 & 0.7885 & 0.9472 & 0.8652 & 0.8610 & 0.7744 & 0.8854 & 0.9724 & 0.9703 & 0.9773 & 0.9753 & 0.9744 & 0.9757 & 0.9761 \\
77 & Mixed. & 0.9365 & 0.9509 & \textbf{0.9687} & 0.9507 & 0.9633 & 0.8874 & 0.8623 & 0.7728 & 0.5744 & 0.8858 & 0.7909 & 0.7955 & 0.7023 & 0.8041 & 0.9460 & 0.9278 & 0.9579 & 0.9509 & 0.9377 & 0.9394 & 0.9435 \\
78 & Motn. & 0.9161 & 0.9281 & 0.9217 & 0.9457 & 0.9681 & 0.9089 & 0.8570 & 0.8690 & 0.8922 & 0.9081 & 0.8626 & 0.8786 & 0.8986 & 0.8770 & 0.9561 & 0.9201 & 0.9377 & 0.9305 & 0.9681 & \textbf{0.9720} & 0.9457 \\
79 & Non1. & 0.9542 & 0.9496 & 0.9252 & \textbf{0.9644} & 0.9506 & 0.9191 & 0.9288 & 0.8952 & 0.8229 & 0.9511 & 0.9257 & 0.8926 & 0.6153 & 0.9079 & 0.9354 & 0.9486 & 0.9267 & 0.9511 & 0.9445 & 0.9221 & 0.9598 \\
80 & Non2. & \textbf{0.9684} & 0.9598 & 0.9394 & \textbf{0.9684} & 0.9654 & 0.9201 & 0.9425 & 0.9232 & 0.8555 & 0.9466 & 0.9191 & 0.9084 & 0.6911 & 0.9064 & 0.9491 & 0.9511 & 0.9344 & 0.9517 & 0.9659 & 0.9649 & 0.9664 \\
81 & Olil. & 0.9333 & 0.9333 & 0.9333 & \textbf{0.9667} & 0.8667 & 0.7000 & 0.9000 & 0.4000 & 0.7667 & 0.8667 & 0.7333 & 0.7667 & 0.4000 & 0.5000 & 0.7333 & 0.7000 & 0.8667 & 0.8667 & \textbf{0.9667} & \textbf{0.9667} & \textbf{0.9667} \\
82 & OSUf. & 0.9421 & 0.9504 & 0.9815 & 0.9628 & 0.9669 & 0.9174 & 0.8512 & 0.5165 & 0.5041 & 0.8678 & 0.8430 & 0.5165 & 0.7438 & 0.5909 & 0.9380 & 0.9256 & 0.9545 & 0.9835 & 0.9463 & 0.9545 & \textbf{0.9917} \\
83 & Phat. & 0.8275 & 0.8333 & 0.8089 & 0.8450 & 0.8380 & 0.8508 & 0.8263 & 0.7075 & 0.7879 & 0.8298 & 0.7657 & 0.7727 & 0.6340 & 0.7389 & 0.8159 & 0.8462 & 0.8263 & \textbf{0.8590} & 0.8497 & 0.8485 & 0.8485 \\
84 & Phoe. & 0.2811 & 0.2911 & 0.3370 & 0.3539 & \textbf{0.4056} & 0.3402 & 0.3085 & 0.1218 & 0.0860 & 0.3133 & 0.2036 & 0.1160 & 0.3017 & 0.1313 & 0.3238 & 0.3291 & 0.3576 & 0.3729 & 0.3354 & 0.3571 & 0.3877 \\
85 & PicZ. & 0.8000 & 0.8800 & 0.8400 & 0.8200 & 0.8400 & 0.8000 & 0.6200 & 0.6200 & 0.5800 & 0.8600 & 0.7600 & 0.5400 & 0.6600 & 0.7200 & 0.7400 & 0.8600 & 0.8000 & 0.8600 & 0.7200 & \textbf{0.9000} & 0.8800 \\
86 & PigA. & 0.4038 & 0.8510 & 0.8269 & 0.7115 & \textbf{0.9808} & 0.1202 & 0.3269 & 0.0433 & 0.0337 & 0.1154 & 0.3269 & 0.0673 & 0.0577 & 0.0769 & 0.5913 & 0.3029 & 0.4279 & 0.3510 & 0.6298 & 0.4808 & 0.4663 \\
87 & Pige. & 0.9567 & 0.9856 & 0.9519 & 0.9567 & \textbf{1.0000} & 0.8317 & 0.1490 & 0.1154 & 0.0192 & 0.9663 & 0.7404 & 0.0769 & 0.1154 & 0.1346 & 0.8846 & 0.7067 & 0.9038 & 0.9856 & 0.9471 & 0.9712 & 0.9952 \\
88 & PigP. & 0.9423 & 0.9519 & 0.9327 & 0.8942 & \textbf{0.9567} & 0.8837 & 0.6827 & 0.0577 & 0.0288 & 0.7933 & 0.7212 & 0.1010 & 0.0865 & 0.0529 & 0.8750 & 0.7356 & 0.8558 & 0.7500 & 0.9038 & 0.9038 & 0.8654 \\
89 & PLAD. & 0.5791 & \textbf{0.9441} & 0.9199 & 0.9311 & 0.9106 & 0.7831 & 0.5289 & 0.2197 & 0.5289 & 0.5698 & 0.5531 & 0.4916 & 0.3575 & 0.5009 & 0.9050 & 0.5717 & 0.8864 & 0.8939 & 0.8845 & 0.8734 & 0.9050 \\
90 & Plane & \textbf{1.0000} & \textbf{1.0000} & \textbf{1.0000} & \textbf{1.0000} & \textbf{1.0000} & 0.9905 & 0.9905 & 0.9619 & 0.9714 & \textbf{1.0000} & \textbf{1.0000} & 0.9810 & 0.9810 & 0.9714 & \textbf{1.0000} & \textbf{1.0000} & \textbf{1.0000} & 0.9905 & \textbf{1.0000} & \textbf{1.0000} & \textbf{1.0000} \\
91 & Pows. & 0.9333 & 0.9944 & 0.9944 & 0.9833 & 0.9833 & 0.9889 & 0.9556 & \textbf{1.0000} & 0.9944 & 0.9778 & 0.9556 & 0.9944 & 0.8722 & 0.9889 & \textbf{1.0000} & 0.9722 & \textbf{1.0000} & \textbf{1.0000} & \textbf{1.0000} & 0.9889 & 0.9944 \\
92 & Prop. & 0.8585 & 0.8439 & 0.8488 & 0.8634 & 0.8634 & 0.8585 & 0.8341 & 0.8098 & 0.8634 & 0.8537 & 0.8439 & 0.8537 & 0.8488 & 0.8488 & \textbf{0.8732} & 0.8341 & 0.8293 & 0.8585 & \textbf{0.8732} & 0.8585 & 0.8585 \\
93 & Prot. & 0.8969 & 0.9038 & 0.8935 & 0.9278 & 0.8900 & 0.9107 & 0.8625 & 0.7732 & 0.9175 & 0.8900 & 0.8866 & 0.8660 & 0.7560 & 0.8557 & 0.8797 & 0.9107 & 0.8454 & \textbf{0.9347} & 0.8935 & 0.8832 & 0.9278 \\
94 & ProW. & 0.7951 & 0.7951 & 0.8293 & 0.8049 & 0.8195 & 0.7854 & 0.8244 & 0.7707 & 0.8000 & 0.7951 & 0.7854 & 0.8000 & 0.7561 & 0.7659 & 0.7659 & 0.7805 & 0.7902 & 0.8341 & 0.7951 & \textbf{0.8390} & \textbf{0.8390} \\
95 & Refs. & 0.5280 & 0.4773 & 0.5680 & 0.5173 & 0.5680 & 0.5253 & 0.6000 & 0.3493 & 0.4160 & 0.5360 & 0.5360 & 0.4187 & 0.5173 & 0.4160 & 0.5653 & 0.5547 & 0.5227 & 0.6667 & 0.5653 & 0.5493 & \textbf{0.6800} \\
96 & Rock & \textbf{0.9000} & 0.8200 & 0.8800 & 0.8600 & \textbf{0.9000} & 0.6200 & 0.6800 & 0.6800 & 0.7600 & 0.8600 & 0.4600 & 0.8600 & 0.6400 & 0.8400 & \textbf{0.9000} & 0.8600 & 0.8600 & 0.8600 & \textbf{0.9000} & 0.8400 & \textbf{0.9000} \\
97 & Scre. & 0.4613 & 0.4853 & 0.5360 & 0.5600 & 0.5227 & \textbf{0.5840} & 0.4053 & 0.3440 & 0.4160 & 0.4720 & 0.4667 & 0.3920 & 0.4347 & 0.3813 & 0.5680 & 0.4960 & 0.4667 & 0.5653 & 0.5413 & 0.4800 & 0.5813 \\
98 & Sem2. & 0.9350 & 0.8933 & 0.9267 & 0.9500 & 0.9600 & 0.9200 & 0.8600 & 0.7883 & 0.8600 & 0.9250 & 0.8567 & 0.8283 & 0.7483 & 0.8883 & 0.9250 & 0.8850 & 0.9350 & 0.9600 & 0.9317 & 0.9583 & \textbf{0.9650} \\
99 & Semg. & 0.6289 & 0.6978 & 0.6956 & 0.7733 & 0.8778 & 0.7022 & 0.6156 & 0.3000 & 0.3022 & 0.6422 & 0.5689 & 0.3222 & 0.4222 & 0.4733 & 0.8400 & 0.6556 & \textbf{0.9067} & 0.8422 & 0.8400 & 0.8356 & 0.8667 \\
100 & Seh2. & 0.8889 & 0.8689 & 0.8511 & 0.9022 & 0.9156 & 0.8956 & 0.7444 & 0.2689 & 0.3822 & 0.8800 & 0.7378 & 0.5489 & 0.6444 & 0.8556 & 0.9067 & 0.8711 & 0.9422 & 0.9333 & 0.9089 & \textbf{0.9431} & 0.9267 \\
101 & ShaZ. & 0.9200 & 0.9400 & 0.9000 & 0.9400 & 0.9400 & 0.9400 & 0.8800 & 0.8400 & 0.7400 & 0.9600 & 0.9000 & 0.5400 & 0.8800 & 0.8600 & 0.9000 & 0.9600 & \textbf{0.9800} & 0.9400 & 0.9400 & 0.9600 & 0.9400 \\
102 & Sham. & \textbf{1.0000} & \textbf{1.0000} & \textbf{1.0000} & \textbf{1.0000} & \textbf{1.0000} & 0.9167 & 0.9833 & 0.6000 & 0.4833 & \textbf{1.0000} & \textbf{1.0000} & 0.4833 & 0.9667 & 0.5444 & \textbf{1.0000} & \textbf{1.0000} & \textbf{1.0000} & \textbf{1.0000} & \textbf{1.0000} & 0.9778 & \textbf{1.0000} \\
103 & Shal. & 0.9050 & 0.9200 & 0.9233 & \textbf{0.9283} & 0.9200 & 0.8117 & 0.9033 & 0.3133 & 0.5783 & 0.8850 & 0.8100 & 0.7050 & 0.6717 & 0.7633 & 0.9083 & 0.9050 & 0.9050 & 0.9050 & 0.9050 & 0.9167 & 0.9217 \\
104 & Smals. & 0.8240 & 0.8293 & 0.8213 & 0.8187 & 0.8213 & 0.7760 & 0.7253 & 0.3973 & 0.5787 & 0.7733 & 0.8293 & 0.5360 & 0.6800 & 0.5760 & 0.7840 & 0.7280 & 0.7893 & \textbf{0.8427} & 0.7867 & 0.8133 & 0.8347 \\
105 & Smoe. & 0.9733 & 0.9600 & \textbf{0.9933} & 0.9800 & 0.9867 & 0.9867 & 0.9400 & 0.9467 & 0.9533 & \textbf{0.9933} & \textbf{0.9933} & 0.9000 & 0.7200 & 0.8467 & 0.9800 & \textbf{0.9933} & \textbf{0.9933} & 0.9800 & 0.9867 & 0.9667 & 0.9867 \\
106 & Son1. & 0.9235 & 0.8968 & 0.9068 & 0.8869 & 0.9052 & 0.8286 & 0.8802 & 0.7937 & 0.6789 & 0.7953 & \textbf{0.9434} & 0.7072 & 0.7754 & 0.6705 & 0.8120 & 0.8985 & 0.7987 & 0.9368 & 0.8952 & 0.9235 & 0.9418 \\
107 & Son2. & 0.9234 & 0.9224 & 0.8940 & 0.9323 & 0.9244 & 0.8846 & 0.8709 & 0.8069 & 0.8237 & 0.8395 & 0.8972 & 0.8636 & 0.9014 & 0.8279 & 0.8877 & 0.9318 & 0.8930 & 0.8909 & 0.9307 & 0.9360 & \textbf{0.9559} \\
108 & Stas. & 0.9808 & 0.9826 & 0.9826 & 0.9814 & 0.9826 & 0.9780 & 0.9701 & 0.9216 & 0.9303 & 0.9749 & 0.9750 & 0.9618 & 0.8568 & 0.9596 & 0.9781 & 0.9800 & 0.9771 & 0.9815 & 0.9790 & 0.9826 & \textbf{0.9834} \\
109 & Stry. & 0.9811 & \textbf{0.9838} & 0.9784 & 0.9757 & 0.9784 & 0.9757 & 0.9595 & 0.8541 & 0.9703 & \textbf{0.9838} & 0.9757 & 0.9595 & 0.8189 & 0.9703 & 0.9730 & 0.9703 & 0.9649 & 0.9811 & 0.9757 & 0.9595 & 0.9784 \\
110 & Swef. & 0.9664 & 0.9664 & 0.9632 & 0.9760 & 0.9632 & 0.9536 & 0.9280 & 0.9376 & 0.8512 & 0.9504 & 0.9328 & 0.8912 & 0.8576 & 0.8992 & 0.9664 & 0.9680 & 0.9728 & \textbf{0.9776} & 0.9632 & 0.9744 & 0.9696 \\
111 & Syms. & 0.9739 & 0.9849 & 0.9809 & 0.9809 & 0.9759 & 0.9809 & 0.9678 & 0.8101 & 0.8452 & 0.9688 & 0.8995 & 0.8040 & 0.9578 & 0.8784 & 0.9849 & 0.9809 & 0.9889 & 0.9940 & 0.9809 & 0.9910 & \textbf{0.9950} \\
112 & Synl. & \textbf{1.0000} & 0.9800 & 0.9967 & 0.9967 & \textbf{1.0000} & \textbf{1.0000} & 0.9967 & 0.7900 & 0.9667 & \textbf{1.0000} & 0.9833 & 0.9533 & 0.9967 & 0.9667 & 0.9867 & 0.9900 & 0.9967 & \textbf{1.0000} & 0.9900 & 0.9967 & \textbf{1.0000} \\
113 & Toe1. & 0.9561 & 0.9561 & 0.9518 & 0.9561 & 0.9649 & 0.9605 & 0.9430 & 0.7237 & 0.6711 & 0.9430 & \textbf{0.9693} & 0.5789 & 0.9254 & 0.7719 & 0.9386 & 0.9649 & 0.9474 & 0.9561 & 0.9298 & \textbf{0.9693} & \textbf{0.9693} \\
114 & Toe2. & 0.9308 & 0.9154 & 0.9077 & 0.9154 & 0.9308 & 0.9231 & 0.9308 & 0.5154 & 0.7077 & 0.9385 & 0.8846 & 0.6462 & 0.9000 & 0.7692 & 0.9077 & 0.9308 & 0.9385 & \textbf{0.9538} & 0.9077 & 0.9154 & 0.9462 \\
115 & Trace & \textbf{1.0000} & \textbf{1.0000} & \textbf{1.0000} & \textbf{1.0000} & \textbf{1.0000} & \textbf{1.0000} & \textbf{1.0000} & \textbf{1.0000} & 0.8600 & \textbf{1.0000} & \textbf{1.0000} & 0.8400 & 0.9600 & \textbf{1.0000} & \textbf{1.0000} & \textbf{1.0000} & \textbf{1.0000} & \textbf{1.0000} & \textbf{1.0000} & \textbf{1.0000} & \textbf{1.0000} \\
116 & TwoG. & 0.9991 & 0.9982 & 0.9965 & 0.9982 & 0.9991 & 0.9982 & 0.9807 & 0.8595 & 0.7392 & 0.9535 & 0.9183 & 0.8648 & 0.7480 & 0.8824 & 0.9842 & 0.9956 & 0.9965 & \textbf{1.0000} & 0.9684 & 0.9921 & \textbf{1.0000} \\
117 & Twos. & \textbf{1.0000} & 0.9973 & \textbf{1.0000} & \textbf{1.0000} & \textbf{1.0000} & \textbf{1.0000} & 0.9998 & 0.9995 & 0.9618 & \textbf{1.0000} & \textbf{1.0000} & 0.9542 & 0.9908 & 0.9990 & \textbf{1.0000} & \textbf{1.0000} & \textbf{1.0000} & \textbf{1.0000} & \textbf{1.0000} & \textbf{1.0000} & \textbf{1.0000} \\
118 & UMD & 0.9931 & 0.9931 & \textbf{1.0000} & 0.9931 & 0.9931 & 0.9931 & \textbf{1.0000} & 0.9722 & \textbf{1.0000} & 0.9861 & 0.9931 & 0.9653 & 0.8958 & 0.9653 & 0.9931 & 0.9931 & 0.9931 & \textbf{1.0000} & 0.9931 & 0.9931 & 0.9931 \\
119 & UWal. & 0.9765 & 0.9735 & \textbf{0.9802} & 0.9799 & 0.9752 & 0.9037 & 0.9372 & 0.9199 & 0.9238 & 0.9710 & 0.9129 & 0.9196 & 0.8939 & 0.9534 & 0.9704 & \textbf{0.9802} & 0.9788 & 0.9796 & 0.9738 & 0.9785 & 0.9799 \\
120 & UWaX. & 0.8565 & 0.8479 & 0.8554 & 0.8713 & 0.8559 & 0.8102 & 0.8035 & 0.7457 & 0.6965 & 0.8378 & 0.7783 & 0.7563 & 0.8138 & 0.7892 & 0.8691 & 0.8462 & 0.8864 & 0.8786 & 0.8733 & \textbf{0.8898} & 0.8811 \\
121 & UWaY. & 0.7786 & 0.7781 & 0.7641 & 0.8085 & 0.7806 & 0.7610 & 0.7264 & 0.6563 & 0.6119 & 0.7560 & 0.6857 & 0.6309 & 0.6924 & 0.6991 & 0.8051 & 0.7976 & 0.8185 & 0.8269 & 0.8009 & 0.8208 & \textbf{0.8275} \\
122 & UWaZ. & 0.7956 & 0.8009 & 0.7998 & 0.8222 & 0.7937 & 0.7803 & 0.7688 & 0.6924 & 0.6189 & 0.7892 & 0.7415 & 0.6454 & 0.7367 & 0.7125 & 0.8130 & 0.8071 & 0.8236 & 0.8361 & 0.8093 & 0.8283 & \textbf{0.8370} \\
123 & Wafer & 0.9981 & 0.9994 & \textbf{1.0000} & 0.9997 & \textbf{1.0000} & 0.9982 & 0.9966 & 0.9943 & 0.9955 & 0.9959 & 0.9989 & 0.9951 & 0.9927 & 0.9942 & 0.9989 & 0.9984 & 0.9989 & 0.9997 & \textbf{1.0000} & \textbf{1.0000} & \textbf{1.0000} \\
124 & Wine & 0.8148 & 0.8333 & 0.8704 & 0.9059 & 0.8704 & 0.6111 & 0.8519 & 0.5000 & 0.7963 & 0.6481 & 0.6852 & \textbf{0.9259} & 0.5000 & 0.5741 & 0.7222 & 0.5926 & 0.6296 & 0.7963 & 0.8889 & 0.8889 & 0.9074 \\
125 & Wors. & 0.7508 & 0.7524 & 0.7492 & 0.7774 & 0.7555 & 0.6536 & 0.6818 & 0.5470 & 0.5188 & 0.7429 & 0.5658 & 0.5549 & 0.5690 & 0.6066 & 0.7038 & 0.7555 & \textbf{0.7837} & 0.7492 & 0.7476 & 0.7696 & 0.7492 \\
126 & Worms & 0.7273 & 0.7532 & 0.6883 & 0.7792 & 0.7792 & 0.7532 & 0.7273 & 0.4935 & 0.3247 & 0.7792 & 0.7662 & 0.4545 & 0.5714 & 0.5584 & 0.7143 & 0.7532 & 0.7532 & 0.7922 & 0.7662 & 0.7792 & \textbf{0.8182} \\
127 & Wors. & 0.8182 & 0.7792 & 0.7662 & 0.7792 & 0.8182 & 0.7922 & 0.8312 & 0.5195 & 0.4416 & 0.8052 & 0.8052 & 0.5974 & 0.7792 & 0.6234 & 0.7273 & 0.7792 & 0.8052 & \textbf{0.8831} & 0.7143 & 0.7922 & 0.8442 \\
128 & Yoga & 0.9177 & 0.9050 & 0.9277 & 0.9290 & \textbf{0.9347} & 0.8997 & 0.8833 & 0.7963 & 0.7843 & 0.8860 & 0.8573 & 0.7927 & 0.6573 & 0.7833 & 0.8497 & 0.8910 & 0.8907 & 0.9197 & 0.8687 & 0.8813 & 0.9127 \\ \hline
\multicolumn{2}{c}{Avg. Rank} & 7.62 & 7.75 & 7.38 & 6.27 & 5.16 & 9.41 & 12.65 & 18.19 & 17.11 & 9.20 & 13.13 & 17.06 & 17.58 & 16.74 & 10.26 & 9.31 & 8.07 & 4.47 & 7.90 & 6.18 & \textbf{3.01} \\
\multicolumn{2}{c}{Win} & 18 & 13 & 20 & 22 & 31 & 13 & 8 & 6 & 5 & 14 & 14 & 4 & 2 & 2 & 15 & 11 & 21 & 31 & 23 & 29 & \textbf{48} \\ \bottomrule
% \end{tabular}
\end{longtable}
% \end{adjustbox}
\normalsize % 恢复正常字号
\setlength{\tabcolsep}{6pt} % 如果之前改了列间距，建议也恢复默认（通常是 6pt）

\section{Details of results on 30  additional datasets}\label{appendC}

R2R, together with a Random Forest (RF) classifier applied directly to raw time series, and ten pre-trained TSFMs without any parameter fine-tuning, are evaluated for zero-shot feature extraction on 30 additional time series datasets. Specifically, embeddings extracted from each pre-trained TSFM are used as inputs to a Random Forest classifier for classification evaluation. The detailed test accuracy results are reported in Table~\ref{tab:append-30-set}.
Notably, TS-RAG~\cite{ningts2025} is originally designed for zero-shot time series forecasting and achieves strong performance when combined with the pre-trained Chronos~\cite{ansari2024chronos} model. In this work, we adapt TS-RAG for zero-shot feature-extraction classification using Chronos, in which the knowledge base used for retrieval is constructed consistently within R2R.

% \usepackage{booktabs}
\begin{table}[h]
\centering
\caption{Detailed test classification accuracy results on 30 additional time series datasets under the zero-shot feature extraction setting. The best results are in \textbf{bold}.}
\label{tab:append-30-set}
\resizebox{\textwidth}{!}{
\begin{tabular}{@{}cl|ccccclccccc|c@{}}
\toprule
ID & Dataset name & RF & GPT4TS & Moment & Chronos & Moirai & UniTS & NuTime & Mantis & UniShape & TS-RAG & TSPulse & R2R \\ \midrule
1 & AconityMINIPrinterLarge\_eq & 0.9096 & 0.6596 & 0.8660 & 0.9265 & 0.8049 & 0.5819 & 0.8438 & 0.8710 & 0.8691 & 0.9206 & 0.9215 & \textbf{0.9417} \\
2 & AconityMINIPrinterSmall\_eq & 0.9486 & 0.7158 & 0.9589 & 0.9315 & 0.8562 & 0.6575 & 0.8562 & 0.9315 & 0.9486 & 0.9555 & 0.9692 & \textbf{0.9863} \\
3 & AllGestureWiimoteX\_eq & 0.5286 & 0.4186 & 0.5343 & 0.4100 & 0.2929 & 0.0800 & 0.5414 & 0.4943 & 0.5343 & 0.5786 & 0.5341 & \textbf{0.6571} \\
4 & AllGestureWiimoteY\_eq & 0.5571 & 0.4157 & 0.5857 & 0.3829 & 0.3014 & 0.0971 & 0.5129 & 0.5086 & 0.5771 & 0.6200 & 0.5729 & \textbf{0.7000} \\
5 & AllGestureWiimoteZ\_eq & 0.4643 & 0.3871 & 0.5286 & 0.4343 & 0.2614 & 0.1029 & 0.4543 & 0.4757 & 0.5557 & 0.5700 & 0.4886 & \textbf{0.6943} \\
6 & AsphaltObstaclesUni\_eq & 0.7366 & 0.7417 & 0.7647 & 0.8159 & 0.4731 & 0.2379 & 0.7954 & 0.7980 & 0.8235 & \textbf{0.8568} & 0.8210 & 0.8261 \\
7 & AsphaltPavementTypeUni\_eq & 0.7794 & 0.7992 & 0.7670 & 0.6960 & 0.4830 & 0.3598 & 0.8239 & 0.7367 & 0.7860 & 0.7045 & 0.6894 & \textbf{0.8636} \\
8 & AsphaltRegularityUni\_eq & 0.9228 & 0.9414 & 0.9174 & 0.8935 & 0.6591 & 0.4807 & 0.9534 & 0.9534 & 0.9627 & 0.8961 & 0.9048 & \textbf{0.9694} \\
9 & Colposcopy & 0.4059 & 0.4059 & 0.3762 & \textbf{0.4257} & 0.3267 & 0.3168 & 0.4158 & \textbf{0.4257} & 0.3960 & 0.3564 & 0.3861 & 0.4158 \\
10 & Covid3Month\_disc & 0.6393 & 0.6066 & 0.5410 & 0.6885 & 0.6885 & 0.4754 & 0.6066 & 0.6721 & 0.5738 & \textbf{0.7049} & 0.6393 & 0.6721 \\
11 & DodgerLoopDay\_nmv & \textbf{0.6883} & 0.1818 & 0.3377 & 0.4286 & 0.4026 & 0.1948 & 0.4286 & 0.4286 & 0.5455 & 0.5714 & 0.5414 & 0.5714 \\
12 & DodgerLoopGame\_nmv & 0.7638 & 0.6063 & 0.7717 & 0.7165 & 0.8346 & 0.4803 & 0.6772 & 0.7044 & 0.6614 & 0.7795 & 0.7717 & \textbf{0.8740} \\
13 & DodgerLoopWeekend\_nmv & \textbf{0.9841} & 0.4841 & 0.9444 & 0.7540 & 0.7460 & 0.5476 & 0.9524 & 0.9562 & 0.9206 & 0.9365 & \textbf{0.9841} & 0.9762 \\
14 & ElectricDeviceDetection & 0.8758 & 0.8559 & 0.8559 & 0.8646 & \textbf{0.8766} & 0.8299 & 0.8697 & 0.8612 & 0.8607 & 0.8652 & 0.8615 & 0.8620 \\
15 & FloodModeling1\_disc & 0.8119 & 0.7475 & 0.7673 & 0.7871 & 0.7624 & 0.7030 & 0.8218 & 0.7475 & 0.7871 & 0.7970 & 0.8168 & \textbf{0.8812} \\
16 & FloodModeling2\_disc & 0.9353 & 0.9303 & 0.9254 & 0.9303 & 0.9254 & 0.9204 & 0.9303 & 0.9254 & 0.9254 & 0.9204 & 0.9254 & \textbf{0.9502} \\
17 & FloodModeling3\_disc & 0.7609 & 0.6957 & 0.6793 & 0.6630 & 0.6685 & 0.6304 & 0.7065 & 0.6576 & 0.6793 & 0.7065 & 0.7174 & \textbf{0.9022} \\
18 & GestureMidAirD1\_eq & 0.5154 & 0.3000 & 0.6231 & 0.4231 & 0.2846 & 0.0385 & 0.4846 & 0.5077 & 0.6615 & 0.7615 & 0.7000 & \textbf{0.7692} \\
19 & GestureMidAirD2\_eq & 0.4154 & 0.0923 & 0.6015 & 0.5769 & 0.2154 & 0.1000 & 0.4769 & 0.4923 & 0.6077 & \textbf{0.6846} & 0.6462 & \textbf{0.6846} \\
20 & GestureMidAirD3\_eq & 0.2231 & 0.1615 & 0.4308 & 0.2462 & 0.1923 & 0.0385 & 0.2538 & 0.2846 & 0.3923 & \textbf{0.5077} & 0.4846 & \textbf{0.5077} \\
21 & GesturePebbleZ1\_eq & 0.6570 & 0.6453 & 0.7500 & 0.6337 & 0.4767 & 0.1628 & 0.7442 & 0.8447 & 0.8488 & 0.8547 & 0.8198 & \textbf{0.9012} \\
22 & GesturePebbleZ2\_eq & 0.6646 & 0.5190 & 0.7785 & 0.6962 & 0.5823 & 0.1329 & 0.7848 & 0.8608 & 0.7722 & 0.7911 & 0.8101 & \textbf{0.9114} \\
23 & KeplerLightCurves & 0.7143 & 0.2982 & \textbf{0.9349} & 0.8897 & 0.8296 & 0.2682 & 0.7268 & 0.7945 & 0.8296 & 0.8296 & 0.7744 & 0.8521 \\
24 & MelbournePedestrian\_nmv & 0.8906 & 0.7861 & 0.6270 & 0.8185 & 0.7628 & 0.1035 & 0.8952 & 0.8323 & 0.8991 & 0.8711 & 0.8758 & \textbf{0.9560} \\
25 & PhoneHeartbeatSound & 0.5989 & 0.5824 & 0.6429 & \textbf{0.6978} & 0.6209 & 0.5165 & 0.6538 & 0.6923 & \textbf{0.6978} & 0.6648 & 0.6264 & 0.6538 \\
26 & PickupGestureWiimoteZ\_eq & 0.6800 & 0.5200 & 0.5200 & 0.6800 & 0.3400 & 0.0600 & 0.6800 & 0.7800 & 0.8000 & 0.7400 & 0.6700 & \textbf{0.8800} \\
27 & PLAID\_eq & 0.7933 & 0.4562 & 0.7263 & 0.7858 & 0.5605 & 0.1415 & 0.6648 & 0.6909 & 0.7393 & 0.7896 & 0.7349 & \textbf{0.8771} \\
28 & ShakeGestureWiimoteZ\_eq & 0.7800 & 0.7200 & 0.8400 & 0.8000 & 0.5600 & 0.1200 & 0.8800 & 0.8900 & 0.8600 & \textbf{0.9000} & 0.7800 & 0.8800 \\
29 & SharePriceIncrease & 0.6884 & \textbf{0.6915} & 0.6781 & 0.6801 & 0.6718 & 0.6387 & 0.6894 & 0.6822 & 0.6822 & 0.6863 & 0.6814 & \textbf{0.6915} \\
30 & Tools & 0.4552 & 0.4328 & 0.6418 & 0.7015 & 0.6119 & 0.2761 & 0.6269 & 0.6567 & 0.5896 & 0.6567 & 0.6189 & \textbf{0.7164} \\ \hline
\multicolumn{2}{c|}{Avg. Rank} & 5.80 & 9.00 & 6.53 & 6.33 & 9.00 & \multicolumn{1}{c}{11.87} & 5.63 & 5.80 & 5.20 & 3.97 & 5.37 & \textbf{1.73} \\
\multicolumn{2}{c|}{Win} & 2 & 1 & 1 & 2 & 1 & \multicolumn{1}{c}{0} & 0 & 1 & 1 & 5 & 1 & \textbf{21} \\ \bottomrule
\end{tabular}}
\end{table}

\section{Details of ablation results}\label{appendD}

The detailed ablation study results for R2R on feature- and parameter-transfer settings across all 128 UCR time series datasets are reported in Table~\ref{tab:append-aba-128ucr}.
In addition, Table~\ref{tab:append-kd-size-ucr18} presents an analysis of the constructed knowledge base in R2R, examining the impact of its size and different sampling strategies applied to the original 1.89M pre-training samples. These experiments are conducted on a selected subset of 18 UCR datasets. Following prior work~\cite{liu2026unishape}, the 18 UCR datasets are selected based on criteria such as dataset size, sequence length, application domain, and number of classes, ensuring diversity while keeping the computational cost of large-scale knowledge base evaluation manageable.

% \usepackage{booktabs}
% \usepackage{multirow}
% \begin{longtable}[]
% \centering
% \caption{Detailed test classification accuracy results of the ablation study on 128 UCR time series datasets.}
% \label{tab:append-aba-128ucr}
% \begin{tabular}{@{}clcccccccc@{}}
% \toprule
% \multirow{3}{*}{ID} & \multirow{3}{*}{Dataset Name} & \multirow{3}{*}{\begin{tabular}[c]{@{}c@{}}R2R-\\ UniShape\end{tabular}} & \multirow{3}{*}{\begin{tabular}[c]{@{}c@{}}w/o \\ Feature \\ transfer\end{tabular}} & \multicolumn{6}{c}{w/o Parameter transfer} \\ \cmidrule(l){5-10} 
%  &  &  &  & MiniRocket & MiniRocket & MiniRocket & Rocket & MultiRocket & UniShape \\
%  &  &  &  & with Retrieval & w/o Retrieval & w/o ReMGP & w/o Retrieval & w/o Retrieval & w/o Retrieval \\ \cmidrule(r){1-4}

\normalsize
\fontsize{6.5pt}{5pt}\selectfont  % 第一个参数字号，第二个参数行距
\setlength{\tabcolsep}{2.0pt}
\begin{longtable}[h]{@{}cl|c|c|cccccc@{}}

\caption{Detailed test classification accuracy results of the ablation study on 128 UCR time series datasets. The best results are in \textbf{bold}.}
\label{tab:append-aba-128ucr} \\

\toprule
\multirow{3}{*}{ID} & \multirow{3}{*}{Dataset name} & \multirow{3}{*}{\makecell{R2R+\\UniShape}} & \multirow{3}{*}{\makecell{w/o \\ Feature \\ transfer}} & \multicolumn{6}{c}{w/o Parameter transfer} \\ 
\cmidrule(l){5-10} 
 & & & & MiniRocket & MiniRocket & MiniRocket & Rocket & MultiRocket & UniShape \\
 & & & & with Retrieval & w/o Retrieval & w/o ReMGP & w/o Retrieval & w/o Retrieval & w/o Retrieval \\ 
\midrule
\endfirsthead

% \toprule
% \multicolumn{10}{c}{\textit{Table \ref{tab:append-aba-128ucr} Continued from previous page}} \\
\toprule
\multirow{3}{*}{ID} & \multirow{3}{*}{Dataset name} & \multirow{3}{*}{\makecell{R2R-\\UniShape}} & \multirow{3}{*}{\makecell{w/o \\ Feature \\ transfer}} & \multicolumn{6}{c}{w/o Parameter transfer} \\ 
\cmidrule(l){5-10} 
 & & & & MiniRocket & MiniRocket & MiniRocket & Rocket & MultiRocket & UniShape \\
 & & & & with Retrieval & w/o Retrieval & w/o ReMGP & w/o Retrieval & w/o Retrieval & w/o Retrieval \\ 
\midrule
\endhead

\midrule
\multicolumn{10}{r}{\textit{Continued on next page}} \\
% \footlow
\endfoot

% \bottomrule
\endlastfoot
 
1 & ACSF1 & \textbf{0.9300} & 0.8000 & 0.9100 & 0.9000 & 0.8900 & 0.6800 & 0.8700 & 0.7500 \\
2 & Adiac & \textbf{0.8414} & 0.8289 & 0.7612 & 0.7570 & 0.7428 & 0.6880 & 0.8338 & 0.8312 \\
3 & AllGestureWiimoteX & \textbf{0.7743} & \textbf{0.7743} & 0.6714 & 0.6657 & 0.6186 & 0.5129 & 0.6600 & 0.7357 \\
4 & AllGestureWiimoteY & \textbf{0.8086} & 0.7929 & 0.7271 & 0.7257 & 0.6400 & 0.6600 & 0.7300 & 0.7729 \\
5 & AllGestureWiimoteZ & \textbf{0.7657} & 0.7429 & 0.6857 & 0.6643 & 0.6143 & 0.7300 & 0.6986 & 0.7400 \\
6 & ArrowHead & \textbf{0.9257} & 0.8857 & 0.8986 & 0.8914 & 0.8800 & 0.8229 & 0.8571 & 0.8571 \\
7 & Beef & 0.8333 & \textbf{0.8667} & 0.7500 & 0.6667 & 0.8333 & 0.5000 & 0.7667 & 0.7667 \\
8 & BeetleFly & \textbf{0.9500} & 0.8500 & \textbf{0.9500} & \textbf{0.9500} & 0.9000 & 0.9000 & 0.9000 & \textbf{0.9500} \\
9 & BirdChicken & 0.9000 & \textbf{0.9500} & 0.9000 & 0.9000 & 0.9000 & 0.9000 & \textbf{0.9500} & 0.9000 \\
10 & BME & \textbf{1.0000} & \textbf{1.0000} & 0.9767 & 0.9800 & 0.9733 & 0.9933 & \textbf{1.0000} & \textbf{1.0000} \\
11 & Car & 0.9167 & \textbf{0.9333} & 0.9197 & 0.9167 & 0.9167 & 0.8667 & 0.8500 & 0.8833 \\
12 & CBF & \textbf{1.0000} & \textbf{1.0000} & 0.9989 & 0.9989 & 0.9967 & 0.9889 & 0.9933 & \textbf{1.0000} \\
13 & Chinatown & \textbf{0.9855} & 0.9565 & 0.9739 & 0.9768 & 0.9723 & 0.9623 & 0.9565 & 0.9449 \\
14 & ChlorineConcentration & \textbf{0.8635} & 0.8539 & 0.6764 & 0.6641 & 0.6146 & 0.6401 & 0.7063 & 0.7456 \\
15 & CinCECGTorso & \textbf{0.9732} & 0.9500 & 0.8739 & 0.8667 & 0.7761 & 0.8420 & 0.9623 & 0.8717 \\
16 & Coffee & \textbf{1.0000} & \textbf{1.0000} & \textbf{1.0000} & \textbf{1.0000} & \textbf{1.0000} & \textbf{1.0000} & \textbf{1.0000} & \textbf{1.0000} \\
17 & Computers & \textbf{0.7800} & 0.7680 & 0.7410 & 0.7360 & 0.7340 & 0.6320 & 0.7280 & 0.7600 \\
18 & CricketX & \textbf{0.8308} & 0.8282 & 0.8126 & 0.8026 & 0.7949 & 0.7769 & 0.7846 & 0.7974 \\
19 & CricketY & \textbf{0.8487} & 0.8333 & 0.8308 & 0.8205 & 0.8077 & 0.8205 & 0.8154 & 0.8179 \\
20 & CricketZ & \textbf{0.8462} & 0.8436 & 0.7667 & 0.7282 & 0.6692 & 0.8128 & 0.6821 & 0.8410 \\
21 & Crop & \textbf{0.7839} & 0.7725 & 0.7776 & 0.7679 & 0.7673 & 0.7784 & 0.7804 & 0.7623 \\
22 & DiatomSizeReduction & \textbf{0.9739} & 0.9706 & 0.9281 & 0.9183 & 0.8791 & 0.9542 & 0.9575 & 0.9183 \\
23 & DistalPhalanxOutlineAgeGroup & \textbf{0.7914} & 0.7698 & \textbf{0.7914} & 0.7626 & 0.7410 & 0.7410 & 0.7698 & 0.7482 \\
24 & DistalPhalanxOutlineCorrect & 0.7862 & 0.7899 & 0.7981 & 0.7862 & 0.7717 & 0.7681 & \textbf{0.8080} & 0.7862 \\
25 & DistalPhalanxTW & 0.7050 & \textbf{0.7194} & 0.6863 & 0.6763 & 0.6735 & 0.6403 & 0.6906 & 0.6763 \\
26 & DodgerLoopDay & \textbf{0.7250} & 0.6250 & 0.6695 & 0.6625 & 0.7125 & 0.6125 & 0.6250 & 0.6625 \\
27 & DodgerLoopGame & \textbf{0.9420} & 0.9203 & 0.8879 & 0.8841 & 0.8813 & 0.9275 & 0.9203 & 0.9203 \\
28 & DodgerLoopWeekend & \textbf{0.9855} & \textbf{0.9855} & 0.9783 & 0.9783 & 0.9783 & 0.9565 & 0.9783 & 0.9783 \\
29 & Earthquakes & 0.7770 & 0.7554 & 0.7422 & 0.7050 & 0.7022 & 0.7194 & 0.7554 & \textbf{0.7986} \\
30 & ECG200 & \textbf{0.9400} & 0.9200 & 0.9300 & 0.9300 & 0.9200 & 0.9100 & 0.9200 & 0.8900 \\
31 & ECG5000 & 0.9462 & \textbf{0.9498} & 0.9353 & 0.9322 & 0.9317 & 0.9464 & 0.9453 & 0.9402 \\
32 & ECGFiveDays & \textbf{1.0000} & \textbf{1.0000} & 0.9844 & 0.9826 & 0.9919 & 0.9849 & 0.9954 & 0.9373 \\
33 & ElectricDevices & 0.7567 & 0.7248 & 0.7339 & 0.7236 & 0.7145 & 0.7082 & 0.7485 & \textbf{0.7712} \\
34 & EOGHorizontalSignal & \textbf{0.6796} & \textbf{0.6796} & 0.6133 & 0.6105 & 0.6022 & 0.5884 & 0.6105 & 0.6105 \\
35 & EOGVerticalSignal & 0.5442 & 0.5193 & 0.4558 & 0.4337 & 0.4254 & \textbf{0.5580} & 0.5138 & 0.5276 \\
36 & EthanolLevel & 0.8140 & \textbf{0.8500} & 0.6020 & 0.5040 & 0.4440 & 0.5480 & 0.6220 & 0.5680 \\
37 & FaceAll & 0.7852 & \textbf{0.8994} & 0.7721 & 0.7556 & 0.7504 & 0.8728 & 0.7385 & 0.7373 \\
38 & FaceFour & \textbf{0.9773} & 0.9545 & 0.9177 & 0.9091 & 0.9091 & 0.8068 & 0.9318 & 0.9432 \\
39 & FacesUCR & 0.9566 & \textbf{0.9620} & 0.9351 & 0.9346 & 0.9298 & 0.9259 & 0.9429 & 0.9054 \\
40 & FiftyWords & \textbf{0.8462} & \textbf{0.8462} & 0.8264 & 0.8154 & 0.8142 & 0.7780 & 0.8374 & 0.8154 \\
41 & Fish & 0.9829 & \textbf{0.9886} & 0.9429 & 0.9314 & 0.8686 & 0.9086 & 0.9714 & 0.9543 \\
42 & FordA & \textbf{0.9644} & 0.9629 & 0.9371 & 0.9356 & 0.9500 & 0.9341 & 0.9553 & 0.9402 \\
43 & FordB & \textbf{0.8704} & 0.8605 & 0.7952 & 0.7901 & 0.7914 & 0.7975 & 0.8395 & 0.8123 \\
44 & FreezerRegularTrain & \textbf{0.9993} & 0.9972 & 0.9877 & 0.9895 & 0.9816 & 0.9975 & 0.9982 & 0.9940 \\
45 & FreezerSmallTrain & 0.9846 & 0.9828 & 0.9465 & 0.9396 & 0.9282 & 0.8912 & 0.9554 & \textbf{0.9958} \\
46 & Fungi & \textbf{1.0000} & \textbf{1.0000} & 0.9624 & 0.9624 & 0.9624 & 0.9570 & 0.9409 & 0.9677 \\
47 & GestureMidAirD1 & \textbf{0.8077} & 0.8000 & 0.7200 & 0.7000 & 0.7000 & 0.7308 & 0.7385 & 0.7538 \\
48 & GestureMidAirD2 & \textbf{0.7231} & \textbf{0.7231} & 0.6400 & 0.6308 & 0.5923 & 0.6077 & 0.5923 & 0.6615 \\
49 & GestureMidAirD3 & \textbf{0.5385} & 0.4692 & 0.4738 & 0.4692 & 0.5154 & 0.4231 & 0.4615 & 0.4462 \\
50 & GesturePebbleZ1 & 0.9419 & \textbf{0.9477} & 0.8953 & 0.8721 & 0.8895 & 0.8023 & 0.9186 & 0.8663 \\
51 & GesturePebbleZ2 & 0.9557 & \textbf{0.9747} & 0.8328 & 0.8291 & 0.7975 & 0.8544 & 0.8797 & 0.8861 \\
52 & GunPoint & \textbf{1.0000} & \textbf{1.0000} & 0.9933 & 0.9867 & 0.9933 & 0.9933 & 0.9933 & 0.9800 \\
53 & GunPointAgeSpan & \textbf{1.0000} & 0.9968 & 0.9715 & 0.9715 & 0.9620 & 0.9905 & \textbf{1.0000} & 0.9968 \\
54 & GunPointMaleVersusFemale & \textbf{1.0000} & \textbf{1.0000} & \textbf{1.0000} & \textbf{1.0000} & \textbf{1.0000} & 0.9968 & \textbf{1.0000} & 0.9968 \\
55 & GunPointOldVersusYoung & \textbf{1.0000} & \textbf{1.0000} & \textbf{1.0000} & \textbf{1.0000} & \textbf{1.0000} & 0.9937 & \textbf{1.0000} & \textbf{1.0000} \\
56 & Ham & 0.7333 & 0.7619 & 0.6486 & 0.6476 & 0.6476 & \textbf{0.7810} & 0.6381 & 0.5810 \\
57 & HandOutlines & \textbf{0.9703} & 0.9649 & 0.9316 & 0.9270 & 0.9189 & 0.9135 & 0.9324 & 0.9486 \\
58 & Haptics & \textbf{0.5747} & 0.5682 & 0.5487 & 0.5487 & 0.5714 & 0.5519 & 0.5584 & 0.5649 \\
59 & Herring & 0.7656 & \textbf{0.7969} & 0.6406 & 0.5781 & 0.6406 & 0.6250 & 0.6250 & 0.6406 \\
60 & HouseTwenty & \textbf{0.9916} & 0.9832 & 0.9244 & 0.9244 & 0.9328 & 0.8992 & 0.9328 & 0.9580 \\
61 & InlineSkate & \textbf{0.5200} & 0.4909 & 0.4527 & 0.4509 & 0.4345 & 0.4473 & 0.4673 & 0.4218 \\
62 & InsectEPGRegularTrain & \textbf{1.0000} & \textbf{1.0000} & \textbf{1.0000} & \textbf{1.0000} & \textbf{1.0000} & 0.9960 & \textbf{1.0000} & \textbf{1.0000} \\
63 & InsectEPGSmallTrain & \textbf{1.0000} & \textbf{1.0000} & 0.9679 & 0.9679 & 0.9679 & 0.9598 & 0.9920 & \textbf{1.0000} \\
64 & InsectWingbeatSound & \textbf{0.6737} & 0.6682 & 0.6625 & 0.6535 & 0.6566 & 0.6273 & 0.6601 & 0.5889 \\
65 & ItalyPowerDemand & 0.9650 & 0.9640 & 0.9514 & 0.9475 & 0.9640 & 0.9660 & \textbf{0.9670} & 0.9359 \\
66 & LargeKitchenAppliances & 0.9200 & \textbf{0.9253} & 0.8253 & 0.8213 & 0.8160 & 0.8187 & 0.8400 & 0.8907 \\
67 & Lightning2 & 0.8361 & \textbf{0.8525} & 0.7705 & 0.7377 & 0.7869 & 0.8361 & 0.7377 & 0.8033 \\
68 & Lightning7 & 0.8356 & \textbf{0.8630} & 0.8082 & 0.7808 & 0.7534 & 0.7671 & 0.8493 & 0.7808 \\
69 & Mallat & \textbf{0.9898} & 0.9659 & 0.9284 & 0.9484 & 0.8601 & 0.9407 & 0.9134 & 0.9267 \\
70 & Meat & \textbf{0.9833} & 0.9500 & 0.9533 & 0.9500 & 0.9667 & 0.9667 & 0.9167 & 0.9667 \\
71 & MedicalImages & 0.8066 & 0.8039 & \textbf{0.8100} & 0.8053 & 0.8053 & 0.7855 & 0.7974 & 0.7829 \\
72 & MelbournePedestrian & \textbf{0.9751} & 0.9702 & 0.9580 & 0.9571 & 0.9633 & 0.9416 & 0.9416 & 0.9649 \\
73 & MiddlePhalanxOutlineAgeGroup & \textbf{0.6039} & 0.5714 & 0.3592 & 0.2273 & 0.5260 & 0.4221 & 0.3701 & 0.3831 \\
74 & MiddlePhalanxOutlineCorrect & \textbf{0.8763} & 0.8660 & 0.8482 & 0.8454 & 0.7801 & 0.8385 & 0.8557 & 0.8660 \\
75 & MiddlePhalanxTW & 0.5714 & \textbf{0.5974} & 0.5625 & 0.5584 & 0.5909 & 0.5779 & 0.5065 & 0.4935 \\
76 & MixedShapesRegularTrain & \textbf{0.9761} & 0.9753 & 0.9693 & 0.9518 & 0.9485 & 0.9530 & 0.9633 & 0.9546 \\
77 & MixedShapesSmallTrain & 0.9435 & \textbf{0.9509} & 0.8664 & 0.7918 & 0.7410 & 0.8169 & 0.9027 & 0.9423 \\
78 & MoteStrain & 0.9457 & 0.9305 & 0.9641 & 0.9593 & 0.9165 & 0.8355 & \textbf{0.9665} & 0.9073 \\
79 & NonInvasiveFetalECGThorax1 & \textbf{0.9598} & 0.9511 & 0.8843 & 0.8768 & 0.8224 & 0.9125 & 0.9247 & 0.9354 \\
80 & NonInvasiveFetalECGThorax2 & 0.9664 & 0.9517 & \textbf{0.9688} & 0.9644 & 0.9567 & 0.9608 & 0.9649 & 0.9511 \\
81 & OliveOil & \textbf{0.9667} & 0.8667 & \textbf{0.9667} & 0.9333 & \textbf{0.9667} & 0.8000 & \textbf{0.9667} & 0.9333 \\
82 & OSULeaf & \textbf{0.9917} & 0.9835 & 0.9298 & 0.9298 & 0.9174 & 0.8802 & 0.9132 & 0.9380 \\
83 & PhalangesOutlinesCorrect & 0.8485 & 0.8590 & 0.8380 & 0.8240 & 0.8228 & 0.8089 & \textbf{0.8660} & 0.8392 \\
84 & Phoneme & \textbf{0.3877} & 0.3729 & 0.3501 & 0.3075 & 0.2948 & 0.2590 & 0.3755 & 0.3539 \\
85 & PickupGestureWiimoteZ & \textbf{0.8800} & 0.8600 & 0.8700 & \textbf{0.8800} & 0.8600 & 0.7600 & 0.7600 & 0.7800 \\
86 & PigAirwayPressure & 0.4663 & 0.3510 & 0.5096 & \textbf{0.5192} & 0.4663 & 0.3462 & 0.4712 & 0.3798 \\
87 & PigArtPressure & \textbf{0.9952} & 0.9856 & 0.9519 & 0.9423 & 0.9279 & 0.8221 & 0.9183 & 0.8558 \\
88 & PigCVP & \textbf{0.8654} & 0.7500 & 0.8558 & 0.8221 & 0.7981 & 0.7788 & 0.7837 & 0.7404 \\
89 & PLAID & \textbf{0.9050} & 0.8939 & 0.5885 & 0.6965 & 0.7300 & 0.3408 & 0.4376 & 0.6127 \\
90 & Plane & \textbf{1.0000} & 0.9905 & \textbf{1.0000} & \textbf{1.0000} & \textbf{1.0000} & \textbf{1.0000} & \textbf{1.0000} & \textbf{1.0000} \\
91 & PowerCons & 0.9944 & \textbf{1.0000} & 0.9944 & 0.9889 & 0.9944 & 0.9667 & 0.9944 & 0.9778 \\
92 & ProximalPhalanxOutlineAgeGroup & \textbf{0.8585} & \textbf{0.8585} & 0.8146 & 0.8146 & 0.8293 & 0.8488 & 0.8488 & 0.8488 \\
93 & ProximalPhalanxOutlineCorrect & 0.9278 & \textbf{0.9347} & 0.8763 & 0.8454 & 0.9072 & 0.8900 & 0.9107 & 0.8832 \\
94 & ProximalPhalanxTW & \textbf{0.8390} & 0.8341 & 0.7951 & 0.8244 & 0.8195 & 0.7805 & 0.8146 & 0.7707 \\
95 & RefrigerationDevices & \textbf{0.6800} & 0.6667 & 0.5273 & 0.5200 & 0.4800 & 0.5413 & 0.5360 & 0.5387 \\
96 & Rock & \textbf{0.9000} & 0.8600 & 0.8600 & 0.8600 & 0.7600 & 0.8200 & 0.8400 & 0.8400 \\
97 & ScreenType & \textbf{0.5813} & 0.5653 & 0.5187 & 0.5173 & 0.4987 & 0.4293 & 0.4987 & 0.4987 \\
98 & SemgHandGenderCh2 & \textbf{0.9650} & 0.9600 & 0.9533 & 0.9500 & 0.9017 & 0.8383 & 0.9500 & 0.9250 \\
99 & SemgHandMovementCh2 & \textbf{0.8667} & 0.8422 & 0.7689 & 0.7467 & 0.6289 & 0.6178 & 0.7800 & 0.8356 \\
100 & SemgHandSubjectCh2 & 0.9267 & \textbf{0.9333} & 0.8778 & 0.8511 & 0.8267 & 0.7289 & 0.9067 & 0.8556 \\
101 & ShakeGestureWiimoteZ & 0.9400 & 0.9400 & 0.9600 & \textbf{0.9800} & 0.9400 & 0.9200 & 0.9400 & 0.8800 \\
102 & ShapeletSim & \textbf{1.0000} & \textbf{1.0000} & 0.9833 & 0.9833 & \textbf{1.0000} & 0.9944 & \textbf{1.0000} & \textbf{1.0000} \\
103 & ShapesAll & \textbf{0.9217} & 0.9050 & 0.9000 & 0.8883 & 0.9000 & 0.8733 & 0.9033 & 0.9117 \\
104 & SmallKitchenAppliances & 0.8347 & \textbf{0.8427} & 0.7867 & 0.8080 & 0.7707 & 0.7733 & 0.8187 & 0.8293 \\
105 & SmoothSubspace & \textbf{0.9867} & 0.9800 & 0.9133 & 0.9067 & 0.8867 & 0.9333 & 0.9333 & 0.9533 \\
106 & SonyAIBORobotSurface1 & 0.9418 & 0.9368 & 0.9368 & 0.9334 & 0.9567 & \textbf{0.9584} & 0.9301 & 0.9085 \\
107 & SonyAIBORobotSurface2 & \textbf{0.9559} & 0.8909 & 0.9433 & 0.9402 & 0.9339 & 0.9465 & 0.8909 & 0.9328 \\
108 & StarLightCurves & \textbf{0.9834} & 0.9815 & 0.9823 & 0.9811 & 0.9823 & 0.9786 & 0.9820 & 0.9809 \\
109 & Strawberry & 0.9784 & \textbf{0.9811} & 0.9595 & 0.9757 & 0.9027 & 0.9081 & 0.9676 & 0.9622 \\
110 & SwedishLeaf & 0.9696 & \textbf{0.9776} & 0.9492 & 0.9440 & 0.9264 & 0.9616 & 0.9728 & 0.9616 \\
111 & Symbols & \textbf{0.9950} & 0.9940 & 0.9859 & 0.9839 & 0.9809 & 0.9729 & 0.9829 & 0.9839 \\
112 & SyntheticControl & \textbf{1.0000} & \textbf{1.0000} & 0.9933 & 0.9933 & 0.9900 & 0.9967 & 0.9933 & 0.9833 \\
113 & ToeSegmentation1 & \textbf{0.9693} & 0.9561 & 0.9561 & 0.9518 & 0.9518 & 0.9605 & 0.9342 & 0.9649 \\
114 & ToeSegmentation2 & 0.9462 & \textbf{0.9538} & 0.9154 & 0.9154 & 0.8769 & 0.8615 & 0.9000 & 0.9231 \\
115 & Trace & \textbf{1.0000} & \textbf{1.0000} & \textbf{1.0000} & \textbf{1.0000} & \textbf{1.0000} & \textbf{1.0000} & \textbf{1.0000} & \textbf{1.0000} \\
116 & TwoLeadECG & \textbf{1.0000} & \textbf{1.0000} & 0.9798 & 0.9833 & 0.9728 & 0.9991 & 0.9991 & 0.9851 \\
117 & TwoPatterns & \textbf{1.0000} & \textbf{1.0000} & \textbf{1.0000} & 0.9998 & 0.9998 & \textbf{1.0000} & \textbf{1.0000} & 0.9998 \\
118 & UMD & 0.9931 & \textbf{1.0000} & 0.9444 & 0.9306 & 0.9275 & 0.8819 & 0.8681 & 0.9861 \\
119 & UWaveGestureLibraryAll & \textbf{0.9799} & 0.9796 & 0.9673 & 0.9648 & 0.9562 & 0.9724 & 0.9796 & 0.9648 \\
120 & UWaveGestureLibraryX & \textbf{0.8811} & 0.8786 & 0.8509 & 0.8492 & 0.8300 & 0.8437 & 0.8571 & 0.8663 \\
121 & UWaveGestureLibraryY & \textbf{0.8275} & 0.8269 & 0.7920 & 0.7842 & 0.7753 & 0.7753 & 0.7962 & 0.8074 \\
122 & UWaveGestureLibraryZ & \textbf{0.8370} & 0.8361 & 0.8093 & 0.8032 & 0.7856 & 0.7822 & 0.8138 & 0.8227 \\
123 & Wafer & \textbf{1.0000} & 0.9997 & \textbf{1.0000} & \textbf{1.0000} & 0.9985 & \textbf{1.0000} & \textbf{1.0000} & 0.9994 \\
124 & Wine & \textbf{0.9074} & 0.7963 & 0.8333 & 0.8148 & 0.8519 & 0.5000 & 0.6481 & 0.7407 \\
125 & WordSynonyms & 0.7492 & 0.7492 & 0.7273 & 0.7116 & 0.7194 & 0.7210 & \textbf{0.7539} & 0.7335 \\
126 & Worms & \textbf{0.8182} & 0.7922 & 0.7662 & 0.7403 & 0.7532 & 0.7403 & 0.7792 & 0.7792 \\
127 & WormsTwoClass & 0.8442 & \textbf{0.8831} & 0.7662 & 0.7662 & 0.7792 & 0.7792 & 0.8182 & 0.7922 \\
128 & Yoga & 0.9127 & \textbf{0.9197} & 0.8900 & 0.8810 & 0.8823 & 0.8570 & 0.8967 & 0.9087 \\ 
 \noalign{\vskip 2pt} \hline \noalign{\vskip 2pt}
\multicolumn{2}{c|}{Avg. Rank} & \textbf{1.48} & 2.45 & 4.14 & 5.09 & 5.59 & 5.91 & 4.06 & 4.52 \\
\multicolumn{2}{c|}{Win} & \textbf{86} & 48 & 13 & 11 & 8 & 8 & 18 & 13 \\ \bottomrule
% \end{tabular}
\end{longtable}
\normalsize % 恢复正常字号
\setlength{\tabcolsep}{6pt} % 如果之前改了列间距，建议也恢复默认（通常是 6pt）

% \usepackage{booktabs}
% \usepackage{multirow}
\begin{table}[h]
\centering
\caption{Detailed test classification accuracy results on 18 UCR time series datasets under varying knowledge base sizes and sampling strategies. The best results are in \textbf{bold}.}
\label{tab:append-kd-size-ucr18}
\resizebox{\textwidth}{!}{
\begin{tabular}{@{}cl|cccccc|c|c@{}}
\toprule
\multirow{2}{*}{ID} & \multirow{2}{*}{Dataset name} & \multicolumn{6}{c|}{Balanced} & Random & Unbalanced \\ \cmidrule(l){3-10} 
 &  & 50K & 100K & 200K & 500K & 1M & 1.89M & 100K & 100K \\ \cmidrule(r){1-10}
1 & ArrowHead & 0.8628 & \textbf{0.8857} & \textbf{0.8857} & 0.8800 & 0.8571 & 0.8685 & \textbf{0.8857} & 0.8628 \\
2 & CBF & \textbf{1.0000} & \textbf{1.0000} & \textbf{1.0000} & \textbf{1.0000} & \textbf{1.0000} & \textbf{1.0000} & \textbf{1.0000} & \textbf{1.0000} \\
3 & CricketX & 0.8077 & \textbf{0.8282} & 0.8205 & 0.8205 & 0.8205 & 0.8154 & 0.8203 & 0.7997 \\
4 & DistalPhalanxOutlineAgeGroup & 0.7698 & 0.7698 & 0.7626 & 0.7842 & 0.7914 & \textbf{0.8058} & 0.7698 & 0.7483 \\
5 & DistalPhalanxOutlineCorrect & 0.7826 & 0.7899 & 0.7754 & \textbf{0.8152} & 0.7826 & 0.7826 & 0.7863 & 0.7790 \\
6 & ECG5000 & 0.9445 & 0.9498 & \textbf{0.9500} & 0.9498 & 0.9498 & 0.9487 & 0.9474 & 0.9403 \\
7 & EOGVerticalSignal & 0.5082 & 0.5193 & 0.5165 & 0.5054 & 0.5054 & \textbf{0.5220} & 0.5082 & 0.4916 \\
8 & EthanolLevel & 0.8520 & 0.8500 & \textbf{0.8660} & 0.8260 & 0.8380 & 0.8520 & 0.8360 & 0.8280 \\
9 & Fish & \textbf{1.0000} & 0.9886 & \textbf{1.0000} & \textbf{1.0000} & \textbf{1.0000} & \textbf{1.0000} & \textbf{1.0000} & 0.9829 \\
10 & GunPoint & 0.9970 & \textbf{1.0000} & \textbf{1.0000} & 0.9934 & \textbf{1.0000} & 0.9934 & \textbf{1.0000} & 0.9934 \\
11 & InsectWingbeatSound & 0.6612 & 0.6682 & 0.6637 & \textbf{0.6773} & 0.6566 & 0.6587 & 0.6647 & 0.6531 \\
12 & ItalyPowerDemand & 0.9679 & 0.9640 & \textbf{0.9737} & 0.9689 & 0.9456 & 0.9640 & 0.9669 & 0.9611 \\
13 & MelbournePedestrian & 0.9698 & 0.9702 & 0.9690 & 0.9694 & \textbf{0.9718} & 0.9653 & 0.9702 & 0.9669 \\
14 & MiddlePhalanxTW & 0.5520 & \textbf{0.5974} & 0.5195 & 0.5650 & 0.5715 & 0.5260 & 0.5520 & 0.5440 \\
15 & MixedShapesRegularTrain & 0.9745 & \textbf{0.9753} & 0.9745 & 0.9741 & 0.9749 & 0.9737 & 0.9749 & 0.9704 \\
16 & OSULeaf & 0.9635 & \textbf{0.9835} & 0.9629 & 0.9753 & 0.9753 & 0.9753 & 0.9794 & 0.9711 \\
17 & Trace & \textbf{1.0000} & \textbf{1.0000} & \textbf{1.0000} & \textbf{1.0000} & \textbf{1.0000} & \textbf{1.0000} & \textbf{1.0000} & \textbf{1.0000} \\
18 & WordSynonyms & 0.7476 & 0.7492 & \textbf{0.7555} & \textbf{0.7555} & \textbf{0.7555} & 0.7414 & 0.7398 & 0.7398 \\ \hline
\multicolumn{2}{c|}{Avg. Rank} & 4.11 & \textbf{2.33} & 3.28 & 3.11 & 3.17 & 4.17 & 3.17 & 6.50 \\
\multicolumn{2}{c|}{Win} & 3 & 8 & \textbf{9} & 6 & 6 & 5 & 5 & 2 \\ \bottomrule
\end{tabular}}
\end{table}

\section{Details of knowledge-enhanced results}\label{appendE}

\begin{figure}[htbp]
\centering
\includegraphics[width=5.5in]{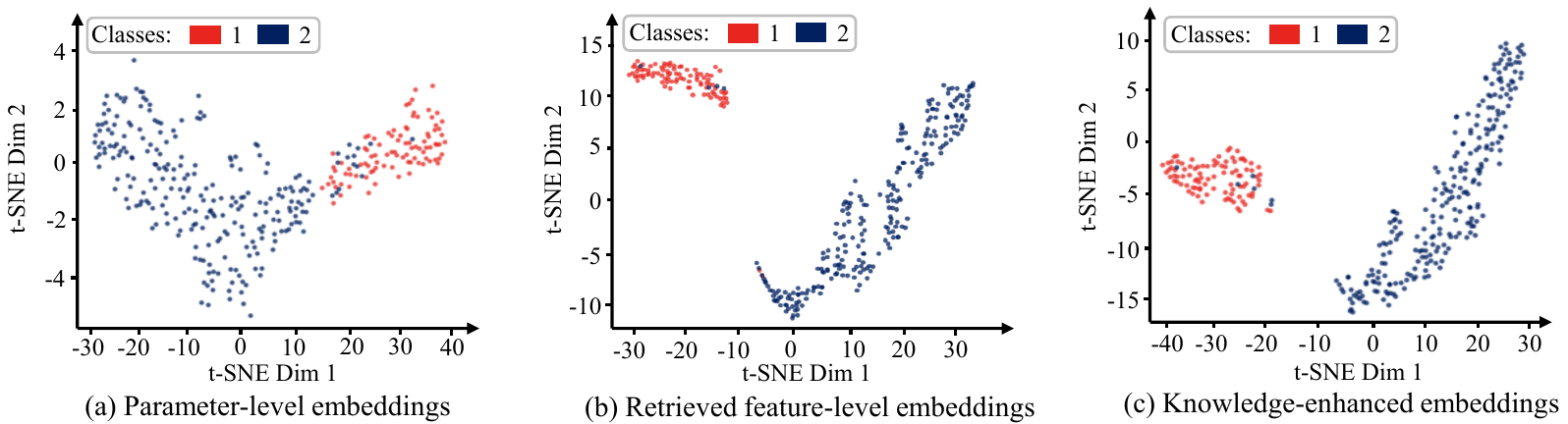}
\caption{t-SNE visualization on the \textit{Chinatown} dataset.
}
\label{fig_append_chinatown_tsne}
\end{figure}

\begin{figure}[htbp]
\centering
\includegraphics[width=5.4in]{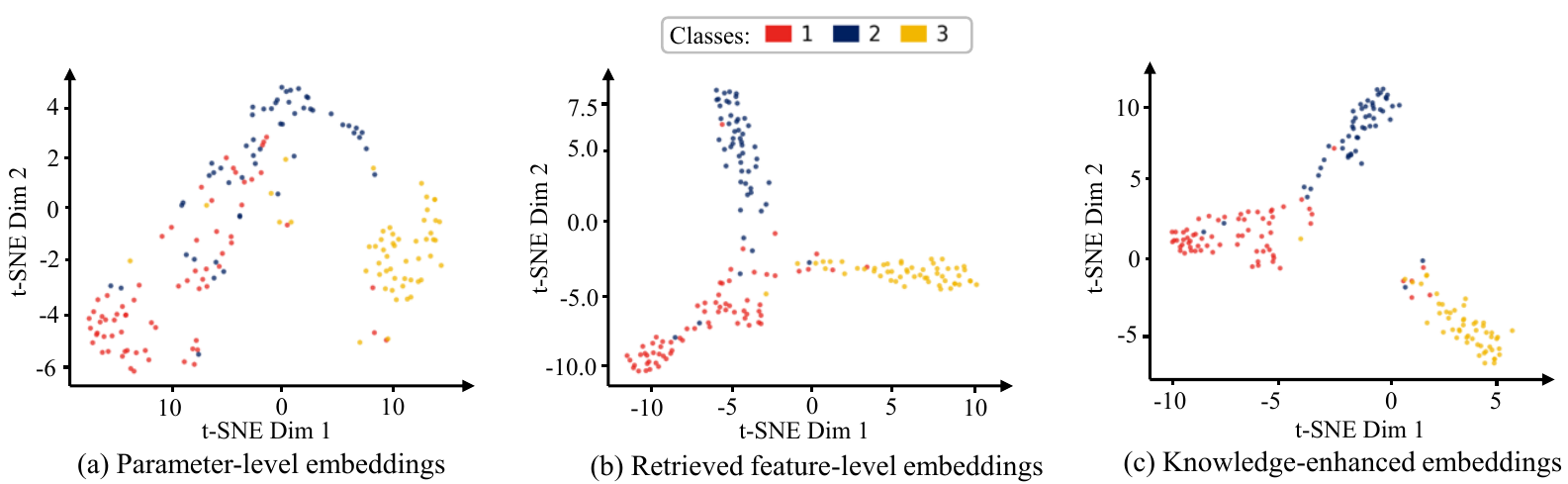}
\caption{t-SNE visualization on the \textit{ArrowHead} dataset.
}
\label{fig_append_arrow_tsne}
\end{figure}

\begin{figure}[htbp]
\centering
\includegraphics[width=5.5in]{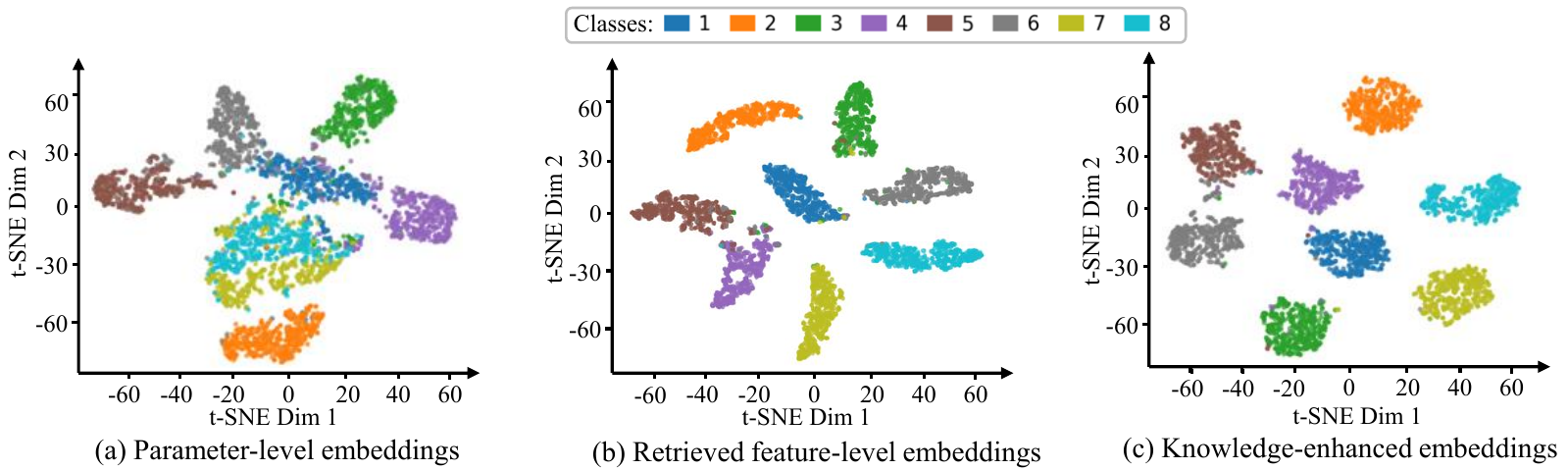}
\caption{t-SNE visualization on the \textit{UWaveGestureLibraryAll} dataset.
}
\label{fig_append_uwave_tsne}
\end{figure}

\begin{figure}[htbp]
\centering
\includegraphics[width=5.5in]{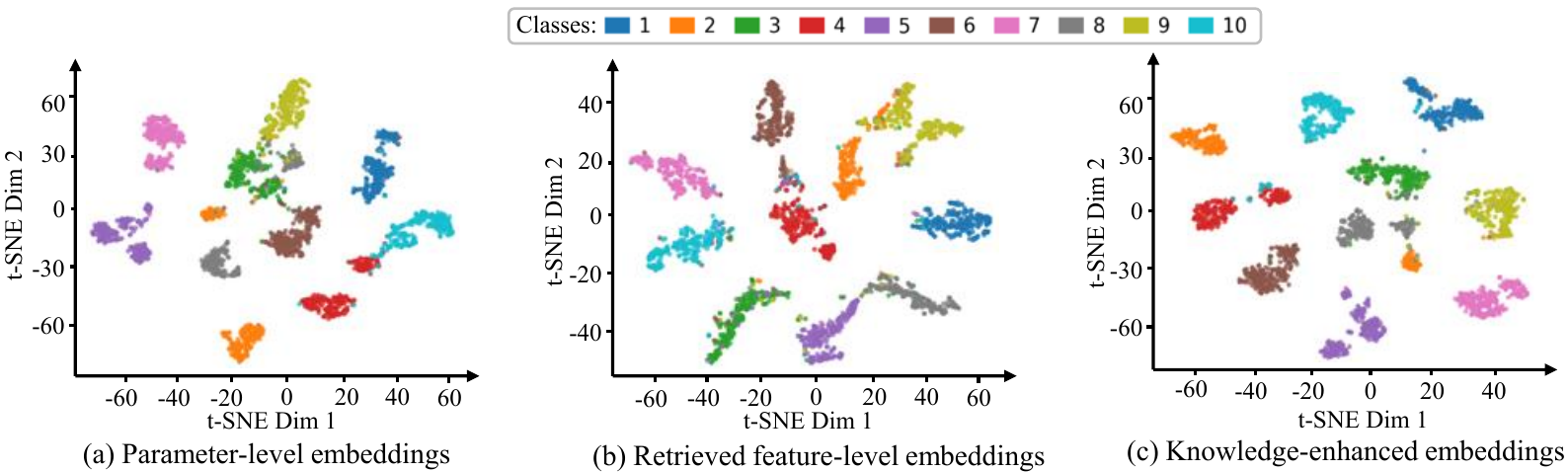}
\caption{t-SNE visualization on the \textit{MelbournePedestrian} dataset.
}
\label{fig_append_melb_tsne}
\end{figure}

We first present t-SNE visualizations~\cite{van2008visualizing} to examine the discriminative quality of embeddings learned by R2R across UCR time series datasets with varying domains and class cardinalities. Figures~\ref{fig_append_chinatown_tsne},~\ref{fig_append_arrow_tsne},~\ref{fig_append_uwave_tsne}, and~\ref{fig_append_melb_tsne} illustrate the results on four representative UCR datasets, namely \textit{Chinatown}, \textit{ArrowHead}, \textit{UWaveGestureLibraryAll}, and \textit{MelbournePedestrian}. These datasets differ in application domains and number of classes, enabling a comprehensive analysis. The results demonstrate that incorporating retrieved embeddings improves class separability compared to embeddings obtained solely via parameter fine-tuning.

Furthermore, we compare R2R with seven commonly used time series data augmentation methods on the same 18 UCR datasets used in Table~\ref{tab:append-kd-size-ucr18}. This experiment evaluates whether retrieval-based numerical embeddings can provide more effective knowledge transfer than conventional data augmentation strategies. Those seven time series data augmentation methods~\cite{wen2021time,iglesias2023data} are summarized as follows:

\begin{itemize}
\item \textbf{Jitter}: Adds random Gaussian noise to the original time series.
\item \textbf{Time warp}: Randomly distorts the temporal axis by varying the speed of the sequence.
\item \textbf{Magnitude warp}: Applies smooth, non-linear changes to the amplitude of the series.
\item \textbf{Window slice}: Extracts a subsegment via downsampling and resizes it back to the original length.
\item \textbf{Time flip}: Reverses the temporal order of the sequence.
\item \textbf{Permutation}: Divides the series into segments and randomly shuffles their order.
\item \textbf{Scaling}: Multiplies the entire series by a random scaling factor.
\end{itemize}

The detailed classification accuracy results of R2R and the seven augmentation methods on the 18 datasets are reported in Table~\ref{tab:append-18ucr-aug}. Additionally, Table~\ref{tab:append-continue-4set} compares R2R’s transfer paradigm with conventional fine-tuning-based transfer across datasets with different numbers of classes (2, 6, 10, and 26), under a continual fine-tuning setting.

% \usepackage{booktabs}
\begin{table}[h]
\centering
\caption{Detailed test classification accuracy on 18 UCR datasets compared with different time series data augmentation methods.  
The best results are in \textbf{bold}.}
\label{tab:append-18ucr-aug}
\resizebox{\textwidth}{!}{
\begin{tabular}{@{}cl|cccccccc|c@{}}
\toprule
ID & Dataset name & Raw & Jitter & Time warp & Magnitude warp & Window slice & Time flip & Permutation & Scaling & R2R \\ \midrule
1 & ArrowHead & 0.8857 & 0.8629 & 0.8457 & 0.8686 & 0.8629 & 0.8857 & 0.8400 & 0.8743 & \textbf{0.9257} \\
2 & CBF & \textbf{1.0000} & \textbf{1.0000} & \textbf{1.0000} & \textbf{1.0000} & \textbf{1.0000} & 0.8389 & \textbf{1.0000} & \textbf{1.0000} & \textbf{1.0000} \\
3 & CricketX & 0.8282 & 0.8154 & 0.8205 & 0.8128 & 0.8179 & \textbf{0.8308} & \textbf{0.8308} & 0.8179 & \textbf{0.8308} \\
4 & DistalPhalanxOutlineAgeGroup & 0.7698 & 0.7698 & 0.7698 & 0.7554 & 0.7842 & 0.7842 & 0.7482 & 0.7338 & \textbf{0.7914} \\
5 & DistalPhalanxOutlineCorrect & 0.7899 & 0.7935 & 0.7899 & \textbf{0.7971} & 0.7609 & 0.7609 & \textbf{0.7971} & 0.7935 & 0.7862 \\
6 & ECG5000 & \textbf{0.9498} & 0.9464 & 0.9387 & 0.9458 & 0.9444 & 0.9442 & 0.9464 & 0.9451 & 0.9462 \\
7 & EOGVerticalSignal & 0.5193 & 0.5387 & 0.5608 & 0.5276 & \textbf{0.5718} & 0.4613 & 0.5110 & 0.5110 & 0.5442 \\
8 & EthanolLevel & \textbf{0.8500} & 0.7520 & 0.7140 & 0.7260 & 0.8100 & 0.7580 & 0.7980 & 0.7560 & 0.8140 \\
9 & Fish & \textbf{0.9886} & 0.9486 & 0.9714 & 0.9657 & 0.9771 & 0.9771 & 0.9657 & 0.9714 & 0.9829 \\
10 & GunPoint & \textbf{1.0000} & \textbf{1.0000} & \textbf{1.0000} & 0.9867 & 0.9867 & 0.9933 & 0.9933 & 0.9867 & \textbf{1.0000} \\
11 & InsectWingbeatSound & 0.6682 & 0.6515 & 0.6399 & 0.6611 & 0.6283 & 0.6667 & 0.6540 & 0.6551 & \textbf{0.6737} \\
12 & ItalyPowerDemand & 0.9640 & \textbf{0.9660} & 0.9534 & 0.9621 & 0.9534 & 0.9621 & 0.9563 & 0.9611 & 0.9650 \\
13 & MelbournePedestrian & 0.9702 & 0.9610 & 0.9668 & 0.9668 & 0.9672 & 0.9631 & 0.9606 & 0.9639 & \textbf{0.9751} \\
14 & MiddlePhalanxTW & \textbf{0.5974} & 0.5649 & 0.5714 & 0.5260 & 0.5649 & 0.5714 & 0.5390 & 0.5779 & 0.5714 \\
15 & MixedShapesRegularTrain & 0.9753 & 0.9654 & 0.9699 & 0.9761 & \textbf{0.9773} & 0.9695 & 0.9736 & 0.9765 & 0.9761 \\
16 & OSULeaf & 0.9835 & 0.9545 & 0.9628 & 0.9669 & 0.9835 & 0.9711 & 0.9504 & 0.9669 & \textbf{0.9917} \\
17 & Trace & \textbf{1.0000} & \textbf{1.0000} & \textbf{1.0000} & \textbf{1.0000} & \textbf{1.0000} & \textbf{1.0000} & \textbf{1.0000} & \textbf{1.0000} & \textbf{1.0000} \\
18 & WordSynonyms & \textbf{0.7492} & 0.7241 & 0.7320 & 0.7414 & 0.7320 & 0.6771 & 0.7116 & 0.7288 & \textbf{0.7492} \\ \hline
\multicolumn{2}{c|}{Avg. Rank} & 2.39 & 5.11 & 5.06 & 4.89 & 4.33 & 5.06 & 5.50 & 4.83 & \textbf{2.00} \\
\multicolumn{2}{c|}{Win} & 8 & 4 & 3 & 3 & 4 & 2 & 4 & 2 & \textbf{10} \\ \bottomrule
\end{tabular}}
\end{table}

% \usepackage{booktabs}
% \usepackage{multirow}
\begin{table}[h]
\centering
\caption{Impact of continual fine-tuning for classification on another four group datasets selected from the additional 30 time series datasets. Acc$_{A_1}$: Initial accuracy on dataset $A$; Acc$_{A_2}$: Accuracy on dataset $A$ after continual training on dataset $B$; Forgetting: Acc$_{A_1}$ - Acc$_{A_2}$.}
\label{tab:append-continue-4set}
\resizebox{\textwidth}{!}{
\begin{tabular}{@{}ll|ccc|ccc@{}}
\toprule
\multicolumn{1}{c}{\multirow{2}{*}{Dataset   A}} & \multicolumn{1}{c|}{\multirow{2}{*}{Dataset B}} & \multicolumn{3}{c|}{Only Parameter   transfer} & \multicolumn{3}{c}{R2R transfer} \\ \cmidrule(l){3-8} 
\multicolumn{1}{c}{} & \multicolumn{1}{c|}{} & \multicolumn{1}{c}{Acc$_{A_1}$} & \multicolumn{1}{l}{Acc$_{A_2}$} & \multicolumn{1}{c|}{Forgetting} & \multicolumn{1}{l}{Acc$_{A_1}$} & \multicolumn{1}{l}{Acc$_{A_2}$} & \multicolumn{1}{l}{Forgetting} \\ \cmidrule(r){1-8}
AconityMINIPrinterSmall\_eq & AsphaltRegularityUni\_eq & 0.9795 & 0.8048 & 0.1747 & 0.9829 & 0.8253 & \textbf{0.1576} \\
Colposcopy & GesturePebbleZ1\_eq & 0.3564 & 0.2574 & 0.0990 & 0.3762 & 0.2970 & \textbf{0.0792} \\
AllGestureWiimoteY\_eq & AllGestureWiimoteZ\_eq & 0.7971 & 0.2829 & 0.5142 & 0.8071 & 0.3286 & \textbf{0.4785} \\
GestureMidAirD2\_eq & GestureMidAirD3\_eq & 0.6923 & 0.1769 & 0.5154 & 0.7000 & 0.2308 & \textbf{0.4692} \\ \bottomrule
\end{tabular}}
\end{table}

\section{Details of hyperparameter analysis results}\label{appendF}

Table~15 presents the classification performance of R2R under three distance metrics used for retrieval index construction, evaluated on the same 18 UCR datasets as in Table~\ref{tab:append-kd-size-ucr18}. We consider \textit{Inner Product (IP)}, \textit{Euclidean Distance (L2)}, and \textit{Cosine Similarity}. These metrics differ in both formulation and geometric properties. Specifically, IP computes similarity as $\langle \mathbf{x}, \mathbf{y} \rangle = \sum_{i=1}^{d} x_i y_i$, favoring vectors with large aligned magnitudes; L2 measures dissimilarity as $\|\mathbf{x} - \mathbf{y}\|_2^2 = \sum_{i=1}^{d} (x_i - y_i)^2$, capturing absolute geometric distance; while Cosine similarity evaluates angular similarity as $\frac{\langle \mathbf{x}, \mathbf{y} \rangle}{\|\mathbf{x}\|_2 \|\mathbf{y}\|_2}$, effectively normalizing vector magnitudes and focusing on directional alignment.
As shown in Table~\ref{tab:append-dist-rekin}, Cosine similarity achieves the best average rank, indicating its superiority in retrieving semantically aligned embeddings. Therefore, R2R adopts Cosine similarity as the default distance metric for nearest neighbor retrieval.

Tables~\ref{tab:append-nlist-18ucr} and~\ref{tab:append-topk-18ucr} further examine the influence of retrieval hyperparameters, including the index partition size $n_{\text{list}}$ and the number of retrieved neighbors $K$. The results demonstrate that $n_{\text{list}}=256$ and $K=20$ yield the best average rank, and are thus selected as default settings. Additionally, comparing $K=1$ and $K=20$ in Table~\ref{tab:append-topk-18ucr} shows that using only a single nearest neighbor significantly degrades performance, as it effectively removes the optimal transport-based weighted fusion mechanism. In contrast, aggregating multiple retrieved neighbors ($K=20$) improves both average rank and win rate, highlighting the effectiveness of multi-neighbor retrieval combined with optimal transport-based weighting.

% \usepackage{booktabs}
\begin{table}[h]
\centering
\caption{Detailed test classification accuracy results on 18 UCR time series datasets using different distance metrics within R2R.  
The best results are in \textbf{bold}.}
\label{tab:append-dist-rekin}
\begin{tabular}{@{}cl|ccc@{}}
\toprule
ID & Dataset name & IP & L2 & Cosine \\ \midrule
1 & ArrowHead & 0.8800 & 0.8743 & \textbf{0.9257} \\
2 & CBF & \textbf{1.0000} & \textbf{1.0000} & \textbf{1.0000} \\
3 & CricketX & 0.8257 & 0.8257 & \textbf{0.8308} \\
4 & DistalPhalanxOutlineAgeGroup & 0.7554 & 0.7339 & \textbf{0.7914} \\
5 & DistalPhalanxOutlineCorrect & 0.7754 & \textbf{0.8080} & 0.7862 \\
6 & ECG5000 & 0.9400 & 0.9420 & \textbf{0.9462} \\
7 & EOGVerticalSignal & 0.5332 & \textbf{0.5497} & 0.5442 \\
8 & EthanolLevel & \textbf{0.8360} & 0.8080 & 0.8140 \\
9 & Fish & 0.9658 & 0.9658 & \textbf{0.9829} \\
10 & GunPoint & \textbf{1.0000} & \textbf{1.0000} & \textbf{1.0000} \\
11 & InsectWingbeatSound & 0.6727 & 0.6646 & \textbf{0.6737} \\
12 & ItalyPowerDemand & 0.9620 & 0.9591 & \textbf{0.9650} \\
13 & MelbournePedestrian & \textbf{0.9763} & 0.9730 & 0.9751 \\
14 & MiddlePhalanxTW & 0.5130 & 0.5325 & \textbf{0.5714} \\
15 & MixedShapesRegularTrain & 0.9736 & \textbf{0.9765} & 0.9761 \\
16 & OSULeaf & 0.9834 & \textbf{1.0000} & 0.9917 \\
17 & Trace & \textbf{1.0000} & \textbf{1.0000} & \textbf{1.0000} \\
18 & WordSynonyms & 0.7461 & \textbf{0.7570} & 0.7492 \\ \hline
\multicolumn{2}{c|}{Avg. Rank} & 2.11 & 1.89 & \textbf{1.39} \\ 
\multicolumn{2}{c|}{Win} & 5 & 8 & \textbf{11} \\ \bottomrule
\end{tabular}
\end{table}

% \usepackage{booktabs}
\begin{table}[h]
\centering
\caption{Detailed test classification accuracy results on 18 UCR time series datasets using different $n_{\text{list}}$ for retrieval within R2R.  
The best results are in \textbf{bold}.}
\resizebox{\textwidth}{!}{
\label{tab:append-nlist-18ucr}
\begin{tabular}{@{}cl|ccccccc@{}}
\toprule
ID & \multicolumn{1}{c}{Dataset name} &  $n_{\text{list}}$=128 &  $n_{\text{list}}$=256 &  $n_{\text{list}}$=512 &  $n_{\text{list}}$=1024 &  $n_{\text{list}}$=2048 &  $n_{\text{list}}$=4096 &  $n_{\text{list}}$=8192 \\ \midrule
1 & ArrowHead & 0.9200 & 0.9257 & 0.9314 & 0.9314 & 0.9314 & \textbf{0.9543} & 0.9486 \\
2 & CBF & \textbf{1.0000} & \textbf{1.0000} & \textbf{1.0000} & \textbf{1.0000} & \textbf{1.0000} & \textbf{1.0000} & \textbf{1.0000} \\
3 & CricketX & 0.8283 & \textbf{0.8308} & 0.8026 & 0.8232 & 0.8283 & 0.7975 & 0.8283 \\
4 & DistalPhalanxOutlineAgeGroup & 0.7842 & \textbf{0.7914} & 0.7842 & 0.7886 & 0.7770 & 0.7698 & 0.7555 \\
5 & DistalPhalanxOutlineCorrect & 0.8007 & 0.7862 & \textbf{0.8079} & 0.8043 & 0.7971 & 0.8007 & \textbf{0.8079} \\
6 & ECG5000 & 0.9444 & 0.9462 & 0.9462 & \textbf{0.9509} & 0.9431 & 0.9480 & 0.9391 \\
7 & EOGVerticalSignal & 0.5415 & 0.5442 & \textbf{0.5497} & 0.5332 & 0.5470 & 0.5111 & 0.5055 \\
8 & EthanolLevel & 0.8040 & 0.8140 & \textbf{0.8160} & 0.8020 & 0.8120 & 0.7880 & 0.7920 \\
9 & Fish & 0.9772 & 0.9829 & 0.9772 & 0.9772 & 0.9829 & 0.9886 & \textbf{0.9943} \\
10 & GunPoint & \textbf{1.0000} & \textbf{1.0000} & \textbf{1.0000} & \textbf{1.0000} & \textbf{1.0000} & \textbf{1.0000} & \textbf{1.0000} \\
11 & InsectWingbeatSound & 0.6576 & \textbf{0.6737} & 0.6712 & 0.6641 & 0.6682 & 0.6732 & 0.6727 \\
12 & ItalyPowerDemand & \textbf{0.9708} & 0.9650 & 0.9582 & 0.9660 & 0.9670 & 0.9621 & 0.9689 \\
13 & MelbournePedestrian & 0.9731 & 0.9751 & 0.9747 & \textbf{0.9763} & 0.9735 & 0.9751 & 0.9759 \\
14 & MiddlePhalanxTW & 0.6078 & 0.5714 & 0.5559 & \textbf{0.6143} & 0.5689 & 0.5624 & 0.5819 \\
15 & MixedShapesRegularTrain & 0.9761 & 0.9761 & 0.9761 & \textbf{0.9765} & \textbf{0.9765} & 0.9761 & 0.9761 \\
16 & OSULeaf & \textbf{0.9917} & \textbf{0.9917} & \textbf{0.9917} & \textbf{0.9917} & \textbf{0.9917} & 0.9876 & 0.9876 \\
17 & Trace & \textbf{1.0000} & \textbf{1.0000} & \textbf{1.0000} & \textbf{1.0000} & \textbf{1.0000} & \textbf{1.0000} & \textbf{1.0000} \\
18 & WordSynonyms & 0.7492 & 0.7492 & 0.7492 & 0.7461 & 0.7477 & 0.7539 & \textbf{0.7571} \\ \hline
\multicolumn{2}{c|}{Avg. Rank} & 3.39 & \textbf{2.72} & 3.11 & 2.94 & 3.33 & 3.67 & 3.11 \\
\multicolumn{2}{c|}{Win} & 5 & 7 & 7 & \textbf{8} & 5 & 4 & 6 \\ \bottomrule
\end{tabular}}
\end{table}

% \usepackage{booktabs}
\begin{table}[h]
\centering
\caption{Detailed test classification accuracy results on 18 UCR time series datasets using different Top-$K$ retrieval embeddings within R2R.  
The best results are in \textbf{bold}.}
\label{tab:append-topk-18ucr}
% \resizebox{\textwidth}{!}{
\begin{tabular}{@{}cl|cccccc@{}}
\toprule
ID & Dataset name & K=1 & K=5 & K=10 & K=20 & K=32 & K=64 \\ \midrule
1 & ArrowHead & 0.9128 & \textbf{0.9543} & 0.9371 & 0.9257 & 0.9428 & 0.9486 \\
2 & CBF & \textbf{1.0000} & \textbf{1.0000} & \textbf{1.0000} & \textbf{1.0000} & \textbf{1.0000} & \textbf{1.0000} \\
3 & CricketX & 0.8287 & \textbf{0.8461} & 0.8385 & 0.8308 & 0.8410 & \textbf{0.8461} \\
4 & DistalPhalanxOutlineAgeGroup & 0.7626 & 0.7770 & \textbf{0.7914} & \textbf{0.7914} & 0.7770 & 0.7842 \\
5 & DistalPhalanxOutlineCorrect & 0.7608 & 0.7789 & 0.7681 & \textbf{0.7862} & 0.7789 & 0.7391 \\
6 & ECG5000 & 0.9409 & 0.9398 & 0.9467 & 0.9462 & \textbf{0.9495} & 0.9438 \\
7 & EOGVerticalSignal & 0.5470 & 0.5359 & 0.5387 & 0.5442 & 0.5470 & \textbf{0.5497} \\
8 & EthanolLevel & 0.8060 & \textbf{0.8280} & 0.8200 & 0.8140 & \textbf{0.8280} & 0.8060 \\
9 & Fish & 0.9715 & 0.9715 & 0.9658 & \textbf{0.9829} & 0.9772 & 0.9715 \\
10 & GunPoint & \textbf{1.0000} & 0.9934 & \textbf{1.0000} & \textbf{1.0000} & \textbf{1.0000} & \textbf{1.0000} \\
11 & InsectWingbeatSound & 0.6666 & 0.6697 & 0.6757 & 0.6737 & 0.6732 & \textbf{0.6762} \\
12 & ItalyPowerDemand & 0.9601 & 0.9572 & 0.9640 & \textbf{0.9650} & 0.9523 & 0.9620 \\
13 & MelbournePedestrian & 0.9682 & 0.9727 & 0.9723 & \textbf{0.9751} & 0.9731 & 0.9715 \\
14 & MiddlePhalanxTW & 0.5259 & \textbf{0.5714} & 0.5519 & \textbf{0.5714} & 0.5454 & 0.5064 \\
15 & MixedShapesRegularTrain & 0.9741 & 0.9720 & 0.9736 & 0.9761 & \textbf{0.9774} & 0.9732 \\
16 & OSULeaf & 0.9793 & \textbf{0.9917} & 0.9793 & \textbf{0.9917} & 0.9834 & \textbf{0.9917} \\
17 & Trace & \textbf{1.0000} & \textbf{1.0000} & \textbf{1.0000} & \textbf{1.0000} & \textbf{1.0000} & \textbf{1.0000} \\
18 & WordSynonyms & 0.7429 & \textbf{0.7492} & 0.7398 & \textbf{0.7492} & 0.7351 & 0.7445 \\ \hline
\multicolumn{2}{c|}{Avg. Rank} & 4.11 & 3.00 & 3.17 & \textbf{2.06} & 2.67 & 2.89 \\
\multicolumn{2}{c|}{Win} & 3 & 8 & 4 & \textbf{11} & 6 & 7 \\ \bottomrule
\end{tabular}
\end{table}

\section{Details of multivariate time series classification results}\label{appendG}

R2R is designed for univariate TSC, and both the query encoder and the knowledge base operate on univariate series. To examine whether the framework also transfers to multivariate time series, we follow the channel-adapter protocol of Mantis~\cite{feofanov2025mantis} and evaluate on the same ten UEA datasets~\cite{bagnall2018uea} reported in its Appendix Table 12. A trainable linear adapter maps the $d$ input channels of each dataset to ten mixed channels before the R2R$_{\text{UniShape}}$ pipeline, so that heterogeneous channel counts and cross-channel mixing are handled by the adapter, while the per-channel query embeddings are averaged before retrieval to keep the retrieval module identical to the univariate case.

Table~\ref{tab:append-uea-mv} reports the detailed multivariate time series classification results. The three Mantis columns correspond to the None, PCA, and Linear channel adapters and are copied directly from Appendix Table 12 of~\cite{feofanov2025mantis}. The two remaining columns are obtained in this work and differ only in whether retrieval is used: \textit{UniShape (Linear)} applies the same adapter and backbone without the query encoder, whereas \textit{R2R$_{\text{UniShape}}$ (Linear)} additionally retrieves numerical knowledge; the two share the same adapter, backbone, ensembling, training budget, and model-selection protocol. Both Mantis and the two UniShape-based settings are trained on the default UEA training splits and evaluated on the default test splits, and neither uses the test set during training or for weight selection, so the comparison is fair with respect to data usage.

As in the main text, we report the average rank of the test classification accuracy and the number of datasets on which each method is best (Win). R2R$_{\text{UniShape}}$ obtains the best average rank (1.40) and the most wins (6 of 10), improving over the parameter-only \textit{UniShape (Linear)} variant (2.10 with 4 wins) as well as over all three Mantis channel adapters. These results indicate that the retrieval-to-recognition paradigm remains effective once a channel adapter supplies multivariate inputs. We regard this as a preliminary extension rather than a full multivariate design, since it covers ten datasets and the retrieved knowledge itself is still univariate.

\begin{table}[h]
\centering
\caption{Test classification accuracy on ten UEA multivariate datasets under the linear channel-adapter setting. The Mantis columns are taken from Appendix Table 12 of~\cite{feofanov2025mantis}, and all accuracies are reported to three decimals. The best results are in \textbf{bold}.}
\label{tab:append-uea-mv}
\resizebox{\textwidth}{!}{
\begin{tabular}{@{}cl|ccc|cc@{}}
\toprule
\multirow{2}{*}{ID} & \multirow{2}{*}{Dataset name} & \multicolumn{3}{c|}{Mantis} & \multicolumn{2}{c}{UniShape-based (ours)} \\ \cmidrule(lr){3-5} \cmidrule(l){6-7}
 & & None & PCA & Linear & UniShape (Linear) & R2R$_{\text{UniShape}}$ (Linear) \\ \midrule
1 & CharacterTrajectories & 0.976 & 0.977 & 0.988 & \textbf{0.995} & \textbf{0.995} \\
2 & DuckDuckGeese & 0.540 & 0.580 & 0.600 & 0.600 & \textbf{0.627} \\
3 & EigenWorms & 0.814 & 0.863 & 0.855 & 0.898 & \textbf{0.906} \\
4 & FaceDetection & 0.549 & 0.553 & 0.592 & \textbf{0.664} & 0.657 \\
5 & HandMovementDirection & 0.297 & 0.338 & 0.338 & 0.351 & \textbf{0.446} \\
6 & Handwriting & 0.281 & 0.315 & \textbf{0.514} & 0.464 & 0.503 \\
7 & MotorImagery & 0.467 & 0.480 & 0.500 & \textbf{0.570} & 0.520 \\
8 & NATOPS & 0.889 & \textbf{0.939} & 0.906 & 0.909 & 0.926 \\
9 & SelfRegulationSCP2 & 0.517 & 0.533 & 0.533 & 0.494 & \textbf{0.546} \\
10 & SpokenArabicDigits & 0.937 & 0.955 & 0.987 & \textbf{0.995} & \textbf{0.995} \\
\hline
\multicolumn{2}{c|}{Avg. Rank} & 4.90 & 3.30 & 2.80 & 2.10 & \textbf{1.40} \\
\multicolumn{2}{c|}{Win} & 0 & 1 & 1 & 4 & \textbf{6} \\ \bottomrule
\end{tabular}}
\end{table}

\section{Limitations}\label{appendH}

R2R is specifically positioned at the post-training stage of TSFMs: it improves how an already pretrained model is adapted to downstream classification, and therefore contributes neither new pretraining objectives and backbone architectures nor evidence about pretraining scale.
R2R relies on a knowledge base constructed from large-scale pretraining datasets to retrieve class-relevant embeddings for downstream time series classification. In this work, both the retrieval mechanism and the knowledge base are designed for univariate time series, and thus do not explicitly account for multivariate settings, where modeling inter-variable dependencies and handling high-dimensional channels are important. Appendix~\ref{appendG} reports a preliminary extension in which a trainable linear channel adapter precedes the R2R pipeline on ten UEA datasets; there, inter-variable dependencies are absorbed by the adapter and the retrieved knowledge remains univariate, so the extension does not yet constitute a native multivariate design. Extending R2R to multivariate scenarios, including efficient channel selection and cross-variable representation learning, is a promising direction for future work. In addition, the effectiveness of retrieval-based enhancement depends on the quality of learned embeddings and the composition of the knowledge base. In particular, imbalanced class distributions in the sampled knowledge base may reduce the benefit of retrieval, indicating that diversity and balance of class patterns are important for robust knowledge transfer. This suggests future efforts on designing more robust retrieval strategies and knowledge base construction methods to further improve generalization in complex real-world settings.

\section{Broader impacts}\label{appendI}

This work proposes R2R, a retrieval-to-recognition post-training framework that enhances pretrained time series foundation models (TSFMs) by incorporating retrieved numerical knowledge from a constructed pretraining knowledge base. By combining implicit parameter knowledge in TSFMs with explicit retrieved feature-level knowledge, R2R provides a plug-and-play mechanism to improve time-series classification across diverse real-world scenarios. It can be seamlessly integrated with existing TSFMs and applied to applications such as activity recognition, healthcare monitoring, and industrial fault diagnosis. The framework relies solely on time-series representations and publicly available benchmark datasets, without involving any personal or sensitive information, and therefore does not introduce additional ethical risks beyond standard machine learning practice.

%%%%%%%%%%%%%%%%%%%%%%%%%%%%%%%%%%%%%%%%%%%%%%%%%%%%%%%%%%%%

\end{document}